\documentclass{article}

\PassOptionsToPackage{numbers, compress}{natbib}

 \usepackage[main, final]{neurips_2026}
\usepackage[utf8]{inputenc} % allow utf-8 input
\usepackage[T1]{fontenc}    % use 8-bit T1 fonts
\usepackage{hyperref}       % hyperlinks
\usepackage{url}            % simple URL typesetting
\usepackage{booktabs}       % professional-quality tables
\usepackage{amsfonts}       % blackboard math symbols
\usepackage{nicefrac}       % compact symbols for 1/2, etc.
\usepackage{microtype}      % microtypography
\usepackage{xcolor}         % colors

\usepackage{amsmath}
\usepackage{amssymb}
\usepackage{enumitem}
\usepackage{mathtools}
\usepackage{amsthm}
\usepackage{wrapfig}
\usepackage{algorithm}
\usepackage{algpseudocode}
\usepackage{pdflscape}
\usepackage{subcaption}
\usepackage{multirow}
\usepackage{multicol}
\usepackage{tikz}
\usepackage{array}
\usepackage[capitalize,noabbrev]{cleveref}
\usepackage[textsize=tiny]{todonotes}

\usepackage{amsmath,amsfonts,bm}

\def\eqref#1{equation~\ref{#1}}
\def\1{\bm{1}}

\def\vmu{{\bm{\mu}}}
\def\vtheta{{\bm{\theta}}}
\def\va{{\bm{a}}}

\def\vc{{\bm{c}}}

\def\vg{{\bm{g}}}
\def\vh{{\bm{h}}}

\def\vm{{\bm{m}}}

\def\vu{{\bm{u}}}
\def\vv{{\bm{v}}}

\def\vx{{\bm{x}}}

\DeclareMathAlphabet{\mathsfit}{\encodingdefault}{\sfdefault}{m}{sl}
\SetMathAlphabet{\mathsfit}{bold}{\encodingdefault}{\sfdefault}{bx}{n}

\def\gB{{\mathcal{B}}}

\def\gD{{\mathcal{D}}}

\def\gF{{\mathcal{F}}}

\def\gH{{\mathcal{H}}}

\def\gL{{\mathcal{L}}}

\def\gQ{{\mathcal{Q}}}

\def\gU{{\mathcal{U}}}

\newcommand{\E}{\mathbb{E}}

\newcommand{\R}{\mathbb{R}}

\definecolor{seablue}{RGB}{0, 105, 148}
\hypersetup{
    colorlinks=true,
    linkcolor=seablue,
    citecolor=seablue,
    urlcolor=seablue,
    filecolor=seablue
}

\definecolor{highlightblue}{RGB}{231, 56, 110}

\definecolor{gold}{rgb}{1.0, 0.84, 0.0}
\definecolor{silver}{rgb}{0.75, 0.75, 0.75}
\definecolor{bronze}{rgb}{0.8, 0.5, 0.2}

\newcommand{\tikzcircle}[2][black,fill=none]{%
    \tikz[baseline=-0.75ex]\draw[#1] (0,0) circle (#2);%
}

\newcommand{\goldmedal}{\tikzcircle[gold,fill=gold]{3pt}}
\newcommand{\silvermedal}{\tikzcircle[silver,fill=silver]{3pt}}
\newcommand{\bronzemedal}{\tikzcircle[bronze,fill=bronze]{3pt}}
\newcommand{\emptymedal}{\tikzcircle[draw=none,fill=none]{3pt}}

\theoremstyle{plain}
\newtheorem{theorem}{Theorem}[section]
\newtheorem{lemma}[theorem]{Lemma}
\newtheorem{proposition}[theorem]{Proposition}
\newtheorem{corollary}[theorem]{Corollary}
\theoremstyle{definition}

\theoremstyle{remark}

\title{Greedy Alignment Principle\\for Optimizer Hyperparameter Selection}

\author{%
  Jaerin Lee \& Kyoung Mu Lee \\
  Computer Vision Lab, ASRI\\
  Seoul National University \\
  Seoul 08826, Korea \\
  \texttt{\{ironjr,kyoungmu\}@snu.ac.kr}
}

\begin{document}

\maketitle

\begin{abstract}
Recent works have shown that gradient--update alignment is a powerful signal for modulating optimizer updates, often leading to faster training.
We promote this update-wise heuristic as a mathematically grounded principle for selecting and tuning optimizer hyperparameters.
By treating gradients and updates as signals and an optimizer as a causal filter that maps between them, we formulate optimizer selection as maximizing the expected drop rate in loss over a prescribed family of optimizers.
We show that this objective is exactly the inner product between the optimizer filter and the gradient autocorrelation, and prove that a greedy optimum exists and has a stability bound under perturbations of the estimated gradient statistics.
Specializing in momentum-based optimizers, the theory yields simple dynamic momentum selection rules for both SGD+Momentum and Adam/AdamW.
Experiments across image classification, language model fine-tuning, and vision transformer fine-tuning show that the resulting dynamic momentum rules match or improve upon the best fixed hyperparameters found via manual sweeps, reducing the need for exhaustive momentum sweeps.
\end{abstract}

\section{Introduction}
\label{sec:1_intro}
\begin{wrapfigure}{r}{0.48\linewidth}
\centering
\vspace{-1.2em}
\includegraphics[width=\linewidth]{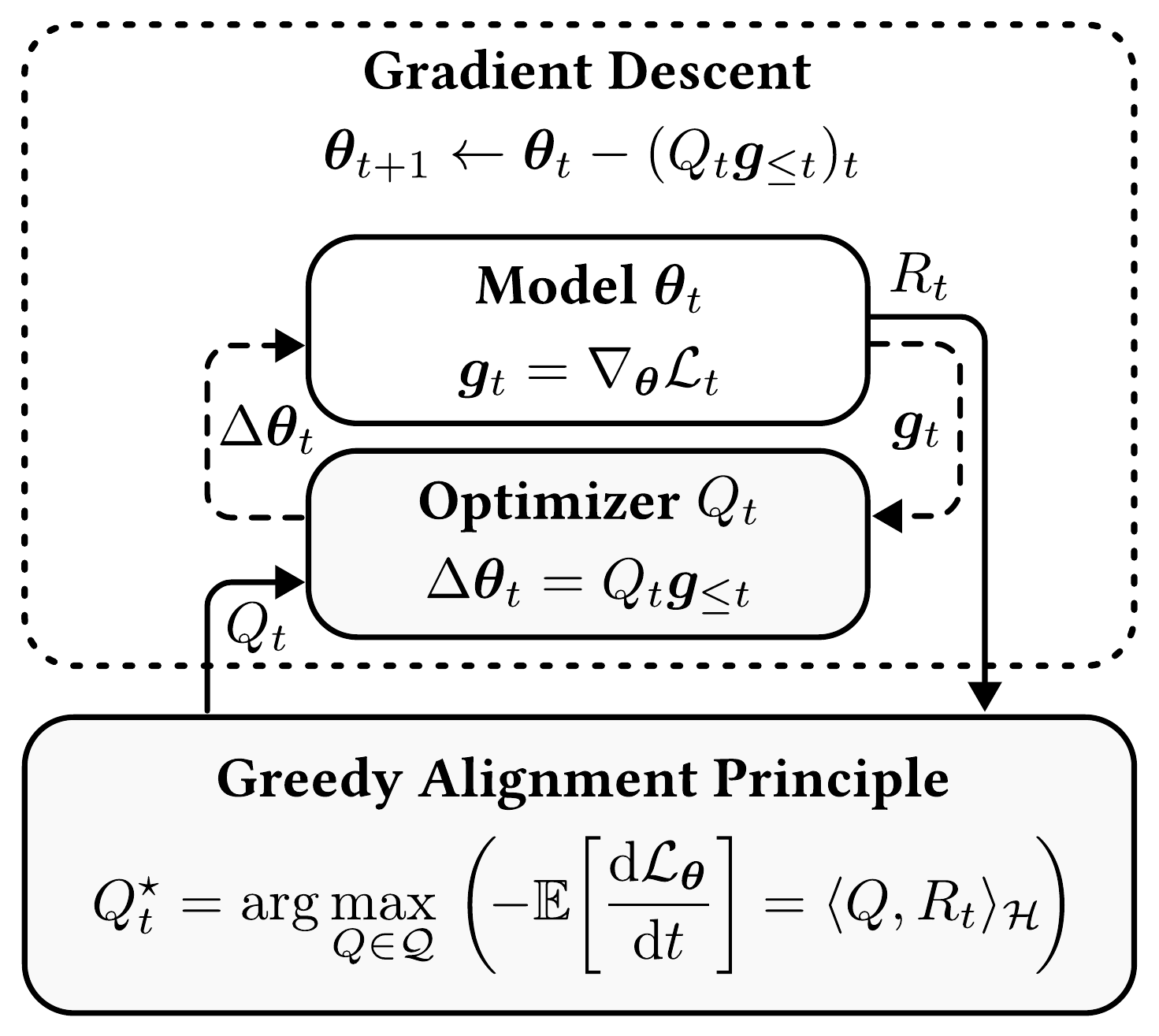}
\vspace{-1.4em}
\caption{%
\small
Just as optimizers train models by parameter updates $\vu$, models can also select an optimizer from a candidate set $\mathcal{Q}$ using the gradient statistics $R$.
}
\label{fig:snakes}
\vspace{-1em}
\end{wrapfigure}

Gradient-based learning has been the \emph{de facto} standard in machine learning practice.
Over the past decades, first-order methods have matured into a rich collection of optimizers.
Dozens of gradient-based optimizers have been proposed in the literature~\cite{robbins1951stochastic,kingma2015adam,loshchilov2017decoupled,chen2023symbolic,yuan2025mars,dozat2016incorporating,liang2026cautious,chang2025mgup,liu2024sophia,pethick2025training,vyas2025soap} and are widely used in practice, improving upon the baseline SGD~\cite{robbins1951stochastic} in their specialized tasks.
Although their philosophies and design choices are distinct, many of them share the same principle of momentum-based filtering of incoming gradient streams~\cite{lee2024grokfast} to stabilize learning dynamics.
The momentum hyperparameter $\beta$ is a crucial design choice that governs the trade-off between stability and learning speed.
However, the selection of this hyperparameter is highly dependent on the underlying task and the type of optimizers used.
Still, the choice is largely left to the practitioner's intuition and trial-and-error.

To address this problem, we consider a different line of work that post-processes the parameter updates produced by optimizers.
Notably, Cautious optimizer~\cite{liang2026cautious} selectively applies the parameter update components based on the sign of coordinate-wise gradient-update products.
MGUP~\cite{chang2025mgup} further uses momentum-gradient alignment to reweight the update components.
These works empirically demonstrate that the \emph{instantaneous alignment} $\vg^\top \vu$ between gradients $\vg$ and parameter updates $\vu$ is an informative signal for assessing, controlling, and improving the learning dynamics of existing gradient-based optimizers~\cite{robbins1951stochastic,kingma2015adam,loshchilov2017decoupled,chen2023symbolic,jordan2024muon}.

This update-wise greedy alignment strategy can be formulated as follows.
In gradient-based learning, we minimize a loss $\gL(\vtheta)$ with respect to the parameter $\vtheta$ using its gradient $\vg = \nabla_\vtheta \gL$.
At each training step $t$, the parameter $\vtheta$ is updated by $\vtheta_{t{+}1} \leftarrow \vtheta_t - \vu_t$.
The heuristics of greedy alignment can then be formulated into an optimization problem:
\begin{equation}
    \label{eq:sec_1:principle}
    \vu_t^\star
    \;=\;\arg \underset{u_t\in\gU_t}{\max}\ \vg_t^\top \,\vu_t,
\end{equation}
where $\gU_t$ is the set of parameter updates proposed in step $t$.
Interestingly, by the chain rule, this problem reduces to maximizing the \emph{instantaneous loss drop rate}:
\begin{equation}
    \label{eq:sec_1:instantaneous_loss_drop}
    \vg_t^\top \vu_t
    \;=\;-\nabla_\vtheta \gL^\top \, \frac{\mathrm d \vtheta}{\mathrm d t}
    \;=\;-\frac{\mathrm d \gL}{\mathrm d t}.
\end{equation}
This partially justifies the community wisdom~\cite{liang2026cautious,chang2025mgup} built on greedy alignment of updates.

Despite their elegance, the idea of greedily maximizing alignment has been applied only by modifying the outputs of fixed optimizers.
We promote this idea into a mathematically grounded principle for selecting and tuning the optimizers themselves.
Adopting the signal processing analogy~\cite{lee2024grokfast}, we treat gradients $\vg_t$ as a time-varying signal.
Similarly, the parameter update $\vu_t$ is a function of the causal gradient stream $\vg_{\le t}$.
Then, the optimizer $Q$ can be regarded as a \emph{signal processing filter} $Q: \vg \mapsto \vu$, rather than an algorithmic procedure, mapping the gradient stream $\vg$ to the update stream $\vu$:
\begin{equation}
    \label{eq:sec_1:operator}
    \vu_t
    \;=\;(Q\vg)_t.
\end{equation}
This filter view of the optimizer $Q$ enables us to apply the greedy alignment principle at the operator-level.
We further specialize to stable causal linear filters $Q$, which includes SGD~\cite{robbins1951stochastic} with momentum as a canonical special case.
Then, the expected alignment becomes a Hilbert inner product between the optimizer filter $Q$ and the gradient autocorrelation $R_{t,k} = \E[\vg_{t}\,\vg_{t{-}k}^\top]$, as elaborated in Section~\ref{sec:2_prelim:problem_statement},
\begin{equation}
    \label{eq:sec_1:optimizer_alignment}
    \E[\vg_t^\top \vu_t]
    \;=\;\langle Q, R \rangle_{\gH}.
\end{equation}
In other words, update-wise \emph{expected} alignment is equivalent to \emph{alignment between the optimizer filter $Q$ and the gradient statistics $R$}.
Allowing $Q$ to be time-varying and input-adaptive, this framework also handles preconditioned methods such as AdamW~\cite{loshchilov2017decoupled}.

Lifting the greedy alignment principle to this filter viewpoint unlocks a theoretical toolbox, especially useful in regimes where the \emph{global optimality criterion is nearly impossible to achieve}.
The heuristic practice of selecting optimizer hyperparameters $\phi$ is now mathematically grounded as an optimization problem within a prescribed family of optimizers $Q\in\gQ_\phi$, which is also equivalent to maximizing the instantaneous expected loss drop rate.
As illustrated in Figure~\ref{fig:snakes}, this dual problem of rate maximization coexists with the primary problem of loss minimization.
By merging equations \ref{eq:sec_1:principle} and \ref{eq:sec_1:optimizer_alignment}, we reveal two design factors that determine the best optimizer according to the greedy alignment principle:
(1) the \emph{hyperparameterized family of optimizers} $Q\in\gQ_\phi$, and (2) the \emph{gradient autocorrelation} $R$.
Therefore, the problem reduces further to selecting the hyperparameter $\phi$ based on the gradient statistics $R$.
In Section~\ref{sec:2_prelim:general_solutions}, we prove that this greedy optimum is well-defined, attained under compactness assumptions, and characterized by the support function of the candidate family.
Section~\ref{sec:3_main} focuses on momentum-based optimizers, a large family of optimizers of our special interest~\cite{robbins1951stochastic,kingma2015adam,loshchilov2017decoupled}.
Section~\ref{sec:4_impl} discusses practical concerns in applying the theory, and experiments in Section~\ref{sec:5_exp} support the resulting dynamic selection rules in this focused case.

In summary, our contributions are as follows:
\begin{itemize}[leftmargin=2em,topsep=0pt]
    \item We lift update-wise greedy alignment rules to an \emph{optimizer-level principle} by modeling a first-order optimizer as a causal filter and formulate optimizer selection as maximizing the expected first-order loss drop rate $-\mathrm d \gL / \mathrm d t$ against the gradient autocorrelation $R$.
    \item We prove that, for causal linear filters and compact candidate families, this problem is well-defined, attains an optimum, and is stable under perturbations of the estimated gradient statistics.
    \item We show that optimizer hyperparameterization reduces this problem into scoring hyperparameters, yielding a principled basis for dynamic hyperparameter selection.
    \item Applying this framework to momentum-based optimizers~\cite{robbins1951stochastic,kingma2015adam,loshchilov2017decoupled}, we derive dynamic selection rules that exhibit matched or improved results compared with manually tuned fixed-momentum baselines, reducing the need for expensive hyperparameter sweeps.
\end{itemize}

\section{Gradient-update alignment to filter-statistics alignment}
\label{sec:2_prelim}
\subsection{Problem statement}
\label{sec:2_prelim:problem_statement}

As mentioned in the previous section, we formalize the \emph{greedy alignment principle} for gradient-based optimizers, starting from the update-wise principle in~\eqref{eq:sec_1:principle}.
Let an optimizer $Q$ be a \emph{causal filter} that translates a gradient stream $\vg_{\le t}$ into an update $\vu_t = -\mathrm d \vtheta / \mathrm dt$.
We then choose the optimizer according to the following maximization problem
\begin{equation}
    \label{eq:sec_2:principle}
    Q^\star
    \;=\;\arg \underset{Q\in\gQ}{\max}\ \E[\vg_t^\top \,\vu_t], \qquad
    \E[\vg_t^\top \,\vu_t]
    \;=\;\E[\vg_t^\top \,(Q\vg)_t]
    \;=\;-\E\!\left[\frac{\mathrm d \gL}{\mathrm d t}\right]
    \;\eqqcolon\; P,
\end{equation}
where $\gQ$ is the candidate set of optimizers.
This lifts the update rule of \eqref{eq:sec_1:principle} to an optimizer-level selection criterion within the prescribed family of optimizers $\gQ$.
For simplicity of notation, define the \emph{learning power} $P \coloneqq -\E[\mathrm d \gL / \mathrm d t]$ as the expected speed of loss drop.
Note that we are deriving the theory in the continuous-time training step $t$ for compactness.
The validity of the results in discrete-time is straightforward and is provided in Section~\ref{sec:3_main:finite-step-descent-bridge}.

Consider a linear dynamic optimizer $Q$ operating as a causal convolution
\begin{equation}
    \label{eq:sec_2:dynamic_optimizer}
    \vu_t
    \;=\; (Q \vg)_t
    \;=\; (Q_t * \vg_{\le t})_t
    \;=\; \sum_{k = 0}^{\infty} Q_{t,k} \, \vg_{t{-}k}.
\end{equation}
This formulation generalizes first-order optimizers, especially those with momentum terms.
Define the \emph{gradient autocorrelation} as $R_{t,k} \coloneqq \E[\vg_t \, \vg_{t{-}k}^\top]$, for $k \ge 0$.
Then $Q$ and $R$ are both elements of an admissible Hilbert space $\gH$ of square-summable causal sequences
\begin{equation}
\label{eq:sec_2:admissible-hilbert}
\gH
\;=\;
\Bigl\{ A = (A_k)_{k \ge 0}
\;:\;
\sum_{k \ge 0} \operatorname{Tr}(A_k^\top A_k) < \infty \Bigr\},
\qquad
\langle A, B \rangle_{\gH}
\;=\;
\sum_{k \ge 0} \operatorname{Tr}(A_k^\top B_k),
\end{equation}
where we can define an inner product between optimizer transfer function $Q$ and gradient autocorrelation $R$.
This is exactly the learning power $P$ in~\eqref{eq:sec_2:principle}.

\begin{proposition}[Learning power is inner product]
\label{prop:learning-power-as-inner-product}
\textnormal{[\texttt{\hyperref[sec:appx:proof-of-proposition-learning-power-as-inner-product]{proof}}]}
Assume that
(A1) the gradient stream $\vg_t$ has finite and uniformly bounded second moments for all $t$: $\sup_{t\in \mathbb Z} \E\|\vg_t\|^2 < \infty$;
(A2) the optimizer filter $Q_t$ is absolutely summable in operator norm: $\sum_{k=0}^{\infty} \|Q_{t,k}\|_{\text{op}} < \infty$;
and (A3) the autocorrelation sequence $R_t = \{R_{t,k}\}_{k\ge 0}$ belongs to $\gH$.
Then for every $t$,
\begin{equation}
    \label{eq:sec_2:instantaneous_power}
    P_t(Q)
    \;=\; \E\big[\vg_t^\top \vu_t\big]
    \;=\; \sum_{k=0}^{\infty} \operatorname{Tr}\!\left(Q_{t,k}^\top R_{t,k}\right)
    \;=\; \langle Q_t, R_t \rangle_{\gH}.
\end{equation}
\end{proposition}

Note that the assumptions (A1-3) generally hold for any stable machine learning system.
The filter-level principle in \eqref{eq:sec_2:principle} can be rewritten into the following maximization problem:
\begin{equation}
    \label{eq:sec_2:optimization_problem}\tag{$\spadesuit$}
    \renewcommand{\boxed}[1]{\colorbox{lightgray!15}{\ensuremath{#1}}}
    \boxed{\ %
    Q_t^\star
    \;=\; \arg \underset{Q\in\gQ}{\max}\ \langle Q, R_t \rangle_{\gH}.\ %
    }
\end{equation}
As long as the set of candidate optimizers $\gQ$ is convex and bounded, selecting the best optimizer under the greedy alignment principle (\ref{eq:sec_2:optimization_problem}) is a \emph{convex optimization problem with a linear objective}, of which we have a rich understanding~\citep{boyd2004convex}.
Note that $\gQ$ should be bounded to avoid arbitrarily large learning rates.
Furthermore, the optimal optimizer $Q_t^\star$ is a function of the (1) candidate set $\gQ$ and the (2) gradient statistics $R_t$.
Note the subscript $t$.
This formulation allows us to dynamically adapt the optimizer $Q$ with respect to time-varying $R_t$.

\subsection{General solutions}
\label{sec:2_prelim:general_solutions}

We start by showing that the solution of problem~(\ref{eq:sec_2:optimization_problem}) is indeed attainable.
The following definitions are required for the theory's formalism.
Let a \emph{Minkowski gauge} $\gamma_{\gQ}(Q)$ and a \emph{polar set} $\gQ^\circ$ be
\begin{equation}
    \label{eq:sec_3:polar_set_and_gauge}
    \gamma_{\gQ}(Q)
    \;\coloneqq\; \inf\{c > 0 : Q \in c \gQ\},\qquad
    \gQ^\circ
    \;\coloneqq\; \{ R\in \gH \mid {\sup}_{Q\in\gQ} \langle Q,R\rangle_{\gH} \le 1\},
\end{equation}
In simpler terms, $\gamma_{\gQ}(Q)$ measures the minimum scaling of a convex set $\gQ$ to contain $Q$, and $\gQ^\circ$ is a set of gradient statistics $R$ that upper-bounds the inner product by one.
Then we can define a \emph{symmetrized polar gauge} $\|\cdot\|_{\gQ^\circ}^{\mathrm{sym}}$ as
\begin{equation}
    \label{eq:sec_3:symmetrized_polar_gauge}
    \|R\|_{\gQ^\circ}^{\mathrm{sym}}
    \;\coloneqq\; \max\{\gamma_{\gQ^\circ}(R),\,\gamma_{\gQ^\circ}(-R)\}.
\end{equation}
We are now ready to introduce our first theorem.
The common subscript $t$ is omitted for brevity.

\begin{theorem}[Greedy optimal first-order optimizers]
\label{thm:dynamic-optimizer-under-convex-constraints}
\textnormal{[\texttt{\hyperref[sec:appx:proof-of-theorem-dynamic-optimizer-under-convex-constraints]{proof}}]}
Let $\gH$ be a Hilbert space.
Let $\gQ \subset \gH$ be a compact nonempty convex set of causal optimizer filters with $0 \in \gQ$.
Then for every $R \in \gH$,
\begin{enumerate}[label=(\roman*),leftmargin=2em]
\item \emph{(Existence):}
The maximum $P^\star (R)$ of problem~(\ref{eq:sec_2:optimization_problem}) is attained, sublinear, and finite everywhere.

\item \emph{(Construction):}
Any optimal optimizer $Q^\star$ is a subgradient of $P^\star = \gamma_{\gQ^\circ}$ at $R$:
$Q^\star \in \partial_R P^\star(R)$.
If the maximizer is unique, $P^\star$ is Gâteaux differentiable at $R$ and $Q^\star = \nabla_{\!R}\, P^\star(R)$.

\item \emph{(Lipschitz continuity):}
The estimated optimal learning power under surrogate gradient statistics $\hat{R}$ is bounded by
$|P^\star(R)-P^\star(\hat{R})|\le \|R-\hat{R}\|_{\gQ^\circ}^{\mathrm{sym}}, \forall R,\hat{R}\in\gH$.
\end{enumerate}
\end{theorem}

Its proof is in Appendix~\ref{sec:appx:proofs}.
From Theorem~\ref{thm:dynamic-optimizer-under-convex-constraints}, we guarantee that:
(1) the optimizer selection problem under the principle of greedy alignment is well-defined, and
(2) there exists the optimal optimizer $Q_t^\star$, and
(3) Lipschitz continuity gives us a stability bound of the optimal learning power $P^\star$ when we use the estimated gradient statistics $\hat{R}$ (for example, autocorrelations from a mini-batch stream) instead of the true statistics $R$.
It should be noted that the optimal solution to the problem~(\ref{eq:sec_2:optimization_problem}) is constructed from the candidate family $\gQ$ and the gradient statistics $R$ in the form $Q^\star = \partial_{R}\, \gamma_{\gQ^\circ}(R)$.
The exact optimality is constructible for various families of optimizers, as we elaborate in Appendix~\ref{sec:appx:families}.

Specifically, the candidate family $\gQ$ can be characterized by a set of hyperparameters $\phi$, (e.g., EMA momentum $\beta$).
These hyperparameters later describe the greedy optimal optimizer $Q^\star$ as well.
In other words, the optimizer-selection principle in \eqref{eq:sec_2:principle} is reducible to a mathematical framework to choose \emph{optimizer hyperparameters}.
We make this formal.
Define a closed convex hull $\operatorname{cl}\operatorname{co}(S)$ of a set $S \subset \gH$ as the closure of all finite convex combinations
\begin{equation}
\label{eq:sec_2:cl-co}
\operatorname{cl}\operatorname{co}(S) \;\coloneqq\; \overline{\operatorname{co} (S)}, \;\; \text{where} \;\;
\operatorname{co}(S) \;\coloneqq\; \Big\{
\sum_{i=1}^{m} \lambda_i s_i : m < \infty, s_i \in S, \lambda_i \ge 0, \sum_{i=1}^{m} \lambda_i = 1
\Big\}.
\end{equation}
This allows us to study hyperparameterized settings as a convex optimization framework.

\begin{corollary}[Family-hull reduction]
\label{cor:family-hull-reduction}
\textnormal{[\texttt{\hyperref[sec:appx:proof-of-corollary-family-hull-reduction]{proof}}]}
Let $\Phi$ be a nonempty set of hyperparameters and $\gF_\Phi = \{Q_\phi : \phi \in \Phi\} \subset \gH$ a bounded family of optimizer filters.
Define $\gQ_\Phi \coloneqq \operatorname{cl}\operatorname{co}(\gF_\Phi \cup \{0\})$.
Then for every $R \in \gH$,
\begin{equation}\label{eq:sec_3:family-hull}
\sup_{Q \in \gQ_\Phi} \langle Q, R \rangle_{\gH}
\;=\;
\Big(\sup_{\phi \in \Phi} \E[\vg_t^\top (Q_\phi \vg)_t]\Big)_{\!+}\!,
\end{equation}
where $(x)_+ \coloneqq \max\{x, 0\}$.
If $\Phi$ is finite, the supremum is a maximum.
\end{corollary}

This maximization of the expected alignment $\E[\vg_t^\top (Q_\phi \vg)_t]$ over the governing hyperparameter $\phi$ is an \emph{exact reduction} of the original problem~(\ref{eq:sec_2:optimization_problem}), not a heuristic approximation.
The learning power $P$ is linear in $Q$.
Provided that the set of hyperparameters $\Phi$ is compact and the mapping $\phi \mapsto Q_\phi$ is continuous, this convex program on a convex set $\gQ_\Phi$ therefore attains its maximum at its extreme points, namely in $\gF_\Phi \cup \{0\}$.
We can therefore apply Theorem~\ref{thm:dynamic-optimizer-under-convex-constraints} to a hyperparameterized family of optimizers.
This allows us to treat various existing families of optimizers as mathematically grounded objects.
In the main text, we focus on linear causal optimizers for simplicity, but the theory can be generalized to nonlinear causal optimizers, as we elaborate in Appendix~\ref{sec:appx:more_math}.

\section{Momentum families and local validity}
\label{sec:3_main}
\subsection{Hyperparameter selection for momentum-based optimizers}
\label{sec:3_main:momentum}

We instantiate the general theory on the most commonly used nontrivial family of optimizers: momentum-based ones.
Many practical optimizers, including SGD+Momentum~\cite{robbins1951stochastic} and Adam/AdamW~\cite{kingma2015adam,loshchilov2017decoupled}, are special cases of this class.
This class of optimizers is particularly interesting because a one-pole IIR filter can represent the EMA momentum:
\begin{equation}
    \label{eq:sec_3:ema-filter}
    \gQ_{\text{1p}} \coloneqq \{Q_{\beta,k} \;=\; \eta\beta^k I : \eta \ge 0, \beta \in \gB \subset [0,1)\}, \qquad
    (Q_{\beta} \vg)_t \;=\; \eta \sum_{k=0}^\infty \beta^k \vg_{t{-}k},
\end{equation}
where $\gB$ is compact, e.g., $\gB = [0, \bar\beta]$ or a finite set of candidates.
The family of 1-pole optimizer filters $\gQ_{\text{1p}}$ is a mathematical description of the family of EMA momentum-based first-order optimizers.
Define the unnormalized momentum $\vm_{\beta,t} \coloneqq \beta \vm_{\beta,t{-}1} + \vg_t$ at time $t$ with the momentum hyperparameter $\beta$.
If we constrain the family to be norm-bounded, i.e., $\gQ_\textnormal{H}(B) \coloneqq \{Q : \|Q\|_\gH {\le} \sqrt{B}\}$ and $\gQ_{\eta,\beta} \coloneqq \gQ_{\text{1p}} \cap \gQ_\textnormal{H}(B)$, then we get the SGD+Momentum optimizer with optimal hyperparameters.

\begin{theorem}[Greedy optimal SGD+Momentum]
\label{thm:greedy-optimal-sgdm}
\textnormal{[\texttt{\hyperref[sec:appx:proof-of-theorem-greedy-optimal-sgdm]{proof}}]}
Consider the norm-bounded family $\gQ_{\eta,\beta} = \gQ_{\text{1p}} \cap \gQ_\textnormal{H}(B)$ of optimizer filters.
Then problem~(\ref{eq:sec_2:optimization_problem}) under the prescribed family $\gQ_{\eta,\beta}$ reduces to the \emph{SGD} optimizer with \emph{greedy optimal momentum hyperparameter}:
\begin{equation}\label{eq:optimal-sgdm}
\renewcommand{\boxed}[1]{\colorbox{lightgray!15}{\ensuremath{#1}}}
\boxed{\ 
\beta^\star_t \;=\; \arg\underset{\beta\in\gB}{\max} \sqrt{1 - \beta^2}\,\, \mathbb{E}[\vg_t^\top  \vm_{\beta,t}],
}
\end{equation}
\end{theorem}

The score in Theorem~\ref{thm:greedy-optimal-sgdm} acts as a normalized probe of the gradient autocorrelation.
Note that the norm budget $B$ disappears.
The factor $\sqrt{1{-}\beta^2}$ comes from the norm of the Hilbert space $\gH$ that defines the inner product between the optimizer $Q_t$ and the statistics $R_t$.
This factor additionally removes trivial preferences for filters with either no memory ($\beta = 0$) or infinite memory ($\beta = 1$).
The following corollary is a neat theoretical result that uncovers the connection between the momentum hyperparameter $\beta$ and the gradient statistics, enriching the meaning of \emph{filter--statistics alignment}.

\begin{corollary}[One-pole gradients recover one-pole momentum]
\label{cor:ar1-momentum}
\textnormal{[\texttt{\hyperref[sec:appx:proof-of-corollary-ar1-momentum]{proof}}]}
Assume that the autocorrelation of the gradient stream $\vg$ has the one-pole scalar trace autocorrelation function:
\begin{equation}
    \label{eq:sec_3:ar1-momentum}
    r_k \;=\; \mathbb E[\vg_t^\top\vg_{t-k}] \;=\; r_0\rho^k, \qquad r_0>0, \quad 0\le\rho<1.
\end{equation}
Then the SGD+Momentum score in Theorem~\ref{thm:greedy-optimal-sgdm} becomes
\begin{equation}
    \label{eq:sec_3:ar1-momentum-score}
    \sqrt{1-\beta^2}\, \mathbb E[\vg_t^\top\vm_{\beta,t}] \;=\; r_0\tfrac{\sqrt{1-\beta^2}}{1-\beta\rho}.
\end{equation}
The maximum is attained where the gradient pole $\rho$ is aligned with the momentum pole $\beta$:
$\beta^\star=\rho$.
Moreover, if the range is restricted to $\beta\in[0,\bar\beta]$ with $0<\bar\beta<1$, then $\beta^\star=\max \{0, \min \{\bar\beta, \rho\}\}$.
\end{corollary}

In other words, the momentum hyperparameter $\beta$ is more than an arbitrary smoothing coefficient.
In this one-pole case, its greedy optimum $\beta^\star$ exactly recovers the temporal coherence of the gradients $\vg$.

Furthermore, we can easily extend Theorem~\ref{thm:greedy-optimal-sgdm} to widely used preconditioned optimizers such as Adam~\citep{kingma2015adam}.
Let normalized first and second moments $\vmu_{\beta_1,t}$ and $\vv_{\beta_2,t}$ as:
\begin{equation}
    \label{eq:3_main:adam-normalized-moments}
    \vmu_{\beta_1,t} \;=\; \beta_1 \vmu_{\beta_1,t{-}1} + (1-\beta_1) \vg_t, \qquad
    \vv_{\beta_2,t} \;=\; \beta_2 \vv_{\beta_2,t{-}1} + (1-\beta_2) \vg_t^2.
\end{equation}
We treat the Adam denominator as a time-varying cost vector $c_j \coloneqq (v_{\beta_2,j} + \epsilon)^{1/2}$, where $j$ is the parameter coordinate index and $\epsilon > 0$ is a small regularization term.
Define an elementwise weighted norm-bounded family $\gQ_\textnormal{D}(B,\vc)$ and an Adam 1-pole family $\gQ_\textnormal{1p}(\vc)$ of optimizer filters.
\begin{equation}
    \label{eq:3_main:adam-helper}
    \gQ_\textnormal{D} \coloneqq \big\{\operatorname{diag}(q_{j}) : \sum_j c_j \sum_{k\ge 0} |q_{j,k}|^2{\le} B\big\}, \;\;
    \gQ_\textnormal{1p}(\vc) \coloneqq \big\{q_{j,k} = \eta (1{-}\beta_1) \beta_1^k c_j^{-1} : \eta {\ge} 0, 0 {<} \beta_1 {<} 1\big\}.
\end{equation}
Corollary~\ref{cor:greedy-optimal-adam} then extends Theorem~\ref{thm:greedy-optimal-sgdm} to Adam~\citep{kingma2015adam}.
In this corollary, we condition on the realized second-moment state $\vv_{\beta_2,t}$ for each candidate $\beta_2$ and view Adam as a local time-varying filter.

\begin{corollary}[Greedy optimal Adam/AdamW]
\label{cor:greedy-optimal-adam}
\textnormal{[\texttt{\hyperref[sec:appx:proof-of-corollary-greedy-optimal-adam]{proof}}]}
Consider the weighted norm-bounded family $\gQ_{\eta,\beta_1, \beta_2} = \gQ_\textnormal{D}(B,\vc) \cap \gQ_\textnormal{1p}(\vc)$  of optimizer filters.
Then problem~(\ref{eq:sec_2:optimization_problem}) under the prescribed family $\gQ_{\eta,\beta_1, \beta_2}$ reduces to the \emph{Adam/AdamW} optimizer with \emph{greedy optimal hyperparameters}:
\begin{equation}\label{eq:optimal-adam}
\renewcommand{\boxed}[1]{\colorbox{lightgray!15}{\ensuremath{#1}}}
\boxed{\ 
(\beta_1^\star, \beta_2^\star)_t \;=\; \arg\underset{\beta_1, \beta_2\in(0,1)^2}{\max} 
\sqrt{\frac{1+\beta_1}{1-\beta_1}}\frac{1}{\sqrt{{\sum_j c_{\beta_2, t, j}^{-1}}}} \,\mathbb{E} [\vg_t^\top  \vu_{\beta_1, \beta_2,t}],
}
\end{equation}
where $c_{\beta_2, t, j} = \sqrt{v_{\beta_2,t,j} + \epsilon}$ and $\vu_{\beta_1, \beta_2,t} \coloneqq \vmu_{\beta_1,t} \odot \vc_{\beta_2,t}^{-1}$ is the Adam update direction without bias correction and decoupled weight decay, which is applied separately in AdamW.
The expectation is understood conditionally on the realized second-moment state $\vc_{\beta_2,t}$.
\end{corollary}

All the proofs are in Appendix~\ref{sec:appx:proofs}.
This derivation shows that the greedy optimal hyperparameters are not universal constants but functions of the gradient autocorrelation structure.
In practice, training algorithms are online and therefore approximate the expectation using either the instantaneous score $\vg_t^\top \vu_t$ or the average of recent batch scores.
Theorem~\ref{thm:greedy-optimal-sgdm} and Corollary~\ref{cor:greedy-optimal-adam} show that the greedy optimal hyperparameters that maximize the expected loss drop are attained by maximizing the alignment score between the gradients and the optimizer responses for both SGD+Momentum and Adam/AdamW.
This opens a new interpretation of these algorithms from the filter viewpoint: SGD+Momentum can be interpreted as an \emph{optimal 1-pole approximation of the norm-bounded optimizer}, and Adam/AdamW can be interpreted as an \emph{optimal 1-pole approximation of the dynamic diagonal optimizer with a preconditioner $\vc_{\beta_2,t}^{-1}$}.

\subsection{Finite-step descent bridge}
\label{sec:3_main:finite-step-descent-bridge}

So far, we have derived the theory based on~\eqref{eq:sec_2:principle}, which assumes continuous-time gradient and update streams.
In reality, training is performed step by step, leaving a gap between the continuous-time theory and the discrete-time algorithm.
The following theorem closes this gap.

\begin{figure}[t]
\begin{minipage}[t]{0.45\textwidth}
\vspace{-9.6em}
\begin{algorithm}[H]
\caption{$K$-Switch SGD+Momentum}
\label{alg:optimal_sgd_momentum}
\begin{algorithmic}[1]
\Require $\eta$, $\{\beta_k\}_{k=1}^K$
\State $\vmu_k \leftarrow \mathbf{0}$ for all $k$
\For{each step $n$}
    \State $\vg \leftarrow \nabla_\theta \gL(\vtheta)$
    \For{$k = 1, \ldots, K$}
        \State $\vmu_k \leftarrow \beta_k \vmu_k + \vg$
        \State \textcolor{highlightblue}{$A_k \leftarrow \sqrt{1 - \beta_k^2}\; \vg^\top \vmu_k$}
    \EndFor
    \State \textcolor{highlightblue}{$k^\star \leftarrow \arg\max_k A_k$}
    \State $\vtheta \leftarrow \vtheta - \eta \vmu_{k^\star}$
\EndFor
\end{algorithmic}
\end{algorithm}
\end{minipage}
\hfill
\begin{minipage}[t]{0.48\textwidth}
\centering
\begin{subfigure}[b]{\textwidth}
\centering
\includegraphics[width=\textwidth]{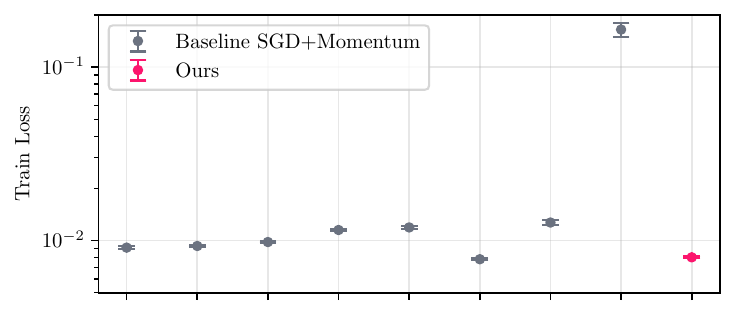}
\end{subfigure}
\\[-0.5em]
\hfill
\begin{subfigure}[b]{.99\textwidth}
\centering
\includegraphics[width=\textwidth]{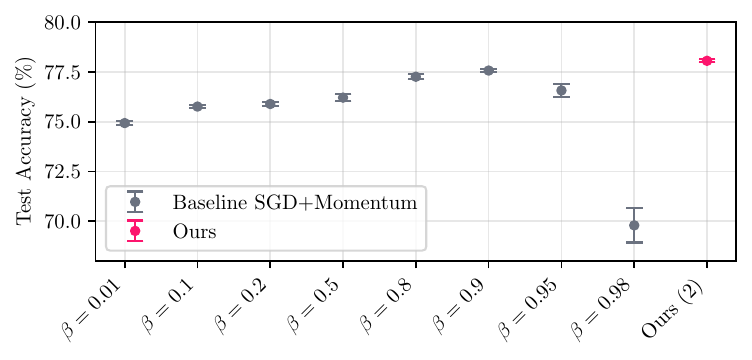}
\end{subfigure}
\end{minipage}
\vspace{-1em}
\caption{\small
Empirical instantiation of Theorem~\ref{thm:greedy-optimal-sgdm} for SGD+Momentum.
The greedy optimal SGD+Momentum of Algorithm~\ref{alg:optimal_sgd_momentum} is compared with fixed-hyperparameter baselines on the CIFAR-100 dataset~\citep{krizhevsky2009learning} with ResNet-18~\citep{he2016deep}.
The error bars indicate the mean and standard deviation over 10 runs.}
\vspace{-1em}
\label{fig:sgd_momentum_test_acc_errbar}
\end{figure}

\begin{table}[t]
\centering
\small
\begin{minipage}[t]{0.49\textwidth}
\caption{\small Demonstration of Theorem~\ref{thm:greedy-optimal-sgdm} for SGD with momentum.
Best baseline at $\beta = 0.9$.
mean $\pm$ std.}
\label{tab:sgd_momentum_results}
% \vspace{-0.5em}
\centering
\resizebox{.95\textwidth}{!}{%
\begin{tabular}{lccc}
\toprule
\textbf{Method} & Test acc. \% & Train loss \\
\midrule
Best baseline & 77.57 $\pm$ 0.09 \emptymedal & 0.0078 $\pm$ 0.0001 \goldmedal \\
\textbf{Ours} & 78.06 $\pm$ 0.07 \goldmedal & 0.0080 $\pm$ 0.0001 \emptymedal \\
\bottomrule
\end{tabular}%
}
\end{minipage}
\hfill
\begin{minipage}[t]{0.49\textwidth}
\caption{\small
Demonstration of Corollary~\ref{cor:greedy-optimal-adam} for Adam.
Best baseline at $(\beta_1, \beta_2) = (0.8, 0.999)$.
mean $\pm$ std.}
\label{tab:adam_results}
% \vspace{-0.5em}
\centering
\resizebox{.95\textwidth}{!}{%
\begin{tabular}{lccc}
\toprule
\textbf{Method} & Test acc. \% & Train loss \\
\midrule
Best baseline & 73.20 $\pm$ 0.21 \emptymedal & 0.0324 $\pm$ 0.0042 \emptymedal \\
\textbf{Ours} & 73.26 $\pm$ 0.31 \goldmedal & 0.0115 $\pm$ 0.0010 \goldmedal \\
\bottomrule
\end{tabular}%
}
\end{minipage}
% \vspace{-1.0em}
\end{table}

\begin{theorem}[Finite-step descent]
\label{thm:finite-step-descent-bridge}
\textnormal{[\texttt{\hyperref[sec:appx:proof-of-theorem-finite-step-descent-bridge]{proof}}]}
Let $\gL : \R^d \to \R$ be differentiable with the $L_s$-Lipschitz gradient.
Fix $\vtheta \in \R^d$ and let $\vg \coloneqq \nabla_\vtheta \gL$.
For any update $\vu \in \R^d$ and learning rate $\eta > 0$, define the alignment score $A(\vu) \coloneqq \vg^\top \vu$ and the one-step loss drop $D_\eta(\vu) \coloneqq \gL(\vtheta) - \gL(\vtheta - \eta \vu)$.
Then
\begin{equation}\label{eq:sec_4:descent_bridge}
\bigl| D_\eta(\vu) - \eta A(\vu) \bigr|
\;\le\;
\frac{L_s \eta^2}{2} \|\vu\|^2.
\end{equation}
\end{theorem}

Therefore, $\vg^\top \vu$ is the correct first-order term of the actual one-step loss decrease.
The approximation is reliable as long as the local quantity $\eta \|\vu\|$ is small enough.
Outside this regime, the alignment score remains a first-order approximation but should be paired with a norm control or other stabilization.
The following corollary further solidifies the greedy alignment as a reliable local decision rule.

\begin{corollary}[Near-optimality of greedy alignment]
\label{cor:greedy-near-optimal}
\textnormal{[\texttt{\hyperref[sec:appx:proof-of-corollary-greedy-near-optimal]{proof}}]}
Let $\mathcal{U}$ be a candidate set of updates with $\|\vu\| \le \rho$ for all $\vu \in \mathcal{U}$.
Let $\vu^\star \in \arg\max_{\vu \in \mathcal{U}} A(\vu)$.
Then
\begin{equation}\label{eq:sec_4:greedy_near_optimal}
D_\eta(\vu^\star) \;\ge\; \max_{\vv \in \mathcal{U}} D_\eta(\vv) - L_s \eta^2 \rho^2.
\end{equation}
In particular, if $L_s \eta^2 \rho^2$ is small, then maximizing $A(\vu) = \vg^\top \vu$ is an $O(\eta^2)$-accurate surrogate for maximizing the one-step decrease.
\end{corollary}

Corollary~\ref{cor:greedy-near-optimal} says that over norm-bounded candidate updates, greedy maximum gives a near-optimal choice for the true one-step loss decrease.
In other words, as long as the bound of the local regime $\eta \|\vu\|$ remains small enough, greedy alignment is a reliable local decision rule.
The operator-level greedy objective (\ref{eq:sec_2:optimization_problem}) is an expected inner product $P(Q) = \E[\vg_t^\top (Q\vg)_t]$.
The online implementation replaces this expectation by an (averaged) instantaneous score $A_t(\vu) \coloneqq \vg_t^\top \vu_t$.
Theorem~\ref{thm:finite-step-descent-bridge} and Corollary~\ref{cor:greedy-near-optimal} justify the use of $\vg^\top \vu$ as a local surrogate by showing that since the second-order error is controlled in the local regime, greedy alignment based on $\vg^\top \vu$ is a reliable first-order approximation of the actual loss decrease.

\section{Realization and practical concerns}
\label{sec:4_impl}
\subsection{A finite-candidate realization of greedy optimizer selection}
\label{sec:4_impl:meta_algorithm}

\begin{figure}[t]
\begin{minipage}[t]{0.47\textwidth}
\vspace{-10.8em}
\begin{algorithm}[H]
\caption{$K$-Switch Adam}
\label{alg:optimal_adam}
\begin{algorithmic}[1]
\Require $\eta$, $\{(\beta_{1,k}, \beta_{2,k})\}_{k=1}^K$
\State $\vmu_k, \vv_k \leftarrow \mathbf{0}$ for all $k$
\For{each step $t$}
    \State $\vg \leftarrow \nabla_\theta \gL(\vtheta)$
    \For{$k = 1, \ldots, K$}
        \State $\vv_k \leftarrow \beta_{2,k} \vv_k + (1{-}\beta_{2,k}) \vg^2$
        \State $\vmu_k \leftarrow \beta_{1,k} \vmu_k + (1{-}\beta_{1,k}) \vg$
        \State $\vu_k \leftarrow \vmu_k / (\sqrt{\vv_k + \epsilon})$
        \State \textcolor{highlightblue}{$a_k \leftarrow \sqrt{\tfrac{1{+}\beta_{1,k}}{(1{-}\beta_{1,k})(\sum_j\! (v_{k,j} + \epsilon)^{-1/2})}}$}
        \State \textcolor{highlightblue}{$A_k \leftarrow a_k\, \vg^\top \vu_k$}
    \EndFor
    \State \textcolor{highlightblue}{$k^\star \leftarrow \arg\max_k A_k$}
    \State $\vtheta \leftarrow \vtheta - \eta\, \vu_{k^\star}$
\EndFor
\end{algorithmic}
\end{algorithm}
\end{minipage}
\hfill
\begin{minipage}[t]{0.53\textwidth}
\centering
\begin{subfigure}[b]{\textwidth}
\centering
\includegraphics[width=\textwidth]{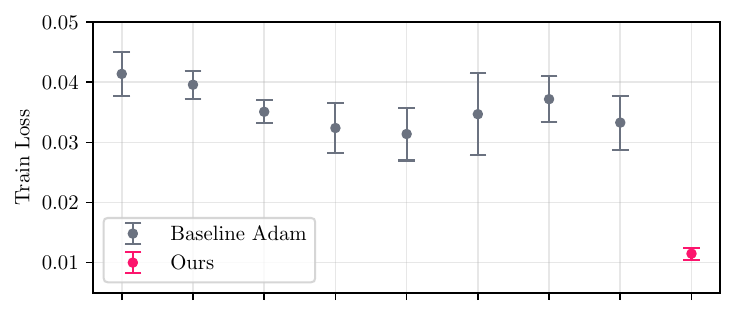}
\end{subfigure}
\\[-0.5em]
\hfill
\begin{subfigure}[b]{0.985\textwidth}
\centering
\includegraphics[width=\textwidth]{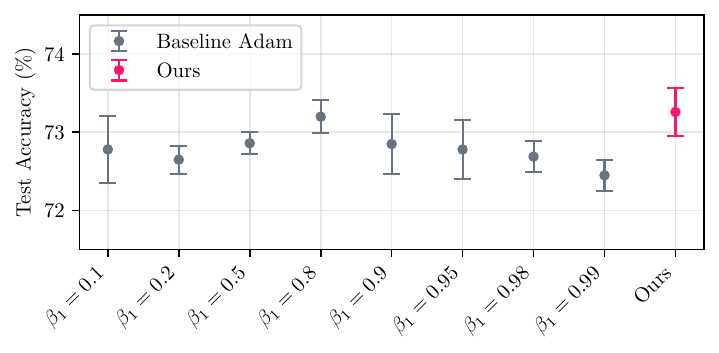}
\end{subfigure}
\end{minipage}
\caption{\small
Empirical instantiation of Corollary~\ref{cor:greedy-optimal-adam} for Adam~\citep{kingma2015adam}.
The greedy optimal Adam of Algorithm~\ref{alg:optimal_adam} is compared with fixed-hyperparameter baselines on the CIFAR-100 dataset~\citep{krizhevsky2009learning} with ResNet-18~\citep{he2016deep}.
The error bars indicate the mean and standard deviation over 10 runs.}
\label{fig:adam_test_acc_errbar}
\vspace{-1em}
\end{figure}

\begin{table}[t]
\centering
\small
\begin{minipage}[t]{0.49\textwidth}
\caption{\small AdamW on Gemma-2B with MetaMathQA-395K validated on GSM8K.
mean $\pm$ std.}
\label{tab:gemma_mathqa}
% \vspace{-0.5em}
\centering
\resizebox{.95\textwidth}{!}{%
\begin{tabular}{lccc}
\toprule
\textbf{Method} & Test acc. \% & Train loss \\
\midrule
Best baseline & 52.57 $\pm$ 1.10 \emptymedal & 0.2080 $\pm$ 0.0004 \goldmedal \\
\textbf{Ours} & 52.77 $\pm$ 0.93 \goldmedal & 0.2084 $\pm$ 0.0003 \emptymedal \\
\bottomrule
\end{tabular}%
}
\end{minipage}
\hfill
\begin{minipage}[t]{0.49\textwidth}
\caption{\small
AdamW on Llama-3-8B with MetaMathQA-395K dataset.
mean $\pm$ std.}
\label{tab:llama_mathqa}
% \vspace{-0.5em}
\centering
\resizebox{.95\textwidth}{!}{%
\begin{tabular}{lccc}
\toprule
\textbf{Method} & Test acc. \% & Train loss \\
\midrule
Best baseline & 76.20 $\pm$ 0.33 \emptymedal & 0.1927 $\pm$ 0.0005 \emptymedal \\
\textbf{Ours} & 76.30 $\pm$ 0.31 \goldmedal & 0.1925 $\pm$ 0.0005 \goldmedal \\
\bottomrule
\end{tabular}%
}
\end{minipage}
% \vspace{-1em}
\end{table}

This section provides a simple realization of the greedy optimal algorithm for optimizer selection, dubbed $K$-Switch.
Rather than attempting a full hyperparameter sweep over the entire training dynamics, $K$-Switch is an online algorithm for the finite-candidate greedy alignment objective.
The argmax operator in the theory can be implemented in multiple ways; we explore the simplest possible option by selecting from a fixed set of $K$ candidate optimizers.
For each candidate $k = 1, \ldots, K$, we maintain an optimizer state $Q_{k,t}$ and compute the candidate update $\vu_{k,t} = (Q_{k,t} \vg_{\leq t})_t$.
Define normalized candidates $\tilde Q_{k,t}$ by $(\tilde Q_{k,t} \vg_{\leq t})_t = a_{k,t} \vu_{k,t}$, with normalization factors $a_{k,t} > 0$ specific to the prescribed family of optimizers, e.g., $a_{k,t} = \sqrt{1-\beta_k^2}$ for SGD+Momentum~\citep{robbins1951stochastic}.
Generally, we find the candidate that maximizes the normalized alignment score $\bar A_{k,t} \coloneqq \E [a_k \vg_t^\top \vu_{k,t}]$:
\begin{equation}\label{eq:sec_4:selected_candidate}
k_t^\star = \arg\max_{k \in \{1,\ldots,K\}} \bar A_{k,t}.
\end{equation}
By the family-hull reduction in Corollary~\ref{cor:family-hull-reduction}, optimizing over the convex hull of these candidates strictly reduces to selecting the best single candidate over the $K$ options:

\begin{corollary}[Exact finite-candidate $K$-Switch]
\label{cor:kswitch-exact}
\textnormal{[\texttt{\hyperref[sec:appx:proof-of-corollary-kswitch-exact]{proof}}]}
For $P_t(Q) \coloneqq \E[\vg_t^\top (Q\vg)_t]$,
\begin{equation}\label{eq:sec_4:kswitch_exact}
\sup_{Q\in\operatorname{co}\{\tilde Q_{k,t}\}_{k=1}^K}P_t(Q)
\;=\;
\max_k P_t(\tilde Q_{k,t})
\;=\;
\max_k \E[a_{k,t}\vg_t^\top \vu_{k,t}].
\end{equation}
\end{corollary}

The implication is straightforward: $K$-Switch is not a heuristic approximation, but the exact finite-candidate solution of Theorem~\ref{thm:dynamic-optimizer-under-convex-constraints} over the convexified normalized family.

\subsection{Practical concerns}
\label{sec:4_impl:practical_concerns}

\paragraph{Online algorithm and stability.}
In practice, the exact expectation $\bar A_{k,t}$ is inaccessible, and we must rely on the mini-batch estimate $\hat A_{k,t} \coloneqq \E_{\text{batch}} [a_k \vg_t^\top \vu_{k,t}]$.
The following proposition ensures that this statistical approximation remains stable.

\begin{proposition}[Online stability]
\label{prop:online-stability}
\textnormal{[\texttt{\hyperref[sec:appx:proof-of-proposition-online-stability]{proof}}]}
Let $\hat A_k$ be an online estimate of $A_k$, and assume $\max_k|\hat A_k-A_k|\le\varepsilon$.
If $\hat k\in\arg\max_k\hat A_k$ and $k^\star\in\arg\max_k A_k$, then $A_{k^\star}-A_{\hat k}\le2\varepsilon$.
If the best candidate is unique with a gap $>2\varepsilon$, then $\hat k=k^\star$.
\end{proposition}

Proposition~\ref{prop:online-stability} translates the Lipschitz stability of Theorem~\ref{thm:dynamic-optimizer-under-convex-constraints}(iii) from $P^\star(R)$ to a finite argmax over the unit-energy family $\gF = \{Q / \|Q\|: Q\text{ an optimizer}\}$.
Thus, the resulting $K$-Switch algorithms in Algorithms~\ref{alg:optimal_sgd_momentum} and~\ref{alg:optimal_adam} are robust streaming estimators of the exact finite-candidate greedy alignment maximizer.
In practice, we can further stabilize the $K$-Switch algorithms by the following modifications.
This leads to Algorithms~\ref{alg:optimal_sgd_momentum} and~\ref{alg:optimal_adam}.
For clarity, Algorithm~\ref{alg:optimal_adam} shows the gradient-driven Adam update and omits bias correction and decoupled weight decay. 
In AdamW experiments, the alignment score is computed only on the gradient-driven update, while the decoupled weight decay is applied identically across all candidates after the selected update.
Lines highlighted in \textcolor{highlightblue}{pink} are the actual augmentations to the baseline algorithms; all other lines are unchanged.

\paragraph{Separating the roles of the optimizer and the learning rate scheduler.}
In practical scenarios, the parameter update $\vu_t$ is scaled by a learning rate scheduler $\eta_t$ which is usually implemented externally to the optimizers.
Respecting this common design choice, we decouple hyperparameter selection from parameter innovation.
The Hilbert-norm factor $a_{k,t}$ is used exclusively in the hyperparameter selection stage, comparing different optimizer hyperparameters on an equal-energy trust region to identify the best-aligned filter pole (e.g., $\beta$) without introducing scale bias (Corollary~\ref{cor:ar1-momentum}).
Once $k_\star$ is selected, we can update the parameters via $\vtheta_{t+1} \leftarrow \vtheta_t - \eta_t \vu_{k^\star,t}$, with $a_{k,t}$ absorbed in the externally set learning rate $\eta_t$.
This ensures compatibility with existing frameworks using learning rate scheduling and prevents this additional step-size modulation from interfering with the directional alignment of the gradient $\vg_t$ and parameter update $\vu_t$.
This is shown in Algorithms~\ref{alg:optimal_sgd_momentum} and~\ref{alg:optimal_adam}.

\section{Experiments}
\label{sec:5_exp}
\begin{table}[t]
\centering
\small
\caption{\small AdamW on Gemma-2B with Commonsense-170K dataset.
mean $\pm$ std.}
\label{tab:llm_commonsense}
% \vspace{-0.5em}
\resizebox{1.0\textwidth}{!}{%
\begin{tabular}{l|cccccc|c}
\toprule
\textit{Gemma-2B (LoRA)} & BoolQ & PIQA & Social IQA & HellaSwag & Winogrande & OBQA & \textbf{Avg} \\
\midrule
Best baseline & 65.69 $\pm$ 0.29 \goldmedal & 78.93 $\pm$ 0.49 \emptymedal & 73.61 $\pm$ 0.16 \goldmedal & 74.07 $\pm$ 0.28 \emptymedal & 71.61 $\pm$ 0.44 \emptymedal & 72.67 $\pm$ 0.68 \emptymedal & 72.76 $\pm$ 0.17 \emptymedal \\
\textbf{Ours} & 65.31 $\pm$ 0.04 \emptymedal & 79.00 $\pm$ 0.36 \goldmedal & 73.58 $\pm$ 0.06 \emptymedal & 75.09 $\pm$ 1.02 \goldmedal & 71.80 $\pm$ 0.39 \goldmedal & 73.27 $\pm$ 1.15 \goldmedal & 73.01 $\pm$ 0.27 \goldmedal \\
\midrule
\midrule
\textit{Gemma-2B (Full FT)} & BoolQ & PIQA & Social IQA & HellaSwag & Winogrande & OBQA & \textbf{Avg} \\
\midrule
Best baseline & 62.79 $\pm$ 0.27 \emptymedal & 74.12 $\pm$ 0.26 \emptymedal & 66.63 $\pm$ 0.33 \emptymedal& 40.50 $\pm$ 1.15 \emptymedal & 61.48 $\pm$ 0.32 \emptymedal & 62.60 $\pm$ 1.02 \emptymedal & 61.35 $\pm$ 0.27 \emptymedal \\
\textbf{Ours} & 63.29 $\pm$ 0.78 \goldmedal & 75.70 $\pm$ 0.22 \goldmedal & 68.41 $\pm$ 0.69 \goldmedal & 42.47 $\pm$ 1.06 \goldmedal & 62.46 $\pm$ 4.64 \goldmedal & 64.40 $\pm$ 0.86 \goldmedal & 62.79 $\pm$ 0.83 \goldmedal \\
\bottomrule
\end{tabular}%
}
\vspace{-1em}
\end{table}

\begin{table}[t]
\centering
\small
\caption{\small AdamW on Vision Transformer fine-tuning tasks with LoRA.
mean $\pm$ std.}
\label{tab:vit_tasks}
% \vspace{-0.5em}
\resizebox{1.0\textwidth}{!}{%
\begin{tabular}{l|ccccccc|c}
\toprule
\textit{ViT-B (rank = 32)} & Cars & CIFAR-100 & CUB-200 & DTD & Food-101 & RESISC45 & SUN397 & \textbf{Avg} \\
\midrule
Best baseline & 77.56 $\pm$ 0.09 \emptymedal & 91.74 $\pm$ 0.07 \emptymedal & 84.67 $\pm$ 0.06 \goldmedal & 78.32 $\pm$ 0.38 \goldmedal & 88.13 $\pm$ 0.03 \emptymedal & 94.54 $\pm$ 0.01 \goldmedal & 72.72 $\pm$ 0.08 \goldmedal & 83.95 $\pm$ 0.06 \emptymedal \\
\textbf{Ours} & 77.95 $\pm$ 0.38 \goldmedal & 91.87 $\pm$ 0.02 \goldmedal & 84.56 $\pm$ 0.13 \emptymedal & 78.23 $\pm$ 0.48 \emptymedal & 88.16 $\pm$ 0.09 \goldmedal & 94.24 $\pm$ 0.09 \emptymedal & 72.71 $\pm$ 0.21 \emptymedal & 83.96 $\pm$ 0.10 \goldmedal \\
\midrule
\midrule
\textit{ViT-L (rank = 8)} & Cars & CIFAR-100 & CUB-200 & DTD & Food-101 & RESISC45 & SUN397 & \textbf{Avg} \\
\midrule
Best baseline & 84.89 $\pm$ 0.12 \emptymedal & 93.20 $\pm$ 0.08 \goldmedal & 87.08 $\pm$ 0.21 \goldmedal & 80.04 $\pm$ 0.18 \emptymedal & 89.98 $\pm$ 0.07 \emptymedal & 95.13 $\pm$ 0.08 \goldmedal & 75.18 $\pm$ 0.10 \goldmedal & 86.50 $\pm$ 0.05 \emptymedal \\
\textbf{Ours} & 85.40 $\pm$ 0.11 \goldmedal & 93.05 $\pm$ 0.01 \emptymedal & 86.80 $\pm$ 0.12 \emptymedal & 80.74 $\pm$ 0.44 \goldmedal & 90.04 $\pm$ 0.15 \goldmedal & 95.07 $\pm$ 0.02 \emptymedal & 74.87 $\pm$ 0.08 \emptymedal & 86.57 $\pm$ 0.07 \goldmedal \\
\bottomrule
\end{tabular}%
}
% \vspace{-1em}
\end{table}

\begin{table}[t]
\centering
\small
\caption{\small Hyperparameter tuning cost comparison between $K$-Switch and full hyperparameter sweep.}
\label{tab:runtime}
% \vspace{-0.5em}
\resizebox{1.0\textwidth}{!}{%
\begin{tabular}{lcccccc}
\toprule
& ResNet-18 & Gemma-2B & Gemma-2B & Llama-3-8B & ViT-Base & ViT-Large \\
& (full model) & ($r=32$ LoRA) & (Full FT) & ($r=32$ LoRA) & ($r=32$ LoRA) & ($r=8$ LoRA) \\
\midrule
Original VRAM & 0.961 GiB & 16.303 GiB & 16.943 GiB & 19.910 GiB & 1.078 GiB & 2.049 GiB \\
$K$-switch VRAM & 1.344 GiB & 16.303 GiB & 27.920 GiB & 21.381 GiB & 1.080 GiB & 2.049 GiB \\
\textbf{Memory overhead} & $+39.84\%$ & $+0.00\%$ & $+64.78\%$ & $+7.39\%$ & $+0.18\%$ & $+0.00\%$ \\
\midrule
\# $\beta$ candidates & 8 & 6 & 6 & 6 & 6 & 6 \\
Per-step runtime & {$+14.0\%$} & {$-0.50\%$} & {$+29.9\%$} & {$+0.33\%$} & {$+2.21\%$} & {$+1.01\%$} \\
\textbf{Full sweep speedup} & {\color{green!60!black}$\mathbf{\times 7.02}$} & {\color{green!60!black}$\mathbf{\times 6.03}$} & {\color{green!60!black}$\mathbf{\times 4.62}$} & {\color{green!60!black}$\mathbf{\times 5.98}$} & {\color{green!60!black}$\mathbf{\times 5.87}$} & {\color{green!60!black}$\mathbf{\times 5.94}$} \\
\bottomrule
\end{tabular}%
}
% \vspace{-1em}
\end{table}

\paragraph{Implementation details.}
One of the practical advantages of our framework is the \emph{automatic} selection of momentum hyperparameters in a greedy sense, as shown in Theorem~\ref{thm:greedy-optimal-sgdm} and Corollary~\ref{cor:greedy-optimal-adam}.
We provide empirical justification of the theory in previous sections by the simplest instantiations of these theoretical frameworks elaborated in Section~\ref{sec:4_impl}: $K$-Switch dynamic optimizers with $K = 2$.
The candidate endpoints are chosen a priori to cover the conventional working range of the corresponding optimizer family, rather than by sweeping over all pairs.
In contrast, fixed baselines are selected by an oracle sweep over the reported grid.
Complete details and results are provided in Appendix~\ref{sec:appx:exp}.

\paragraph{ResNet-18 on CIFAR-100.}
Using this simple instantiation, we empirically demonstrate the theory by training a ResNet-18~\citep{he2016deep} model on the CIFAR-100 dataset~\citep{krizhevsky2009learning}.
Baseline optimizers are trained with fixed hyperparameters, in line with typical machine learning practices.
For the momentum of SGD+Momentum, we tried $\beta \in [0.01, 0.999]$ and reported the best.
For Adam, we tried $\beta_1 \in [0.1, 0.99]$ while keeping $\beta_2 = 0.999$ fixed and reported the best.
Figures~\ref{fig:sgd_momentum_test_acc_errbar} and~\ref{fig:adam_test_acc_errbar} and Tables~\ref{tab:sgd_momentum_results} and~\ref{tab:adam_results} summarize the results.
Our $K$-Switch dynamic optimizers achieve comparable and often better performance than the baseline optimizers with fixed hyperparameters in both final accuracy and convergence speed.
This largely reduces the need for a heavy hyperparameter sweep.

\paragraph{Large language models and vision transformers.}
We also demonstrate our framework in more practical scenarios: training large language models (LLMs) and vision transformers (ViTs).
We train a Gemma-2B~\citep{gemma2024open} and Llama-3-8B~\citep{llama32024herd} using low rank adaptation (LoRA)~\citep{hu2022lora} with standard settings on the MetaMathQA-395K dataset~\citep{yu2024metamath} and compare the results with baseline optimizers of fixed hyperparameters on GSM8K~\citep{cobbe2021gsm8k}.
Table~\ref{tab:gemma_mathqa} and~\ref{tab:llama_mathqa} summarize the results of ten runs for each model.
Furthermore, we also train a Gemma-2B with both LoRA and full fine-tuning on the Commonsense-170K dataset~\citep{hu2023llmadapters} using our and baseline optimizers of fixed hyperparameters.
We then compare the results on various reasoning tasks such as BoolQ~\citep{clark2019boolq}, PIQA~\citep{bisk2020piqa}, Social IQA~\citep{sap2019socialiqa}, HellaSwag~\citep{zellers2019hellaswag}, Winogrande~\citep{sakaguchi2019winogrande}, and OBQA~\citep{hu2023llmadapters}, and summarize the results in Table~\ref{tab:llm_commonsense}.
Finally, we train ViT-B and ViT-L models~\citep{dosovitskiy2021image} by LoRA on various image classification tasks including Cars~\citep{krause20133d}, CIFAR-100~\citep{krizhevsky2009learning}, CUB-200~\citep{wah2011cub200}, DTD~\citep{cimpoi2014describing}, Food-101~\citep{bossard2014food101}, RESISC45~\citep{cheng2017resisc45}, and SUN397~\citep{xiao2010sun}.
Table~\ref{tab:vit_tasks} summarizes the results.
Detailed settings and extended results are provided in Appendix~\ref{sec:appx:exp}.

The strongest large-model evidence appears on Commonsense-170K, where $K$-Switch AdamW improves the unweighted average over the best reported fixed-$\beta_1$ baseline by $+0.25$ points for LoRA and $+1.44$ points for full fine-tuning.
In the MetaMathQA and ViT settings, our $K$-Switch remains competitive with oracle-selected fixed-momentum baselines.

\paragraph{Computational overhead.}
Table~\ref{tab:runtime} compares total hyperparameter search cost of our $K$-Switch against fixed-$\beta$ sweeps.
We report per-step wall-clock time measured with the same fused PyTorch/Triton implementation of the optimizers.
We can reduce the additional memory requirements of $K = 2$ switch to below $+100\%$ with this implementation.
In LoRA-based fine-tuning, the additional memory overhead can be lower than $+10\%$.
For $K = 2$, the practical default, per-step runtime overhead is small in LoRA regimes and moderate in full-model runs.
Nevertheless, full sweep requires multiple runs of the same task to get the best hyperparameter settings, whereas our $K$-switch adapts the hyperparameters in real-time, requiring only a single run for each task.
Regarding the final performance, our $K$-Switch consistently outperforms the best fixed-hyperparameter settings found by dense manual sweep.
Our theory provides a practical alternative to manual hyperparameter sweeps, reducing the estimated hyperparameter tuning cost by at least $\times 4.62$ across various tasks.

\section{Limitations and scope}
\label{sec:7_limit}
\paragraph{Greedy paradigm.}
Our objective $\max_{Q\in\gQ}\langle Q,R_t\rangle_{\gH}$ is a greedy training-statistics objective that maximizes the expected first-order loss drop rate.
Thus, our theory should not be interpreted as a global convergence guarantee, a validation-loss guarantee, or a population-optimality result in arbitrary nonlinear nonconvex landscapes.
Long-horizon guarantees should be incorporated with optimizer-specific analyses of the underlying optimizer family.

\paragraph{Dependence on optimizer family.}
The framework optimizes within a user-specified feasible family $\gQ$ and, by itself, does not determine which class of optimizers is best for a given task.
In practice, this means that our method reduces the inner search over hyperparameters while still relying on a reasonable candidate family chosen by the practitioner.

\paragraph{Adaptive optimizers and overhead.}
Adaptive optimizers such as Adam/AdamW involve nonlinear state-dependent operations, so we treat them through realized update streams or local time-varying filter approximations rather than as globally linear time-invariant filters.
Our $K$-Switch implementation is a simple reference instantiation of the theory; it requires maintaining multiple candidate optimizer states, so memory and runtime costs depend on $K$, model size, etc.

\section{Conclusion}
\label{sec:8_concl}
By interpreting optimizers as statistical filters, we recast optimizer tuning into a principled greedy optimization problem: selecting the causal filter that maximizes the expected rate of loss decrease.
This dual problem, running alongside the primary problem of loss minimization, allows us to choose a greedy-optimal optimizer based on the current gradient statistics while training the model with the selected optimizer.
Specializing in momentum-based optimizers, this perspective transforms optimizer hyperparameter selection into an online alignment problem.
The resulting $K$-Switch algorithm is lightweight and empirically reduces the need for exhaustive fixed-momentum sweeps.

%---------

\begin{ack}
This work was supported in part by the IITP grants [No. RS-2021-II211343, Artificial Intelligence Graduate School Program (Seoul National University), No. RS-2025-02303870, No.2022-0-00156] funded by the Korean government (MSIT).
\end{ack}

\section*{Broader impacts}

The main goal of this paper is two-fold: (1) to reduce the wasteful time and budget currently spent on hyperparameter tuning, and (2) to broaden the accessibility of machine learning techniques by alleviating the need for such efforts.
By advancing this direction, we expect to lower the computational overhead and operational costs of large-scale machine learning systems.
Ultimately, we hope that this contributes to a more democratic and sustainable AI ecosystem, benefiting both society and the environment.

%---------

\bibliographystyle{abbrvnat}
\bibliography{references}

%---------
\newpage

\appendix

\section{Related work}
\label{sec:6_relworks}
\paragraph{Momentum, adaptive optimizers, and the tuning bottleneck.}
Momentum~\cite{polyak1964some,nesterov1983method} has been a central mechanism in first-order optimization for decades~\cite{sutskever2013importance}, and SGD~\cite{robbins1951stochastic} with momentum remains a canonical optimizer in deep learning.
In order to accelerate, stabilize, and better generalize in training large models, dozens of adaptive optimizers have been proposed to modify momentum, by preconditioning or scaling the parameter updates.
To name a few, this includes AdaGrad~\cite{duchi2011adaptive}, RMSProp~\cite{tieleman2012lecture}, Adam~\cite{kingma2015adam}, AdamW~\cite{loshchilov2017decoupled}, Nadam~\cite{dozat2016incorporating}, AMSGrad~\cite{reddi2018convergence}, Adafactor~\cite{shazeer2018adafactor}, AdaBound~\cite{luo2019adaptive}, RAdam~\cite{liu2020variance}, AdaBelief~\cite{zhuang2020adabelief}, LAMB~\cite{you2020large}, Adan~\cite{xie2024adan}, Sophia~\cite{liu2024sophia}, Lion~\cite{chen2023symbolic}, and MARS~\cite{yuan2025mars}.
Most of them introduce their own set of hyperparameters, each with different recommended ranges and tuning strategies.
Optimizer-level mechanisms such as Lookahead~\cite{zhang2019lookahead}, Schedule-Free learning~\cite{defazio2024road}, and D-Adaptation~\cite{defazio2023dadaptation} reduce reliance on manually chosen schedules or step sizes.
Despite this progress, optimizer performance is highly sensitive to the tuning procedure.
Large-scale studies~\cite{choi2019empirical,schmidt2021descending} have shown that optimizer rankings vary under different hyperparameter search spaces and that no single optimizer dominates across tasks.
We address this bottleneck from a different approach: rather than proposing a new fixed optimizer, we formulate the online selection of optimizer hyperparameters as a greedy alignment problem between an optimizer filter and the gradient autocorrelation.

\paragraph{Gradient--update alignment.}
Several works have studied the alignment between gradients and generated parameter updates to accelerate and stabilize training.
Lion~\cite{chen2023symbolic} has discovered sign-based momentum updates through symbolic search.
Cautious Optimizer framework~\cite{liang2026cautious} proposes a simple hard masking rule for momentum-based optimizers, which updates only on coordinates where the proposed optimizer update and the current gradient are positively aligned.
MGUP~\cite{chang2025mgup}, in contrast, relaxes hard rejection with step size modulation to weight the updates on high-alignment coordinates while preserving small but nonzero updates on the remaining coordinates.
These methods operate primarily at the \emph{actuation} level: given an update, they decide how much of it should be applied to each parameter coordinate.
Our framework operates one level above this: we apply alignment to select the optimizer itself.
In this sense, Cautious-style masking and MGUP-style prioritization are complementary safeguards for applying a selected update, whereas our main object is the filter-selection objective.

\paragraph{Hyperparameter optimization.}
Classical hyperparameter optimization methods, including Bayesian optimization, random search~\cite{bergstra2011algorithms}, Hyperband~\cite{li2018hyperband}, BOHB~\cite{falkner2018bohb}, and population-based training~\cite{jaderberg2017population}, tune hyperparameters by repeatedly or partially training models under different configurations.
Instead, gradient-based hyperparameter optimization methods make the training procedure differentiable or approximate hypergradients, enabling optimization of learning rates, momentum schedules, regularization coefficients, and other training hyperparameters~\cite{maclaurin2015gradient,franceschi2017forward,baydin2018online,shaban2019truncated,chandra2022gradient,chen2022optformer}.
These methods typically optimize validation performance or long-horizon objectives, often with nontrivial unrolling, meta-training, or repeated evaluations.
Our method is more lightweight, operating in local, online regimes, rather than global, offline regimes.
We use the current training gradient and candidate updates to estimate the greedy training-power objective, avoiding validation unrolling or HPO outer loops.

\paragraph{Performance-guided or learned optimizer design.}
The Performance Estimation Problem (PEP) framework analyzes or designs first-order methods through worst-case convex optimization objectives~\cite{drori2014performance,kim2017optimized_gradient,goujaud2024pepit}.
On the other hand, optimizer discovery methods search over update-rule programs or symbolic expressions, as in neural optimizer search~\cite{bello2017neural}, symbolic learning~\cite{zheng2022symbolic}, and non-parametric optimizer search~\cite{wang2022efficient}.
Learning-to-optimize approaches train optimizer networks, either RNN-based or hierarchical~\cite{andrychowicz2016learning,wichrowska2017learned}, to return update rules from gradient information.
These approaches are oriented to discovering new optimizers, whereas our framework assumes a prescribed optimizer family and gives a closed-form greedy score for selecting among its members.
Thus, our work is closer to online optimizer selection than to optimizer discovery, learning, or synthesis.

\section{Generalization beyond linear optimizers and finite-candidate exactness}
\label{sec:appx:more_math}
This section provides more detailed mathematical foundations and generalization of the main text, which were omitted for brevity and simplicity.
In the main text, we have developed the theory mainly for \emph{causal linear filters} $Q$.
Linearity gives the inner product structure of the learning power functional $P_t(Q) = \langle Q, R \rangle_{\gH}$, leading to elegant theory uncovering interpretable relationships between the optimizer $Q$ and the gradient autocorrelation $R$.
We showed that this simplest case already captures existing practical optimizers such as SGD+Momentum~\cite{robbins1951stochastic}.
However, the resulting framework relies on restrictive linearity assumptions, and extension to adaptive optimizers such as Adam~\cite{kingma2015adam} and AdamW~\cite{loshchilov2017decoupled} required additional assumptions.
Here, we generalize beyond linearity and extend the theory to general causal optimizers $Q$.
The advantage is clear: we can apply the same \emph{greedy alignment principle} to nonlinear causal optimizers as well.

We begin with the general form of the learning power functional:
\begin{equation}
    \label{eq:sec_3:learning-power-functional}
    P_t(Q) \;=\; \E[\vg_t^\top \, (Q\vg)_t]
\end{equation}
where $\vg_t$ is a gradient stream and $Q$ is a general causal optimizer.
Even though $Q$ is not a linear functional over the gradient stream $\vg_t$, the learning power functional is still linear in $Q$.
The key insight is that the additivity and homogeneity of an optimizer filter $Q$ comes from the additivity and homogeneity of the generated parameter updates $\vu_t = (Q\vg)_t$, not from the relationship between $Q$ and $\vg_t$.

\begin{lemma}[Learning power functional is linear in $Q$]
\label{lem:learning-power-functional-linear-in-Q}
For any two causal optimizers $Q_1$ and $Q_2$, and any scalars $\alpha_1$ and $\alpha_2$,
\begin{equation}
    \label{eq:sec_3:learning-power-functional-linearity}
    P_t(\alpha_1 Q_1 + \alpha_2 Q_2) \;=\; \alpha_1 P_t(Q_1) + \alpha_2 P_t(Q_2)
\end{equation}
\end{lemma}
\begin{proof}
Let $Q_1$ and $Q_2$ be arbitrary causal optimizers (not necessarily linear in $\vg$). Define the combined optimizer $Q = \alpha_1 Q_1 + \alpha_2 Q_2$ by its action on the gradient stream: $(Q\vg)_t \coloneqq \alpha_1 (Q_1\vg)_t + \alpha_2 (Q_2\vg)_t$. Then:
\begin{align*}
    P_t(\alpha_1 Q_1 + \alpha_2 Q_2)
    &\;=\; \E[\vg_t^\top \, (Q\vg)_t]\\
    &\;=\; \E[\vg_t^\top \, (\alpha_1 (Q_1\vg)_t + \alpha_2 (Q_2\vg)_t)]\\
    &\;=\; \E[\alpha_1 \vg_t^\top (Q_1\vg)_t + \alpha_2 \vg_t^\top (Q_2\vg)_t]\\
    &\;=\; \alpha_1 \E[\vg_t^\top \, (Q_1\vg)_t] + \alpha_2 \E[\vg_t^\top \, (Q_2\vg)_t]\\
    &\;=\; \alpha_1 P_t(Q_1) + \alpha_2 P_t(Q_2)
\end{align*}
The key observation is that linearity of $P_t$ in $Q$ follows from (i) the definition of linear combination of optimizers acting pointwise on the output parameter updates $\vu_t = (Q\vg)_t$, and (ii) linearity of expectation. This holds regardless of whether $Q_1$ or $Q_2$ are themselves linear operators on $\vg$.
\end{proof}

Lemma~\ref{lem:learning-power-functional-linear-in-Q} applies to general nonlinear adaptive optimizers.
For example, consider the Adam optimizer family:
The Adam update $Q^\text{Adam}_{\beta_1, \beta_2}: \vg_{\le t} \mapsto \vu_{t}$ is indeed a complex nonlinear functional of the gradient stream $\vg$.
The output $(Q^\text{Adam}_{\beta_1, \beta_2}\vg)_t = \vm_{\beta_1,t} / (\sqrt{v_{\beta_2,t} + \epsilon})$ depends nonlinearly on the entire gradient history through the exponential moving averages $\vm_{\beta_1,t}$ and $\vv_{\beta_2,t}$.
The lemma, however, concerns linearity in $Q$, not linearity of $Q$ in $\vg$.
That is, we can still apply the greedy alignment principle for the compact candidate family $\gQ_\gB^\text{Adam} = \{Q^\text{Adam}_{\beta_1, \beta_2} : (\beta_1, \beta_2) \in \gB\}$ of Adam optimizers to select the optimal hyperparameters $(\beta_1^\star, \beta_2^\star)$ that maximize the learning power $P_t(Q^\text{Adam}_{\beta_1, \beta_2})$ under the constraint $\gQ_\gB^\text{Adam}$.
Other nonlinear optimizers such as AdamW~\cite{loshchilov2017decoupled}, Lion~\cite{chen2023symbolic}, Muon~\cite{jordan2024muon}, Shampoo~\cite{gupta2018shampoo}, and clipping-based optimizers can be treated as black-box causal maps and selected by the same criterion.

The following theorems justify the generalization by extending Section~\ref{sec:2_prelim} to general nonlinear adaptive optimizers.

\begin{theorem}[Greedy optimal causal maps]
\label{thm:greedy-optimal-causal}
Let $\mathcal X$ be a real locally convex vector space of causal maps
$Q:\vg_{\le t}\mapsto \vu_t$, closed under pointwise linear combinations.
Fix the gradient stream $\vg_t$ at time $t$ and define
\begin{equation}
    \label{eq:thm:greedy-optimal-causal:learning-power}
    P_t(Q)
    \;\coloneqq\;
    \mathbb E[\vg_t^\top (Q\vg)_t].
\end{equation}
Assume $P_t\in \mathcal X^\ast$ is a continuous linear functional in the dual space $\mathcal X^\ast$.
Let $\mathcal Q\subset\mathcal X$ be nonempty and compact.
Define the support value
\begin{equation}
    \label{eq:thm:greedy-optimal-causal:support-value}
    \sigma_{\mathcal Q}(P_t)
    \;\coloneqq\;
    \max_{Q\in\mathcal Q}P_t(Q).
\end{equation}
Then:
\begin{enumerate}[label=(\roman*),leftmargin=2em]
\item \emph{(Existence):} The maximum is attained and finite.
\item \emph{(Subgradient characterization):} If $\mathcal Q$ is closed and convex, then
\begin{equation}
    \label{eq:thm:greedy-optimal-causal:subgradient}
    \partial \sigma_{\mathcal Q}(P_t)
    \;=\;
    \arg\max_{Q\in\mathcal Q}P_t(Q),
\end{equation}
where the subgradient is taken with respect to the dual pairing
between $\mathcal X^\ast$ and $\mathcal X$.
In particular, if the maximizer is unique, the support function has a unique
subgradient at $P_t$.
\item \emph{(Lipschitz continuity):} For any $P_t,\hat P_t\in\mathcal X^\ast$,
\begin{equation}
    \label{eq:thm:greedy-optimal-causal:lipschitz}
    |\sigma_{\mathcal Q}(P_t)-\sigma_{\mathcal Q}(\hat P_t)|
    \;\le\;
    \max\{
    \sigma_{\mathcal Q}(P_t-\hat P_t),
    \sigma_{\mathcal Q}(\hat P_t-P_t)
    \}.
\end{equation}
\end{enumerate}
\end{theorem}

\begin{proof}
\textit{(i) Existence:}
Since $\mathcal Q$ is compact and $P_t$ is continuous, the map $Q \mapsto P_t(Q)$ attains its maximum on $\mathcal Q$ by the Weierstrass theorem.
Finiteness follows from compactness and continuity.

\textit{(ii) Subgradient characterization:}
First, we show that $\sigma_{\mathcal Q}$ is sublinear.
For $a \ge 0$,
\begin{equation}
\label{eq:proof:greedy-optimal-causal:positive-homogeneity}
\sigma_{\mathcal Q}(a P_t) \;=\; \max_{Q \in \mathcal Q} a P_t(Q) \;=\; a \max_{Q \in \mathcal Q} P_t(Q) \;=\; a \sigma_{\mathcal Q}(P_t).
\end{equation}
For $P_t, \hat P_t \in \mathcal X^\ast$,
\begin{equation}
\label{eq:proof:greedy-optimal-causal:subadditivity}
\sigma_{\mathcal Q}(P_t + \hat P_t) \;=\; \max_{Q \in \mathcal Q} (P_t(Q) + \hat P_t(Q)) \;\le\; \max_{Q \in \mathcal Q} P_t(Q) + \max_{Q \in \mathcal Q} \hat P_t(Q).
\end{equation}
Thus $\sigma_{\mathcal Q}$ is positively homogeneous and subadditive, hence sublinear.

Now let $Q_t^\star \in \arg\max_{Q \in \mathcal Q} P_t(Q)$.
Then $\sigma_{\mathcal Q}(P_t) = P_t(Q_t^\star)$.
For any $h \in \mathcal X^\ast$,
\begin{equation}
\label{eq:proof:greedy-optimal-causal:subgradient-ineq}
\sigma_{\mathcal Q}(P_t + h) \;=\; \max_{Q \in \mathcal Q} (P_t(Q) + h(Q)) \;\ge\; P_t(Q_t^\star) + h(Q_t^\star) \;=\; \sigma_{\mathcal Q}(P_t) + h(Q_t^\star).
\end{equation}
This is exactly the subgradient inequality, thus $Q_t^\star \in \partial \sigma_{\mathcal Q}(P_t)$.

Conversely, suppose $\bar Q \in \partial \sigma_{\mathcal Q}(P_t)$, i.e.\ for all $h \in \mathcal X^\ast$,
\begin{equation}
\label{eq:proof:greedy-optimal-causal:subgradient-def}
\sigma_{\mathcal Q}(P_t + h) \;\ge\; \sigma_{\mathcal Q}(P_t) + h(\bar Q).
\end{equation}
We first show $\bar Q \in \mathcal Q$. For any $\tilde h \in \mathcal X^\ast$ and any $\tau > 0$, applying the inequality with $h = \tau \tilde h$ and combining with subadditivity and positive homogeneity of $\sigma_{\mathcal Q}$,
\begin{equation}
\label{eq:proof:greedy-optimal-causal:membership-bound}
\sigma_{\mathcal Q}(P_t) + \tau \sigma_{\mathcal Q}(\tilde h)
\;\ge\;
\sigma_{\mathcal Q}(P_t + \tau \tilde h)
\;\ge\;
\sigma_{\mathcal Q}(P_t) + \tau \tilde h(\bar Q).
\end{equation}
Dividing by $\tau$ gives $\sigma_{\mathcal Q}(\tilde h) \ge \tilde h(\bar Q)$ for all $\tilde h \in \mathcal X^\ast$. Since $\mathcal Q$ is closed and convex, the support-function characterization implies $\bar Q \in \mathcal Q$.

Next, applying the subgradient inequality with $h = -P_t$,
\begin{equation}
\label{eq:proof:greedy-optimal-causal:optimality}
0 \;=\; \sigma_{\mathcal Q}(0) \;\ge\; \sigma_{\mathcal Q}(P_t) - P_t(\bar Q),
\end{equation}
so $P_t(\bar Q) \ge \sigma_{\mathcal Q}(P_t)$.
Combined with $\bar Q \in \mathcal Q$, which gives $P_t(\bar Q) \le \sigma_{\mathcal Q}(P_t)$, we conclude $P_t(\bar Q) = \sigma_{\mathcal Q}(P_t)$, i.e., $\bar Q \in \arg\max_{Q \in \mathcal Q} P_t(Q)$.
Thus $\partial \sigma_{\mathcal Q}(P_t) = \arg\max_{Q \in \mathcal Q} P_t(Q)$.

\textit{(iii) Lipschitz continuity:}
Let $\delta = P_t - \hat P_t$. Then
\begin{equation}
\label{eq:proof:greedy-optimal-causal:lipschitz-upper}
\sigma_{\mathcal Q}(P_t) - \sigma_{\mathcal Q}(\hat P_t) \;=\; \sup_{Q \in \mathcal Q} P_t(Q) - \sup_{Q \in \mathcal Q} \hat P_t(Q) \;\le\; \sup_{Q \in \mathcal Q} (P_t(Q) - \hat P_t(Q)) \;=\; \sup_{Q \in \mathcal Q} \delta(Q).
\end{equation}
Swapping $P_t$ and $\hat P_t$,
\begin{equation}
\label{eq:proof:greedy-optimal-causal:lipschitz-lower}
\sigma_{\mathcal Q}(\hat P_t) - \sigma_{\mathcal Q}(P_t) \;\le\; \sup_{Q \in \mathcal Q} (-\delta(Q)).
\end{equation}
Combining both inequalities gives the stated bound.
\end{proof}

We get stronger uniqueness result if $\mathcal X$ is finite-dimensional (so that the number of hyperparameters and the number of delays are finite), or more generally if the standard regularity conditions for support functions $\sigma_{\mathcal Q}: \mathcal X^\ast \to \mathbb R$ on locally convex spaces holds.
In these cases, the support function is Gâteaux differentiable at $P_t$:
\begin{equation}
\label{eq:proof:greedy-optimal-causal:differentiability}
D \sigma_{\mathcal Q}(P_t)(h)
\;=\;
\lim_{\tau\to 0} \frac{\sigma_{\mathcal Q}(P_t + \tau h) - \sigma_{\mathcal Q}(P_t)}{\tau}
\;=\;
h(Q_t^\star).
\end{equation}
This is linear in $h$, so the maximizer $Q_t^\star$ is the unique subgradient of $\sigma_{\mathcal Q}$ at $P_t$.
We can then simplify the construction of the greedy optimal causal map to:
\begin{equation}
\label{eq:proof:greedy-optimal-causal:simplified-construction}
Q_t^\star = \nabla \sigma_{\mathcal Q}(P_t).
\end{equation}
This is coherent with Theorem~\ref{thm:dynamic-optimizer-under-convex-constraints} in Section~\ref{sec:2_prelim} for linear optimizers.

The next corollary extends the family-hull reduction theorem of Section~\ref{sec:2_prelim} to general causal maps.

\begin{corollary}[Family-hull reduction for arbitrary causal maps]
\label{cor:family-hull-causal}
Let $\mathcal F\subset\mathcal X$ be any nonempty set of causal maps, and let $P_t\in\mathcal X^\ast$ be the learning-power functional.
Then
\begin{equation}
    \label{eq:cor:family-hull-causal:hull-reduction}
    \sup_{Q\in\operatorname{co}(\mathcal F)}P_t(Q)
    \;=\;
    \sup_{Q\in\mathcal F}P_t(Q).
\end{equation}
If $\mathcal F$ is compact and $P_t$ is continuous, the supremum on the right-hand side is attained.
Moreover, if a convex combination of elements of $\mathcal F$
\begin{equation}
    \label{eq:cor:family-hull-causal:convex-combination}
    Q^\star\;=\;\sum_{i=1}^m \alpha_i Q_i,
    \qquad \text{where} \quad
    Q_i\in\mathcal F,\;
    \alpha_i>0,\;
    \sum_i\alpha_i=1,
\end{equation}
attains the supremum over $\operatorname{co}(\mathcal F)$, then every active
component $Q_i$ with $\alpha_i>0$ also attains the same maximal score:
\begin{equation}
    \label{eq:cor:family-hull-causal:active-component}
    P_t(Q_i)\;=\;\sup_{Q\in\mathcal F}P_t(Q).
\end{equation}
\end{corollary}

\begin{proof}
Since $\mathcal F \subseteq \operatorname{co}(\mathcal F)$,
\begin{equation}
\label{eq:proof:family-hull-causal:lower-bound}
\sup_{Q \in \operatorname{co}(\mathcal F)} P_t(Q) \;\ge\; \sup_{Q \in \mathcal F} P_t(Q).
\end{equation}
For the reverse inequality, let $Q \in \operatorname{co}(\mathcal F)$. Then $Q = \sum_{i=1}^m \alpha_i Q_i$ for some $Q_i \in \mathcal F$, $\alpha_i \ge 0$, and $\sum_i \alpha_i = 1$. By linearity of $P_t$ (Lemma~\ref{lem:learning-power-functional-linear-in-Q}),
\begin{equation}
\label{eq:proof:family-hull-causal:linearity}
P_t(Q) \;=\; \sum_i \alpha_i P_t(Q_i) \;\le\; \sum_i \alpha_i \sup_{Q' \in \mathcal F} P_t(Q') \;=\; \sup_{Q' \in \mathcal F} P_t(Q').
\end{equation}
Taking the supremum over $Q \in \operatorname{co}(\mathcal F)$ gives
\begin{equation}
\label{eq:proof:family-hull-causal:upper-bound}
\sup_{Q \in \operatorname{co}(\mathcal F)} P_t(Q) \;\le\; \sup_{Q \in \mathcal F} P_t(Q).
\end{equation}
Thus equality holds.

If $\mathcal F$ is compact and $P_t$ is continuous, then $P_t$ attains its maximum on $\mathcal F$.

Finally, suppose a convex combination of active components $Q^\star = \sum_{i=1}^m \alpha_i Q_i$ with $\alpha_i > 0$ and $\sum_i \alpha_i = 1$ attains the supremum.
Let $P^\star = \sup_{Q \in \mathcal F} P_t(Q)$.
Then $P_t(Q_i) \le P^\star$ for all $i$, and by linearity of $P_t$ (Lemma~\ref{lem:learning-power-functional-linear-in-Q}),
\begin{equation}
\label{eq:proof:family-hull-causal:weighted-average}
P^\star \;=\; P_t(Q^\star) \;=\; \sum_i \alpha_i P_t(Q_i) \;\le\; \sum_i \alpha_i P^\star \;=\; P^\star.
\end{equation}
Since all $\alpha_i > 0$, the only way the weighted average equals $P^\star$ is if $P_t(Q_i) = P^\star$ for every active $i$.
Thus every $Q_i$ also attains the same maximal score $P^\star$.
\end{proof}

The corollary implies that the greedy alignment principle can be applied to optimize over families containing general nonlinear adaptive optimizers.
If the set of causal maps $\mathcal X$ is convex, greedy optimizer selection is a convex optimization problem even though each individual optimizer $Q \in \mathcal X$ may be highly nonlinear in the gradient stream $\vg$.
For a nonconvex hyperparameterized family, the family-hull reduction in Corollary~\ref{cor:family-hull-causal} shows that optimizing over its convex hull gives the same value as selecting the best original candidate.
Therefore, our finite $K$-Switch selection algorithm is exact for the convexified family.

\section{Closed-form solutions for five canonical families of optimizers}
\label{sec:appx:families}
This section solves the optimization problem (\ref{eq:sec_2:optimization_problem}) using Theorem~\ref{thm:dynamic-optimizer-under-convex-constraints}.
Exact closed form solutions for the greedy-optimal optimizer $Q^\star$ and the corresponding optimal learning power $P^\star(R)$ are obtained for the prescribed family $\gQ$ and gradient autocorrelation sequence $R$.
This theoretical study reveals how widely-used families of optimizers emerge as special cases of the general optimization problem.
Specifically, we classify the greedy-optimal optimizers $Q^\star$ into five canonical families of causal filters $\gQ$.
As we see in the following, the prescribed family $\gQ$ characterizes the solution optimizer $Q^\star$ and its associated hyperparameters.

We begin by observing that most common practical optimizers are symmetric.
For example, every component-wise updater has a diagonal filter response $Q$, hence is symmetric.
For time lag $k>0$, consider the gradient autocorrelation
\begin{equation}
    \label{eq:appx:autocorrelation}
    R_{t,k} \;=\; \E[\vg_t \, \vg_{t-k}^\top].
\end{equation}
This quantity is not generally symmetric nor PSD.
However, for symmetric or PSD filter families, only its symmetric part contributes to the actual learning power.
Therefore, we focus on symmetric filters by first defining the symmetric part of the autocorrelation $R$:
\begin{equation}
    \label{eq:appx:symmetric-part}
    S_{t,k} \;\coloneqq\; \frac{R_{t,k}+R_{t,k}^\top}{2}.
\end{equation}
$S_{t,k}$ is a symmetric matrix for each time $t$ and delay $k$; it is PSD at $k=0$ (since $S_{t,0}=R_{t,0}=\E[\vg_t\vg_t^\top]$), but for $k>0$ it may be indefinite, motivating the use of its positive part $S_{t,k}^+$ defined below.
Whenever the filter $Q_{t,k}$ is symmetric, i.e., $Q_{t,k}=Q_{t,k}^\top$, the trace of the inner products are equal, i.e.,
\begin{equation}
\label{eq:appx:trace-symmetric}
\operatorname{Tr}(Q_{t,k}^\top R_{t,k})
\;=\;
\operatorname{Tr}(Q_{t,k}S_{t,k}).
\end{equation}
Let
\begin{equation}
\label{eq:appx:symmetric-eigendecomp}
S_{t,k}=U_{t,k}\operatorname{diag}(\lambda_{1,t,k}\ge\cdots\ge\lambda_{d,t,k})\,U_{t,k}^\top
\end{equation}
and denote the positive part by
\begin{equation}
\label{eq:appx:positive-part}
S_{t,k}^+=U_{t,k}\operatorname{diag}([\lambda_{i,t,k}]_+)\,U_{t,k}^\top .
\end{equation}
Consider the following five types of causal filter families:
\begin{itemize}[leftmargin=2em,topsep=0pt]
\item \emph{Hilbert ball type $\gQ_\textnormal{H}(B) = \{Q : \|Q\|_{\gH}^2 = \sum_{k=0}^{\infty} \operatorname{Tr}(Q_{t,k}^\top Q_{t,k}) \leq B\}$} is the simplest and the largest family that does not favor any particular direction in the parameter space, but requires larger memory to store its hyperparameters.
\item \emph{Spectral type $\gQ_\textnormal{S}(\tau,\Lambda) = \{Q \succeq 0: \operatorname{Tr}(Q_{t,k}) \leq \tau_k, \,0 \preceq Q_{t,k} \preceq \Lambda_k I\, \forall k\}$} is a trust region that upper limits the (1) per-direction spectrum for safety and the (2) trace for total update budget, simultaneously at each time delay $k$.
\item \emph{Lyapunov type $\gQ_\textnormal{L}(B) = \{Q \succeq 0: Q_{t,k} = \Pi_{t,k}^+ Q_{t,k} \Pi_{t,k}^+,\, \operatorname{Tr}(Q_{t,k}^\top S_{t,k}^+ Q_{t,k}) \leq B_k\, \forall k\}$}, where $\Pi_{t,k}^+$ is the projection onto $\operatorname{range}(S_{t,k}^+)$, utilizes the positive symmetric part of the lag-correlation sequence itself as the metric, leading to a natural dynamic Lyapunov-like stability condition.
\item \emph{Metric type $\gQ_\textnormal{M}(B, M) = \{Q \succeq 0: \operatorname{Tr}(Q_{t,k}^\top M_{t,k} Q_{t,k}) \leq B_k, M_k \succ 0\, \forall k\}$} is a general preconditioning optimizer that utilizes a general positive-definite metric $M_k$ to control the learning power.
\item \emph{Diagonal type $\gQ_\textnormal{D}(B,c) = \{Q \succeq 0: Q_{t,k} = \operatorname{diag}(q_{j,k}) \succeq 0, \, \sum_j c_{j,k} q_{j,k}^2 \leq B_k\, \forall k\}$} represents element-wise optimizers, a memory-efficient family that are commonly used in large-scale machine learning.
\end{itemize}
Instantiating the construction from Theorem~\ref{thm:dynamic-optimizer-under-convex-constraints} on each of these families, we obtain the closed-form greedy-optimal optimizer $Q^\star$ and the corresponding optimal learning power $P^\star(S)$.

\begin{theorem}[Closed-form solutions for causal filter families]
\label{thm:closed-form-solutions-dynamic}
Omit the time index $t$ for brevity.
Assume throughout that all displayed series converge absolutely.
Note that the Hilbert family uses the full lag moment $R$.
The remaining four families depend only on the symmetric autocorrelation $S_k=\operatorname{Sym}(R_k)$ and their positive parts $S_k^+$.
The closed-form optimal solutions corresponding to each of the five causal filter families are as follows:
\begin{enumerate}[leftmargin=2em,topsep=0pt,label=(\roman*)]
\item \emph{(Hilbert ball):}
The optimal $Q_\textnormal{H}^\star$ is proportional to the autocorrelation $R$, i.e.,
\begin{equation}
Q_\textnormal{H}^\star \;=\; \frac{\sqrt{B}}{\|R\|_{\gH}} R,
\end{equation}
if $R \neq 0$; otherwise any feasible $Q$ is optimal.
This gives the optimal learning power
\begin{equation}
P_\textnormal{H}^\star(R) \;=\; \sqrt{B}\, \|R\|_{\gH}.
\end{equation}

\item \emph{(Spectral):}
The optimal $Q^\star_{\textnormal{S},k}$ shares the same eigenstructure as $S_k$ but with eigenvalues $q_{i,k}^\star$ instead of $\lambda_{i,k}$:
\begin{equation}
Q^\star_{\textnormal{S},k} \;=\; U_k \operatorname{diag}(q_{i,k}^\star)\, U_k^\top.
\end{equation}
The eigenvalues $q_{i,k}^\star$ are determined by \emph{water-filling}:
Let $d_k^+$ be the number of positive eigenvalues of $S_k$ and $m_k = \min\{d_k^+, \lfloor \tau_k / \Lambda_k \rfloor\}$.
The eigenvalues $q_{i,k}^\star$ are then given by:
\begin{equation}
q_{i,k}^\star \;=\; \begin{cases}
\Lambda_k, & \text{if } i \leq m_k, \\
\tau_k - m_k \Lambda_k, & \text{if } i = m_k + 1 \le d_k^+ \text{ and } \tau_k < d_k^+\Lambda_k, \\
0, & \text{otherwise}.
\end{cases}
\end{equation}
This gives the optimal learning power
\begin{equation}
P_\textnormal{S}^\star(S) \;=\; \sum_{k=0}^{\infty} \left[\Lambda_k \sum_{i = 1}^{m_k} \lambda_{i,k} + \mathbf{1}_{m_k<d_k^+}(\tau_k - m_k \Lambda_k)\, \lambda_{m_k + 1, k}\right],
\end{equation}
with the convention $\lambda_{d_k^+ + 1,k} = 0$, so the second term vanishes when $\tau_k \ge d_k^+\Lambda_k$ (i.e., $m_k = d_k^+$).

\item \emph{(Lyapunov):}
For each delay $k$, define $\Pi_k^+$ as the orthogonal projection onto $\operatorname{range}(S_k^+)$.
Then, the optimal $Q^\star_{\textnormal{L},k}$ is given by
\begin{equation}
Q^\star_{\textnormal{L},k} \;=\; \alpha_k \Pi_k^+, \qquad
\alpha_k \;=\; \sqrt{\frac{B_k}{\operatorname{Tr}(S_k^+)}},
\end{equation}
if $S_k^+ \neq 0$; otherwise $Q^\star_{\textnormal{L},k} = 0$.
This gives the optimal learning power
\begin{equation}
P_\textnormal{L}^\star(S) \;=\; \sum_{k=0}^{\infty} \sqrt{B_k \operatorname{Tr}(S_k^+)}.
\end{equation}

\item \emph{(Metric, commuting case):}
For each delay $k$, assume that $M_k \succ 0$ and $S_k$ commute, so that they are simultaneously diagonalizable with eigenvalues $m_{i,k} > 0$ and $\lambda_{i,k}$, respectively.
Let $U_k$ denote their common eigenbasis.
Then the optimal $Q^\star_{\textnormal{M},k}$ is given by
\begin{equation}
Q^\star_{\textnormal{M},k} \;=\; \alpha_k U_k \operatorname{diag}\left(\frac{[\lambda_{i,k}]_+}{m_{i,k}}\right) U_k^\top, \qquad
\alpha_k \;=\; \sqrt{\frac{B_k}{\sum_i [\lambda_{i,k}]_+^2 / m_{i,k}}},
\end{equation}
if $S_k^+ \neq 0$; otherwise $Q^\star_{\textnormal{M},k} = 0$.
This gives the optimal learning power
\begin{equation}
P_\textnormal{M}^\star(S) \;=\; \sum_{k=0}^{\infty} \sqrt{B_k \sum_i [\lambda_{i,k}]_+^2 / m_{i,k}}.
\end{equation}
Equivalently, under this commutativity assumption,
\begin{equation}
Q^\star_{\textnormal{M},k}=\alpha_k M_k^{-1}S_k^+.
\end{equation}

\item \emph{(Diagonal):}
For each delay $k$, let $s_{j,k} \coloneqq (S_k)_{jj}$ denote the diagonal entries of $S_k$.
The optimal $Q^\star_{\textnormal{D},k}$ is a diagonal matrix with elements given by
\begin{equation}
Q^\star_{\textnormal{D},k} \;=\; \operatorname{diag}(q_{j,k}^\star), \qquad
q_{j,k}^\star \;=\; \alpha_k \frac{[s_{j,k}]_+}{c_{j,k}}, \qquad
\alpha_k \;=\; \sqrt{\frac{B_k}{\sum_\ell [s_{\ell,k}]_+^2 / c_{\ell,k}}},
\end{equation}
if $\sum_\ell [s_{\ell,k}]_+ > 0$; otherwise $Q^\star_{\textnormal{D},k} = 0$.
This gives the optimal learning power
\begin{equation}
P_\textnormal{D}^\star(S) \;=\; \sum_{k=0}^{\infty} \sqrt{B_k \sum_j [s_{j,k}]_+^2 / c_{j,k}}.
\end{equation}
\end{enumerate}

\begin{proof}
We apply Theorem~\ref{thm:dynamic-optimizer-under-convex-constraints} to each causal filter family.

\emph{(i) Hilbert ball $\gQ_\textnormal{H}(B) = \{Q : \|Q\|_{\gH} \leq \sqrt{B}\}$.}
The Lagrangian is $L(Q,\lambda) = \langle Q,R\rangle_{\gH} - \lambda(\|Q\|_{\gH}^2 - B)$.
Taking the gradient with respect to $Q$ and setting to zero gives
\begin{equation}
\nabla_Q L \;=\; R - 2\lambda Q \;=\; 0 \quad \Rightarrow \quad Q \;=\; \frac{R}{2\lambda}.
\end{equation}
The constraint $\|Q\|_{\gH} = \sqrt{B}$ gives $\|R/(2\lambda)\|_{\gH} = \sqrt{B}$, so $2\lambda = \|R\|_{\gH}/\sqrt{B}$.
Hence
\begin{equation}
Q_\textnormal{H}^\star \;=\; \sqrt{B}\,\frac{R}{\|R\|_{\gH}}, \qquad
P_\textnormal{H}^\star(R) \;=\; \langle Q_\textnormal{H}^\star,R\rangle_{\gH} \;=\; \sqrt{B}\,\|R\|_{\gH}.
\end{equation}

\emph{(ii) Spectral $\gQ_\textnormal{S}(\tau,\Lambda) = \{Q : \operatorname{Tr}(Q_{t,k}) \leq \tau_k,\ 0 \preceq Q_{t,k} \preceq \Lambda_k I\, \forall k\}$.}
The problem decouples over delays $k$. For each $k$, by von Neumann's trace inequality, the maximizer shares the eigenstructure of $S_k$:
\begin{equation}
Q_k^\star = U_k \operatorname{diag}(q_{i,k}^\star)\, U_k^\top,
\end{equation}
where the eigenvalues $q_{i,k}^\star$ solve the linear program
\begin{equation}
\max_{0 \leq q_{i,k} \leq \Lambda_k} \sum_i q_{i,k}\, \lambda_{i,k} \quad \text{s.t.} \quad \sum_i q_{i,k} \le \tau_k.
\end{equation}
Let $d_k^+$ be the number of positive eigenvalues of $S_k$ and $m_k = \min\{d_k^+, \lfloor \tau_k / \Lambda_k \rfloor\}$.
The optimal solution allocates maximum weight $\Lambda_k$ to the largest eigenvalues $\lambda_{i,k}$ until the trace budget $\tau_k$ is exhausted, giving the water-filling formula:
\begin{equation}
q_{i,k}^\star \;=\; \begin{cases}
\Lambda_k, & \text{if } i \leq m_k, \\
\tau_k - m_k \Lambda_k, & \text{if } i = m_k + 1 \le d_k^+ \text{ and } \tau_k < d_k^+\Lambda_k, \\
0, & \text{otherwise}.
\end{cases}
\end{equation}

\emph{(iii) Lyapunov $\gQ_\textnormal{L}(B) = \{Q : Q_{t,k} \succeq 0,\ Q_{t,k} = \Pi_{t,k}^+ Q_{t,k} \Pi_{t,k}^+,\, \operatorname{Tr}(Q_{t,k}^\top S_{t,k}^+ Q_{t,k}) \leq B_k\, \forall k\}$.}
The problem decouples over delays $k$. For each $k$, the Lagrangian on the support of $S_k^+$ is
\begin{equation}
L_k(Q_k, \mu_k) \;=\; \operatorname{Tr}(Q_k^\top S_k^+) - \mu_k \bigl(\operatorname{Tr}(Q_k^\top S_k^+ Q_k) - B_k\bigr).
\end{equation}
The first-order condition gives
\begin{equation}
S_k^+ - 2 \mu_k S_k^+ Q_k \;=\; 0 \quad \Rightarrow \quad S_k^+ (I - 2 \mu_k Q_k) \;=\; 0.
\end{equation}
This implies $Q_k = \frac{1}{2 \mu_k} I$ on the support of $S_k^+$, i.e., $Q_k = \alpha_k \Pi_k^+$ where $\alpha_k = \frac{1}{2 \mu_k}$ and $\Pi_k^+$ is the projection onto $\operatorname{range}(S_k^+)$.
Since $L_k$ is concave in $Q_k$ for $\mu_k > 0$ (as $S_k^+ \succeq 0$), the first-order condition gives the global maximum.
Using the constraint:
\begin{equation}
\operatorname{Tr}(Q_k^\top S_k^+ Q_k)
\;=\; \alpha_k^2 \operatorname{Tr}(S_k^+)
\;=\; \alpha_k^2 \sum_{i} [\lambda_{i,k}]_+ \;=\; B_k.
\end{equation}
Therefore, $\alpha_k = \sqrt{B_k / \sum_{i} [\lambda_{i,k}]_+}$ if $\sum_i [\lambda_{i,k}]_+ > 0$; otherwise $Q^\star_{\textnormal{L},k} = 0$.

\emph{(iv) Metric $\gQ_\textnormal{M}(B, M) = \{Q : Q_{t,k} \succeq 0,\ \operatorname{Tr}(Q_{t,k}^\top M_{t,k} Q_{t,k}) \leq B_k,\ M_{t,k} \succ 0\, \forall k\}$.}
The problem decouples over delays $k$. For each $k$, assume $M_k$ and $S_k$ commute with common eigenbasis $U_k$ and eigenvalues $m_{i,k} > 0$ and $\lambda_{i,k}$ respectively. The Lagrangian is
\begin{equation}
L_k(Q_k, \mu_k) \;=\; \operatorname{Tr}(Q_k^\top S_k) - \mu_k \bigl(\operatorname{Tr}(Q_k^\top M_k Q_k) - B_k\bigr).
\end{equation}
Since $Q_k \succeq 0$, only positive eigenvalues $[\lambda_{i,k}]_+$ contribute to the objective. The first-order condition in the common eigenbasis gives
\begin{equation}
[\lambda_{i,k}]_+ - 2 \mu_k m_{i,k} q_{i,k} \;=\; 0 \quad \Rightarrow \quad q_{i,k} \;=\; \frac{[\lambda_{i,k}]_+}{2 \mu_k m_{i,k}}.
\end{equation}
Using the constraint $\sum_i m_{i,k} q_{i,k}^2 = B_k$:
\begin{equation}
\sum_i m_{i,k} \left(\frac{[\lambda_{i,k}]_+}{2 \mu_k m_{i,k}}\right)^2
\;=\; \frac{1}{4\mu_k^2} \sum_i \frac{[\lambda_{i,k}]_+^2}{m_{i,k}} \;=\; B_k.
\end{equation}
Therefore, $q_{i,k}^\star = \alpha_k [\lambda_{i,k}]_+ / m_{i,k}$ where $\alpha_k = \sqrt{B_k / \sum_i [\lambda_{i,k}]_+^2 / m_{i,k}}$ if $\sum_i [\lambda_{i,k}]_+ > 0$; otherwise $Q^\star_{\textnormal{M},k} = 0$.

\emph{(v) Diagonal $\gQ_\textnormal{D}(B,c) = \{Q : Q_{t,k} = \operatorname{diag}(q_{j,k}),\ q_{j,k} \geq 0,\ \sum_j c_{j,k} q_{j,k}^2 \leq B_k\, \forall k\}$.}
The problem decouples over delays $k$. Since $Q_k$ is diagonal, $\langle Q_k, S_k\rangle$ depends on $S_k$ only through its diagonal entries; write $s_{j,k} \coloneqq (S_k)_{jj}$. Since $q_{j,k} \geq 0$, only positive diagonal entries $[s_{j,k}]_+$ contribute to the objective. For each $k$, we solve:
\begin{equation}
\max_{q_{j,k} \geq 0} \sum_j [s_{j,k}]_+\, q_{j,k} \quad \text{s.t.} \quad \sum_j c_{j,k} q_{j,k}^2 \leq B_k.
\end{equation}
By Cauchy--Schwarz:
\begin{equation}
\sum_j [s_{j,k}]_+ q_{j,k} \;=\; \sum_j (\sqrt{c_{j,k}} q_{j,k}) \frac{[s_{j,k}]_+}{\sqrt{c_{j,k}}} \;\leq\; \left(\sum_j c_{j,k} q_{j,k}^2\right)^{1/2} \left(\sum_j \frac{[s_{j,k}]_+^2}{c_{j,k}}\right)^{1/2}.
\end{equation}
Equality holds when $q_{j,k}^\star \propto [s_{j,k}]_+ / c_{j,k}$.
Normalizing by the constraint gives the stated result.
\end{proof}
\end{theorem}

These analytic solutions reveal how the characteristics of different types of optimal dynamic optimizers $Q^\star$ are induced by controlling the feasible set $\mathcal{Q}$.

\paragraph{Hilbert family $\leftrightarrow$ Matched causal filter.} Budget $\gQ_\textnormal{H}(B)$ produces a \emph{matched filter} that aligns learning power with lag-correlation evidence $Q^\star \propto R$.
It enjoys implementation simplicity but potentially over-concentrates on dominant temporal modes.
As in Theorem~\ref{thm:greedy-optimal-sgdm}, we can project this general class of optimizers into special geometries to calculate for the optimal hyperparameters.
Momentum-based family is one canonical example of constraining this family to a one-pole filter, i.e., momentum is a one-pole projection of the matched gradient-statistics filter.
This idea aligns with Corollary~\ref{cor:ar1-momentum} in the main manuscript.

\paragraph{Per-lag spectral family $\leftrightarrow$ Water-filling over positive active-power modes.} Budget $\gQ_\textnormal{S}(\tau,\Lambda)$ produces a \emph{water-filling} dynamic optimizer that allocates budget to positive active-power modes until hitting the safety cap $\Lambda_k$.
The spectral cap $\Lambda_k$ acts as a \emph{per-mode safety mechanism} analogous to \emph{gradient clipping}, while the trace constraint $\tau_k$ controls the \emph{per-lag learning-power budget} analogous to \emph{learning rate scheduling}.
This suggests that clipping and learning rate scheduling can be viewed as consequences of spectral constraints in the greedy alignment problem.
Furthermore, we also observe a spectral analogue of \emph{cautious alignment}~\cite{liang2026cautious}: negative active-power modes receive no budget under PSD-constrained filter families.

\paragraph{Lyapunov family $\leftrightarrow$ Equal-power dynamic optimizer.} Budget $\gQ_\textnormal{L}(B)$ produces an \emph{equal-power} dynamic optimizer that equalizes learning power uniformly across informative (positive) lag-correlation directions, preventing over-concentration while maintaining temporal efficiency.
This produces a natural dynamic preconditioning effect, with uniform power allocation across all informative temporal directions.

\paragraph{Metric family $\leftrightarrow$ General preconditioning optimizer.} Budget $\gQ_\textnormal{M}(B, M)$ produces a \emph{general preconditioning} optimizer that adapts the learning rate statically or dynamically based on the metric $M_k$.
This provides an abstract template for many existing preconditioning methods, such as K-FAC~\citep{martens2015optimizing}, Shampoo~\citep{gupta2018shampoo}, and natural gradient descent~\citep{amari1998natural}.

\paragraph{Diagonal family $\leftrightarrow$ Coordinate-wise dynamic optimizer $\sim$ Adam.} Budget $\gQ_\textnormal{D}(B,c)$ produces a \emph{coordinate-wise} dynamic optimizer that adapts per-coordinate learning power proportional to the positive diagonal evidence $[s_{j,k}]_+$ and inversely proportional to the costs $c_j$.
When $c_j \propto \sqrt{v_{\beta_2,t,j} + \epsilon}$ ($v_{\beta_2,t,j}$ being the EMA of $g_j^2$ at time $t$), this recovers the core mechanism of \emph{Adam}: coordinate update is proportional to evidence divided by cost.
This is elaborated in Corollary~\ref{cor:greedy-optimal-adam} in the main manuscript.

These closed-form solutions show that the optimizer family $\gQ$ is not an arbitrary constraint: it is the inductive bias that determines how learning power, i.e., the expected loss drop rate $P^\star(R) = -\E [\mathrm d \gL / \mathrm d t]$, is allocated across temporal and parameter-space modes.
Different feasible geometries allocate the same learning power differently, reflecting the different inductive biases of each family.
The Hilbert family yields a matched filter, the spectral family yields water-filling under safety caps, the Lyapunov family equalizes power on informative modes, and the diagonal family recovers Adam-like cost-weighted coordinate adaptation.

\section{Experimental details}
\label{sec:appx:exp}
This section provides implementation details and full results of the experiments conducted to validate the theory in Section~\ref{sec:3_main}.
First, we provide the hyperparameters and settings for the experiments in Tables~\ref{tab:exp_hyperparams_metamath_gemma}, \ref{tab:exp_hyperparams_metamath_llama}, \ref{tab:exp_hyperparams_commonsense_gemma}, \ref{tab:exp_hyperparams_commonsense_gemma_full}, and \ref{tab:exp_hyperparams_vit} of Appendix~\ref{sec:appx:exp:implementation}.
Additional experimental results are provided in Tables~\ref{tab:sgd_momentum_results_full}, \ref{tab:adam_results_full}, \ref{tab:gemma2b_metamath_lora_base_full}, and \ref{tab:llm_commonsense_full} of Appendix~\ref{sec:appx:exp:results}.
Our automatic hyperparameter tuning shows comparable performance across all datasets and models, including conventional residual networks~\citep{he2016deep}, vision transformers~\citep{dosovitskiy2021image}, and modern large language models~\citep{gemma2024open, llama32024herd} with or without using parameter-efficient fine-tuning methods like low-rank adaptation (LoRA)~\citep{hu2022lora}.
This demonstrates the practical usefulness of our framework.

\subsection{Implementation details}
\label{sec:appx:exp:implementation}

\begin{table}[t]
\centering
\caption{\small Hyperparameters and settings for math finetuning experiments on Gemma-2B~\citep{gemma2024open} with LoRA~\citep{hu2022lora}. Values reflect the experimental script.}
\label{tab:exp_hyperparams_metamath_gemma}
\renewcommand{\arraystretch}{1.1}
\resizebox{.8\textwidth}{!}{%
\begin{tabular}{ll}
\toprule
\textbf{Parameter}        & \textbf{Value(s) / Description} \\
\midrule
Dataset                   & MetaMathQA-395K \\
Training subset size      & 100{,}000 \\
Models tested             & Gemma-2B (\texttt{google/gemma-2b}) \\
Hardware                  & 1 $\times$ A100-80GB \\
Precision                 & Bfloat16 (BF16) \\
Optimizer                 & AdamW~\citep{kingma2015adam,loshchilov2017decoupled} \\
Optimal AdamW-type optimizer & \emph{two-option switch} with $\beta_1$ endpoints of $0.8$ and $0.99$ \\
Epochs                    & $1$ \\
Batch size ($bs$)         & $32$ \\
Learning rate ($lr$)      & $2 \times 10^{-4}$ \\
Weight decay              & $0$ \\
Warmup ratio              & $0$ \\
Adapter type              & LoRA~\citep{hu2022lora} \\
LoRA Rank ($r$)           & $32$ \\
LoRA Scaling ($\alpha$)   & $4$ \\
LoRA Dropout              & $0.05$ \\
Cutoff length             & $256$ \\
Adapter target modules    & \begin{tabular}[t]{@{}l@{}}%
                             \texttt{q\_proj, k\_proj, v\_proj, o\_proj} \\
                             \texttt{down\_proj, up\_proj, gate\_proj}%
                           \end{tabular} \\
\bottomrule
\end{tabular}%
}
\end{table}

% \subsection{Additional concerns}
% \label{sec:appx:exp:concerns}

\paragraph{Increasing the effective averaging horizon with EMA.}
During optimization, the gradients $\vg$ and the parameter updates $\vu$ vary over time, and the exact expectation $\E$ in the argmax objective $A_{k,t}$ is unknown.
In the main implementation, we use the instantaneous mini-batch score $\vg^\top \vu$ to approximate the expected alignment.
The effective averaging horizon can be further increased without additional computational overhead by maintaining an EMA of the argmax objective $A_{k,t}$.
We use an EMA with decay factor $\alpha = 0.9$ for Adam/AdamW experiments in the main manuscript.
Algorithm~\ref{alg:optimal_adam} omits the EMA implementation for brevity.

\paragraph{State decay under persistent negative alignment.}
Section~\ref{sec:3_main:finite-step-descent-bridge} shows that the greedy alignment score $A = \vg^\top \vu$ is a faithful surrogate for the actual one-step drop rate in loss within a local regime where $\eta \|\vu\|$ is small.
Outside this regime, negative alignments $A < 0$ can occur~\citep{liang2026cautious,chang2025mgup} when component-wise opposition dominates component-wise alignment:
$A = \sum_j [g_j u_j]_+ - \sum_j[-g_j u_j]_+$ and $|\sum_j[-g_j u_j]_+| > |\sum_j [g_j u_j]_+|$.
We thus treat a negative alignment as an indicator of low confidence in the accumulated optimizer states, rather than as a proof that every coordinate is harmful, and we apply a soft attenuation by halving the optimizer states, e.g., the running momentum, whenever a negative alignment $A<0$ persists for $5$ consecutive steps.

\paragraph{ResNet-18 on CIFAR-100.}
For ResNet-18~\citep{he2016deep} on CIFAR-100~\citep{krizhevsky2009learning} experiments, we follow the standard settings of~\cite{he2016deep}: 300 epochs with a learning rate decay of 0.1 at epochs 60, 120, and 160.
All hyperparameters other than momentum are held fixed.
We use a weight decay of $5\times10^{-4}$, batch size of $128$, and a base learning rate of $0.1$ for SGD and $0.01$ for Adam~\citep{kingma2015adam}.
For our optimal Adam-type optimizers using the \emph{two-option switch}, we use $\beta_1$ options of $0.8$ and $0.999$, to ensure that these endpoints enclose the typical range of $\beta_1$ values used in practice.
For our optimal SGD+Momentum-type optimizers using the \emph{two-option switch}, we use momentum options of $0.8$ and $0.99$, again to ensure that the endpoints enclose the typical range of momentum values in practice.
For optimal SGD+Momentum-type optimizers using the \emph{five-option switch}, we use momentum options of $0.9$, $0.95$, $0.98$, $0.99$, and $0.995$, to demonstrate the effectiveness of fine-grained control in dynamic hyperparameter tuning.
In the main manuscript, we show only the results of the \emph{two-option switch} for SGD+Momentum and Adam, which incurs lower overhead than the five-option variant.
Our \emph{five-option switch} for SGD+Momentum demonstrates that we can achieve \emph{significantly better performance} than the baseline optimizer with fixed hyperparameters ($77.57\% \rightarrow 78.33\%$ test accuracy).
These extended results are summarized in Tables~\ref{tab:sgd_momentum_results_full} and~\ref{tab:adam_results_full} in the next section.
However, current implementation of multi-option switch larger than two requires a significant amount of computation time (around +100\% compared to the baseline) and memory usage.
Therefore, we did not include the results in the main manuscript, opening up a future direction for more efficient implementation.

\begin{table}[t]
\centering
\caption{\small Hyperparameters and settings for math finetuning experiments on Llama-3-8B~\citep{llama32024herd} with LoRA~\citep{hu2022lora}. Values reflect the experimental script.}
\label{tab:exp_hyperparams_metamath_llama}
\renewcommand{\arraystretch}{1.1}
\resizebox{.8\textwidth}{!}{%
\begin{tabular}{ll}
\toprule
\textbf{Parameter}        & \textbf{Value(s) / Description} \\
\midrule
Dataset                   & MetaMathQA-395K \\
Training subset size      & 100{,}000 \\
Models tested             & Llama-3-8B (\texttt{meta-llama/llama-3-8b}) \\
Hardware                  & 1 $\times$ A100-80GB \\
Precision                 & Bfloat16 (BF16) \\
Optimizer                 & AdamW~\citep{kingma2015adam,loshchilov2017decoupled} \\
Optimal AdamW-type optimizer & \emph{two-option switch} with $\beta_1$ endpoints of $0.8$ and $0.99$ \\
Epochs                    & $1$ \\
Batch size ($bs$)         & $32$ \\
Learning rate ($lr$)      & $1 \times 10^{-4}$ \\
Weight decay              & $0$ \\
Warmup ratio              & $0$ \\
Adapter type              & LoRA~\citep{hu2022lora} \\
LoRA Rank ($r$)           & $32$ \\
LoRA Scaling ($\alpha$)   & $4$ \\
LoRA Dropout              & $0.05$ \\
Cutoff length             & $256$ \\
Adapter target modules    & \begin{tabular}[t]{@{}l@{}}%
                             \texttt{q\_proj, k\_proj, v\_proj, o\_proj} \\
                             \texttt{down\_proj, up\_proj, gate\_proj}%
                           \end{tabular} \\
\bottomrule
\end{tabular}%
}
% \vspace{-0.5em}
\end{table}

\paragraph{Gemma-2B and Llama-3-8B on MetaMathQA-395K.}
We summarize the hyperparameters and training settings for these experiments that produced Table~\ref{tab:gemma_mathqa} and Table~\ref{tab:llama_mathqa} in Section~\ref{sec:3_main} of the main manuscript.
We use low-rank adaptation (LoRA)~\citep{hu2022lora} with a rank of $32$ and a scaling factor of $4$ for both Gemma-2B~\citep{gemma2024open} and Llama-3-8B~\citep{llama32024herd}.
We truncate the MetaMathQA-395K~\citep{yu2024metamath} training dataset to 100,000 examples for both models.
The reported test accuracy is based on a separate, validation-only dataset, GSM8K~\citep{cobbe2021gsm8k}.
Additional experimental results are provided in Table~\ref{tab:gemma2b_metamath_lora_base_full}, where we show results for baseline optimizers with different $\beta_1$ values we have tested.
In the main manuscript, only the best baseline optimizer results are shown for brevity: $\beta_1 = 0.5$ for Gemma-2B and $\beta_1 = 0.9$ for Llama-3-8B.
For our optimal AdamW-type optimizers using the \emph{two-option switch}, we use $\beta_1$ options of $0.8$ and $0.99$, to ensure that these endpoints enclose the typical range of $\beta_1$ values used in practice.

\begin{table}[h!]
\centering
\caption{\small Hyperparameters and settings for the main commonsense finetuning experiments on Gemma-2B~\citep{gemma2024open} with LoRA~\citep{hu2022lora}.}
\label{tab:exp_hyperparams_commonsense_gemma}
\renewcommand{\arraystretch}{1.1}
\resizebox{.8\textwidth}{!}{%
\begin{tabular}{ll}
\toprule
\textbf{Parameter}        & \textbf{Value(s) / Description} \\
\midrule
Dataset                   & Commonsense-170K~\citep{hu2023llmadapters} \\
Models tested             & Gemma-2B~\citep{gemma2024open} \\
Hardware                  & 1 $\times$ A100-80GB \\
Precision                 & Bfloat16 (BF16) \\
Optimizer                 & AdamW~\citep{kingma2015adam,loshchilov2017decoupled} \\
Optimal AdamW-type optimizer & \emph{two-option switch} with $\beta_1$ endpoints of $0.8$ and $0.95$ \\
Epochs                    & $1$ \\
Batch size ($bs$)         & $32$ \\
Learning rate ($lr$)      & $2 \times 10^{-4}$ \\
Weight decay              & $0$ \\
Warmup ratio              & $0$ \\
Adapter type              & LoRA~\citep{hu2022lora} \\
LoRA Rank ($r$)           & $32$ \\
LoRA Scaling ($\alpha$)   & $4$ \\
LoRA Dropout              & $0.05$ \\
Cutoff length             & $256$ \\
Adapter target modules    & \begin{tabular}[t]{@{}l@{}}%
                             \texttt{q\_proj, k\_proj, v\_proj, o\_proj} \\
                             \texttt{down\_proj, up\_proj, gate\_proj}%
                           \end{tabular} \\
\bottomrule
\end{tabular}
}
\end{table}

\begin{table}[h!]
\centering
\caption{\small Hyperparameters and settings for the main commonsense finetuning experiments on Gemma-2B~\citep{gemma2024open} with full fine-tuning.}
\label{tab:exp_hyperparams_commonsense_gemma_full}
\renewcommand{\arraystretch}{1.1}
\resizebox{.8\textwidth}{!}{%
\begin{tabular}{ll}
\toprule
\textbf{Parameter}        & \textbf{Value(s) / Description} \\
\midrule
Dataset                   & Commonsense-170K~\citep{hu2023llmadapters} \\
Models tested             & Gemma-2B~\citep{gemma2024open} \\
Hardware                  & 1 $\times$ A100-80GB \\
Precision                 & Bfloat16 (BF16) \\
Optimizer                 & AdamW~\citep{kingma2015adam,loshchilov2017decoupled} \\
Optimal AdamW-type optimizer & \emph{two-option switch} with $\beta_1$ endpoints of $0.1$ and $0.99$ \\
Epochs                    & $1$ \\
Batch size ($bs$)         & $32$ \\
Learning rate ($lr$)      & $1 \times 10^{-5}$ \\
Weight decay              & $0$ \\
Warmup ratio              & $0$ \\
Adapter type              & Full fine-tuning \\
Cutoff length             & $256$ \\
\bottomrule
\end{tabular}
}
% \vspace{-0.5em}
\end{table}

\paragraph{Gemma-2B on Commonsense-170K.}
We summarize the hyperparameters and training settings for these experiments in Table~\ref{tab:llm_commonsense} in Section~\ref{sec:3_main} of the main manuscript.
We use low-rank adaptation (LoRA)~\citep{hu2022lora} with a rank of $32$ and a scaling factor of $4$ for Gemma-2B~\citep{gemma2024open}.
We also demonstrate our optimizer with full fine-tuning on the Commonsense-170K~\citep{hu2023llmadapters} dataset.
Additional experimental results are provided in Table~\ref{tab:llm_commonsense_full} in the next section, where we show results for baseline optimizers with different $\beta_1$ values we have tested.
In the main manuscript, only the best baseline optimizer results are shown for brevity: $\beta_1 = 0.95$ for LoRA and $\beta_1 = 0.5$ for full fine-tuning.
For our optimal AdamW-type optimizers using the \emph{two-option switch}, we use $\beta_1$ options of $0.1$ and $0.99$ for full fine-tuning and $\beta_1$ options of $0.8$ and $0.95$ for LoRA, to ensure that these endpoints enclose the typical range of $\beta_1$ values used in each type of experiment in practice.
After fitting the Commonsense-170K~\citep{hu2023llmadapters} dataset, we evaluate performance on various reasoning datasets that are commonly used in the LLM literature.
These include BoolQ~\citep{clark2019boolq}, PIQA~\citep{bisk2020piqa}, Social IQA~\citep{sap2019socialiqa}, HellaSwag~\citep{zellers2019hellaswag}, Winogrande~\citep{sakaguchi2019winogrande}, and OBQA~\citep{hu2023llmadapters}.
We also report the average performance across all datasets.

\begin{table}[h!]
\centering
\caption{\small Common hyperparameters and settings for ViT-Base and ViT-Large~\citep{dosovitskiy2021image} finetuning experiments across various classification datasets.}
\label{tab:exp_hyperparams_vit}
\renewcommand{\arraystretch}{1.1}
\resizebox{\textwidth}{!}{
\begin{tabular}{ll}
\toprule
\textbf{Parameter}        & \textbf{Value(s) / Description} \\
\midrule
Models tested & ViT-B / ViT-L~\citep{dosovitskiy2021image} \\
Datasets      & Stanford Cars, CIFAR-100, CUB-200, DTD, Food-101, RESISC45, SUN397 \\
Hardware      & 1 $\times$ RTX 5090-32GB \\
Precision     & Bfloat16 (BF16) \\
Optimizer                 & AdamW~\citep{kingma2015adam,loshchilov2017decoupled} \\
Optimal AdamW-type optimizer & \emph{two-option switch} with $\beta_1$ endpoints of $0.5$ and $0.99$ \\
Batch size ($bs$)      & $256$ \\
Epochs        & 
\begin{tabular}[t]{@{}l@{}}
\textrm{Stanford Cars:} $20$ \\
\textrm{CIFAR-100:} $7$ \\
\textrm{CUB-200:} $25$ \\
\textrm{DTD:} $25$ \\
\textrm{Food-101:} $10$ \\
\textrm{RESISC45:} $10$ \\
\textrm{SUN397:} $15$ \\
\end{tabular} \\
Learning rate ($lr$) &
\begin{tabular}[t]{@{}l@{}}
\textrm{Stanford Cars:} $5\times10^{-3}$ \\
\textrm{CIFAR-100:} $1\times10^{-3}$ \\
\textrm{CUB-200:} $2\times10^{-3}$ \\
\textrm{DTD:} $2\times10^{-3}$ \\
\textrm{Food-101:} $2\times10^{-3}$ \\
\textrm{RESISC45:} $2\times10^{-3}$ \\
\textrm{SUN397:} $1\times10^{-3}$ \\
\end{tabular} \\
Weight decay           & $0$ \\
Warmup ratio           & $0$ \\
Adapter type           & LoRA~\citep{hu2022lora} \\
LoRA Rank ($r$)        & $8$, $32$ \\
LoRA Scaling ($\alpha$)& $4$ \\
LoRA Dropout           & $0$ \\
Target modules         & \texttt{query}, \texttt{value} \\
\bottomrule
\end{tabular}
}
% \vspace{-0.5em}
\end{table}

\paragraph{ViT-B and ViT-L on various classification datasets.}
We summarize the hyperparameters and training settings for these experiments in Table~\ref{tab:vit_tasks} in Section~\ref{sec:3_main} of the main manuscript.
We tested two different configurations:
ViT-Base~\citep{dosovitskiy2021image} with rank-$32$ LoRA and ViT-Large with rank-$8$ LoRA.
The same pre-trained weights are fine-tuned for various task-specific datasets, including Stanford Cars~\citep{krause20133d}, CIFAR-100~\citep{krizhevsky2009learning}, CUB-200~\citep{wah2011cub200}, DTD~\citep{cimpoi2014describing}, Food-101~\citep{bossard2014food101}, RESISC45~\citep{cheng2017resisc45}, and SUN397~\citep{xiao2010sun}.
For our optimal AdamW-type optimizers using the \emph{two-option switch}, we use $\beta_1$ options of $0.5$ and $0.99$ for all configurations.
This covers the working range of $\beta_1$ values used in practice for each type of experiment.
We also report the average performance across all datasets.

\begin{figure}[t]
\centering
\begin{subfigure}[b]{0.48\textwidth}
\centering
\includegraphics[width=\textwidth]{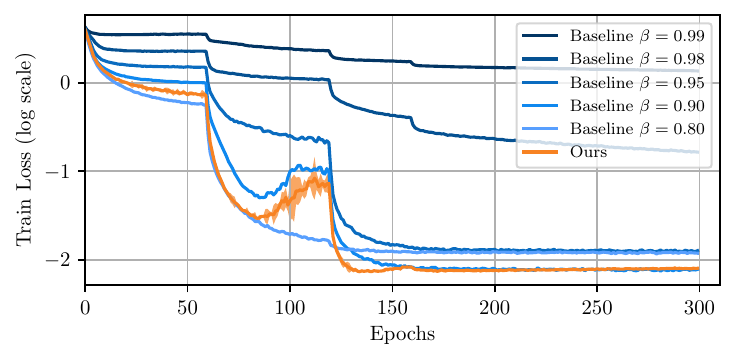}
\end{subfigure}
\hfill
\begin{subfigure}[b]{0.48\textwidth}
\includegraphics[width=\textwidth]{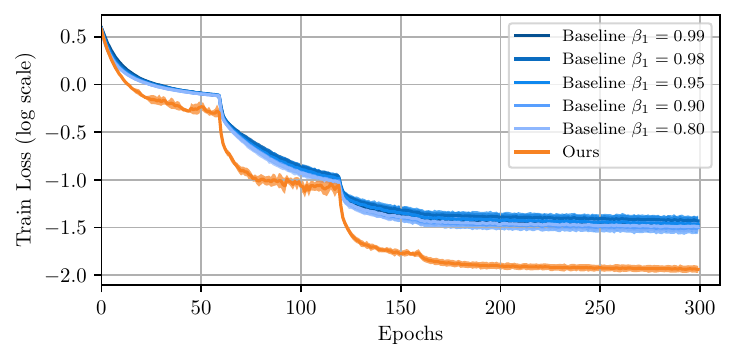}
\end{subfigure}
\\[-1em]
\begin{subfigure}[b]{0.48\textwidth}
\centering
\includegraphics[width=\textwidth]{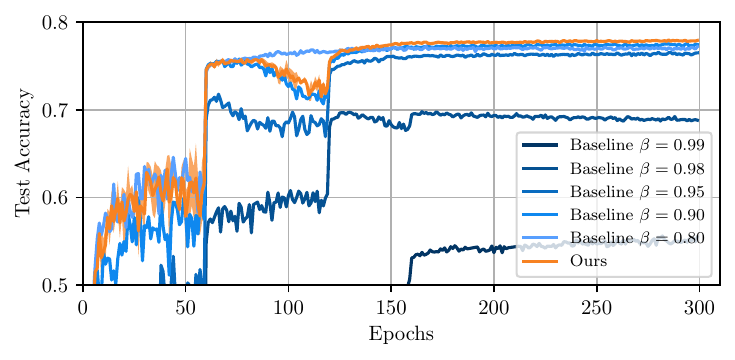}
\caption{\small Training curve of SGD+M optimizers.}
\label{fig:sgd_momentum_train_loss_curve}
\end{subfigure}
\hfill
\begin{subfigure}[b]{0.48\textwidth}
\centering
\includegraphics[width=\textwidth]{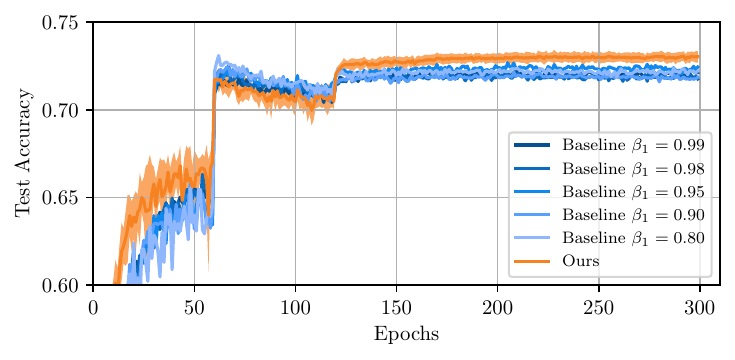}
\caption{\small Training curve of Adam optimizers.}
\label{fig:adam_train_loss_curve}
\end{subfigure}%
% \vspace{-0.5em}
\caption{%
\small
Demonstration of Theorem~\ref{thm:greedy-optimal-sgdm} and Corollary~\ref{cor:greedy-optimal-adam}.
Our instantiations of optimal optimizers are compared with baselines having fixed hyperparameters on the CIFAR-100 dataset~\citep{krizhevsky2009learning} with ResNet-18~\citep{he2016deep}, following the standard settings of \cite{he2016deep}.
The line and shaded area indicate the mean and standard deviation over 10 runs.
For clear visualization, each baseline plot shows only the best run.
For SGD+Momentum, momentum below 0.8 showed suboptimal performance.}
\label{fig:hyperparameter_experiments_curve}
% \vspace{-0.5em}
\end{figure}

\begin{table}[t]
\begin{minipage}[t]{0.49\linewidth}
\centering
\small
\caption{\small
Full test results of SGD+Momentum on CIFAR-100 with ResNet-18.
mean $\pm$ std.}
\label{tab:sgd_momentum_results_full}
\resizebox{\linewidth}{!}{%
\begin{tabular}{lcc}
\toprule
\textbf{Method} & Test acc. \% & Train loss \\
\midrule
$\beta = 0.01$ & 74.93 $\pm$ 0.11 \emptymedal & 0.0091 $\pm$ 0.0002 \emptymedal \\
$\beta = 0.1$ & 75.76 $\pm$ 0.09 \emptymedal & 0.0093 $\pm$ 0.0001 \emptymedal \\
$\beta = 0.2$ & 75.89 $\pm$ 0.08 \emptymedal & 0.0098 $\pm$ 0.0001 \emptymedal \\
$\beta = 0.5$ & 76.21 $\pm$ 0.17 \emptymedal & 0.0115 $\pm$ 0.0001 \emptymedal \\
$\beta = 0.8$ & 77.26 $\pm$ 0.12 \emptymedal & 0.0119 $\pm$ 0.0002 \emptymedal \\
$\beta = 0.9$ & 77.57 $\pm$ 0.09 \bronzemedal & 0.0078 $\pm$ 0.0001 \silvermedal \\
$\beta = 0.95$ & 76.57 $\pm$ 0.32 \emptymedal & 0.0127 $\pm$ 0.0004 \emptymedal \\
$\beta = 0.98$ & 69.79 $\pm$ 0.86 \emptymedal & 0.1648 $\pm$ 0.0148 \emptymedal \\
$\beta = 0.99$ & 55.62 $\pm$ 5.32 \emptymedal & 1.3609 $\pm$ 0.2312 \emptymedal \\
$\beta = 0.995$ & 62.54 $\pm$ 3.49 \emptymedal & 0.7271 $\pm$ 0.0892 \emptymedal \\
$\beta = 0.999$ & 68.48 $\pm$ 3.25 \emptymedal & 0.1772 $\pm$ 0.0229 \emptymedal \\
\midrule
\textbf{Ours (2 options)} & 78.06 $\pm$ 0.07 \silvermedal & 0.0080 $\pm$ 0.0001 \bronzemedal \\
\textbf{Ours (5 options)} & 78.33 $\pm$ 0.49 \goldmedal & 0.0073 $\pm$ 0.0001 \goldmedal \\
\bottomrule
\end{tabular}%
}
\end{minipage}\hfill
\begin{minipage}[t]{0.49\linewidth}
\centering
\small
\caption{\small
Full test results of Adam on CIFAR-100 with ResNet-18.
mean $\pm$ std.}
\label{tab:adam_results_full}
\resizebox{\linewidth}{!}{%
\begin{tabular}{lcc}
\toprule
\textbf{Method} & Test acc. \% & Train loss \\
\midrule
$\beta_1 = 0.1$ & 72.78$\pm$ 0.43 \emptymedal & 0.0414 $\pm$ 0.0037 \emptymedal \\
$\beta_1 = 0.2$ & 72.65 $\pm$ 0.18 \emptymedal & 0.0396 $\pm$ 0.0023 \emptymedal \\
$\beta_1 = 0.5$ & 72.86 $\pm$ 0.14 \bronzemedal & 0.0351 $\pm$ 0.0019 \emptymedal \\
$\beta_1 = 0.8$ & 73.20 $\pm$ 0.21 \silvermedal & 0.0324 $\pm$ 0.0042 \bronzemedal \\
$\beta_1 = 0.9$ & 72.85 $\pm$ 0.38 \emptymedal & 0.0314 $\pm$ 0.0044 \silvermedal \\
$\beta_1 = 0.95$ & 72.78 $\pm$ 0.38 \emptymedal & 0.0347 $\pm$ 0.0068 \emptymedal \\
$\beta_1 = 0.98$ & 72.69 $\pm$ 0.20 \emptymedal & 0.0372 $\pm$ 0.0038 \emptymedal \\
$\beta_1 = 0.99$ & 72.45 $\pm$ 0.20 \emptymedal & 0.0333 $\pm$ 0.0045 \emptymedal \\
\midrule
\textbf{Ours (2 options)} & 73.26 $\pm$ 0.31 \goldmedal & 0.0115 $\pm$ 0.0010 \goldmedal \\
\bottomrule
\end{tabular}%
}
\end{minipage}
\end{table}

\begin{table}[t]
\centering
\small
\caption{\small Full test results of Gemma-2B trained with MetaMathQA-395K, validated on GSM8K.
mean $\pm$ std.}
\label{tab:gemma2b_metamath_lora_base_full}
% \vspace{-0.5em}
\resizebox{0.47\textwidth}{!}{%
\begin{tabular}{lcc}
\toprule
Method & GSM8K acc. (\%) & Train loss \\
\midrule
$\beta_1 =0.5$ & 52.57 $\pm$ 1.10 \silvermedal & 0.2080 $\pm$ 0.0004 \goldmedal \\
$\beta_1 =0.8$ & 52.31 $\pm$ 1.00 \bronzemedal & 0.2080 $\pm$ 0.0004 \goldmedal \\
$\beta_1 =0.9$ & 51.76 $\pm$ 0.99 \emptymedal & 0.2081 $\pm$ 0.0004 \silvermedal \\
$\beta_1 =0.95$ & 51.12 $\pm$ 0.77 \emptymedal & 0.2085 $\pm$ 0.0004 \emptymedal \\
$\beta_1 =0.98$ & 51.25 $\pm$ 0.38 \emptymedal & 0.2093 $\pm$ 0.0005 \emptymedal \\
$\beta_1 =0.99$ & 50.97 $\pm$ 0.68 \emptymedal & 0.2103 $\pm$ 0.0004 \emptymedal \\
\midrule
\textbf{Ours} & 52.77 $\pm$ 0.93 \goldmedal & 0.2084 $\pm$ 0.0003 \bronzemedal \\
\bottomrule
\end{tabular}%
}
\end{table}

\subsection{Extended experimental results}
\label{sec:appx:exp:results}

This section provides additional experimental results for the main manuscript.
For visual clarity, we use gold, silver, and bronze medals to denote the best, second-best, and third-best results, respectively, in all tables hereafter.

\paragraph{ResNet-18 on CIFAR-100.}
Figures~\ref{fig:sgd_momentum_test_acc_errbar} and~\ref{fig:adam_test_acc_errbar} in the main manuscript demonstrate how optimizer hyperparameters affect the final training loss and validation accuracy, and how our optimal optimizers compare to baseline optimizers with fixed hyperparameters.
Here, the complete results are provided in Tables~\ref{tab:sgd_momentum_results_full} and~\ref{tab:adam_results_full}.
Figure~\ref{fig:hyperparameter_experiments_curve} further compares the training curves of baseline optimizers and our optimal optimizers.
For clarity, each baseline plot shows only the best run, and only baselines with robust momentum values are displayed.
Specifically, we show results for $\beta \in [0.8, 0.99]$ for SGD+Momentum and $\beta_1 \in [0.8, 0.999]$ for Adam~\citep{kingma2015adam}.
Although the abrupt learning rate decay of the scheduler introduces perturbations that are not considered in our theory, our implementation of the greedy optimal optimizer generally reduces the training loss rapidly and achieves better performance than the baselines.
Moreover, the complete results in Tables~\ref{tab:sgd_momentum_results_full} and~\ref{tab:adam_results_full} show that by increasing the number of selectable options from two to five, our instantiation of the greedy optimal optimizers achieves significantly better performance than baseline optimizers with any fixed hyperparameters.
This opens up new opportunities for research into the \emph{dynamic hyperparameter tuning} framework, which is first enabled by our theory.

\begin{table}[t]
\centering
\small
\caption{\small Full test results of Gemma-2B trained with Commonsense-170K.
mean $\pm$ std.}
\label{tab:llm_commonsense_full}
% \vspace{-0.5em}
\resizebox{1.0\textwidth}{!}{%
\begin{tabular}{l|cccccc|c}
\toprule
\textit{Gemma-2B (LoRA)} & BoolQ & PIQA & Social IQA & HellaSwag & Winogrande & OBQA & \textbf{Avg} \\
\midrule
$\beta_1 = 0.5$ & 65.25 $\pm$ 0.28 \emptymedal & 78.73 $\pm$ 0.27 \emptymedal & 73.97 $\pm$ 0.50 \goldmedal & 72.81 $\pm$ 1.54 \emptymedal & 70.96 $\pm$ 0.90 \emptymedal & 72.07 $\pm$ 0.34 \emptymedal & 72.30 $\pm$ 0.32 \emptymedal \\
$\beta_1 = 0.8$ & 65.42 $\pm$ 0.20 \emptymedal & 78.80 $\pm$ 0.71 \emptymedal & 73.51 $\pm$ 0.23 \emptymedal & 73.46 $\pm$ 1.21 \bronzemedal & 71.43 $\pm$ 0.28 \emptymedal & 72.20 $\pm$ 0.34 \emptymedal & 72.47 $\pm$ 0.25 \emptymedal \\
$\beta_1 = 0.9$ & 65.31 $\pm$ 0.27 \emptymedal & 78.87 $\pm$ 0.67 \bronzemedal & 73.66 $\pm$ 0.37 \silvermedal & 72.97 $\pm$ 1.47 \emptymedal & 71.40 $\pm$ 0.30 \emptymedal & 73.20 $\pm$ 0.65 \silvermedal & 72.57 $\pm$ 0.30 \bronzemedal \\
$\beta_1 = 0.95$ & 65.69 $\pm$ 0.29 \goldmedal & 78.93 $\pm$ 0.49 \silvermedal & 73.61 $\pm$ 0.16 \bronzemedal & 74.07 $\pm$ 0.28 \silvermedal & 71.61 $\pm$ 0.44 \bronzemedal & 72.67 $\pm$ 0.68 \emptymedal & 72.76 $\pm$ 0.17 \silvermedal \\
$\beta_1 = 0.98$ & 65.65 $\pm$ 0.27 \silvermedal & 78.82 $\pm$ 0.40 \emptymedal & 73.52 $\pm$ 0.34 \emptymedal & 68.47 $\pm$ 1.83 \emptymedal & 71.45 $\pm$ 0.21 \emptymedal & 71.67 $\pm$ 1.23 \emptymedal & 71.60 $\pm$ 0.38 \emptymedal \\
$\beta_1 = 0.99$ & 65.48 $\pm$ 0.43 \bronzemedal & 78.69 $\pm$ 0.56 \emptymedal & 73.13 $\pm$ 0.15 \emptymedal & 68.66 $\pm$ 0.95 \emptymedal & 72.01 $\pm$ 0.52 \goldmedal & 73.00 $\pm$ 0.43 \bronzemedal & 71.83 $\pm$ 0.23 \emptymedal \\
\midrule
\textbf{Ours} & 65.31 $\pm$ 0.04 \emptymedal & 79.00 $\pm$ 0.36 \goldmedal & 73.58 $\pm$ 0.06 \emptymedal & 75.09 $\pm$ 1.02 \goldmedal & 71.80 $\pm$ 0.39 \silvermedal & 73.27 $\pm$ 1.15 \goldmedal & 73.01 $\pm$ 0.27 \goldmedal \\
\midrule
\midrule
\textit{Gemma-2B (Full FT)} & BoolQ & PIQA & Social IQA & HellaSwag & Winogrande & OBQA & \textbf{Avg} \\
\midrule
$\beta_1 = 0.5$ & 62.79 $\pm$ 0.27 \silvermedal & 74.12 $\pm$ 0.26 \silvermedal & 66.63 $\pm$ 0.33 \silvermedal & 40.50 $\pm$ 1.15 \silvermedal & 61.48 $\pm$ 0.32 \silvermedal & 62.60 $\pm$ 1.02 \silvermedal & 61.35 $\pm$ 0.27 \silvermedal \\
$\beta_1 = 0.8$ & 62.50 $\pm$ 0.22 \bronzemedal & 72.62 $\pm$ 0.49 \bronzemedal & 64.02 $\pm$ 0.19 \bronzemedal & 40.11 $\pm$ 0.21 \bronzemedal & 54.38 $\pm$ 0.62 \bronzemedal & 57.20 $\pm$ 0.71 \bronzemedal & 58.47 $\pm$ 0.19 \bronzemedal \\
$\beta_1 = 0.9$ & 62.42 $\pm$ 0.24 \emptymedal & 72.05 $\pm$ 0.38 \emptymedal & 62.88 $\pm$ 0.54 \emptymedal & 39.45 $\pm$ 0.33 \emptymedal & 52.28 $\pm$ 0.26 \emptymedal & 55.47 $\pm$ 0.52 \emptymedal & 57.43 $\pm$ 0.16 \emptymedal \\
$\beta_1 = 0.95$ & 62.38 $\pm$ 0.20 \emptymedal & 71.60 $\pm$ 0.16 \emptymedal & 62.20 $\pm$ 0.36 \emptymedal & 39.14 $\pm$ 0.16 \emptymedal & 51.33 $\pm$ 0.48 \emptymedal & 54.13 $\pm$ 0.98 \emptymedal & 56.80 $\pm$ 0.20 \emptymedal \\
$\beta_1 = 0.98$ & 62.42 $\pm$ 0.18 \emptymedal & 70.84 $\pm$ 0.65 \emptymedal & 61.19 $\pm$ 0.24 \emptymedal & 38.10 $\pm$ 0.24 \emptymedal & 51.09 $\pm$ 0.15 \emptymedal & 53.07 $\pm$ 0.34 \emptymedal & 56.12 $\pm$ 0.14 \emptymedal \\
$\beta_1 = 0.99$ & 62.25 $\pm$ 0.01 \emptymedal & 70.82 $\pm$ 0.64 \emptymedal & 60.70 $\pm$ 0.25 \emptymedal & 37.41 $\pm$ 0.55 \emptymedal & 50.72 $\pm$ 0.43 \emptymedal & 52.87 $\pm$ 1.09 \emptymedal & 55.80 $\pm$ 0.24 \emptymedal \\
\midrule
\textbf{Ours} & 63.29 $\pm$ 0.78 \goldmedal & 75.70 $\pm$ 0.22 \goldmedal & 68.41 $\pm$ 0.69 \goldmedal & 42.47 $\pm$ 1.06 \goldmedal & 62.46 $\pm$ 4.64 \goldmedal & 64.40 $\pm$ 0.86 \goldmedal & 62.79 $\pm$ 0.83 \goldmedal \\
\bottomrule
\end{tabular}%
}
% \vspace{-0.5em}
\end{table}

\paragraph{Gemma-2B and Llama-3-8B on MetaMathQA-395K.}
We extend the experimental results in Table~\ref{tab:gemma_mathqa} of the main manuscript by providing the full baseline results in Table~\ref{tab:gemma2b_metamath_lora_base_full}.
The results show that our optimal optimizer yields a training loss comparable to the best baseline optimizer, while achieving better validation accuracy.
Importantly, this achievement does not result from tedious manual hyperparameter tuning, but from our \emph{dynamic hyperparameter tuning} framework enabled by our theory.

\begin{figure}[t]
\centering
\begin{minipage}{0.48\linewidth}
    \centering
    \includegraphics[width=\linewidth]{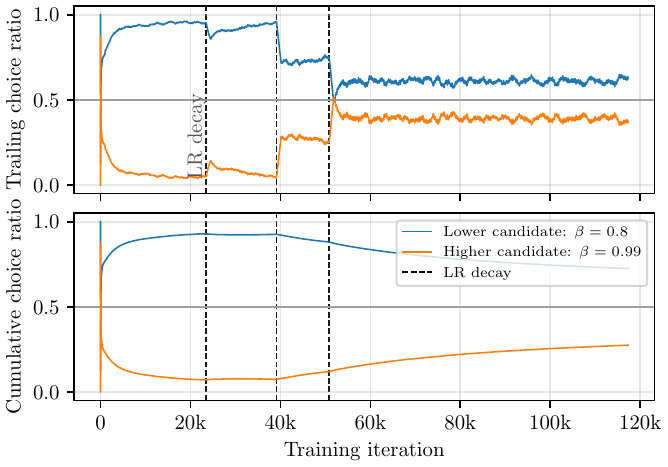}
    \vspace{-1.5em}
    \subcaption{\small SGDM: selection ratio over training.}
\end{minipage}
\hfill
\begin{minipage}{0.48\linewidth}
    \centering
    \includegraphics[width=\linewidth]{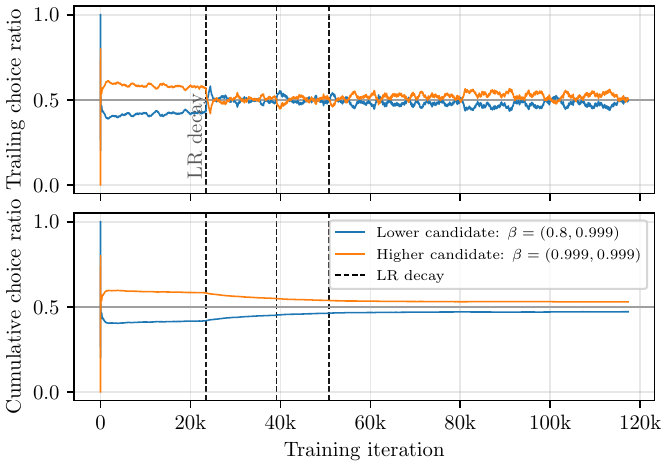}
    \vspace{-1.5em}
    \subcaption{\small Adam: selection ratio over training.}
\end{minipage}
\caption{\small Selection ratios of the candidate hyperparameters over training for SGDM and Adam.
The experiments are the same ResNet-18 / CIFAR-100 runs in Figures~\ref{fig:sgd_momentum_test_acc_errbar} and~\ref{fig:adam_test_acc_errbar}.
}
\label{fig:beta-score-2x2}
\end{figure}

\paragraph{Gemma-2B on Commonsense-170K.}
An extended version of the results in Table~\ref{tab:llm_commonsense} of the main manuscript is provided in Table~\ref{tab:llm_commonsense_full}.
As in the previous experiments, our optimal optimizer achieves comparable and occasionally better performance than the best baseline optimizer with fixed hyperparameters.
In the main manuscript, we provided an abbreviated version of this table that only includes the best baseline optimizer.
For reference, the best baseline is $\beta_1 = 0.95$ for LoRA training and $\beta_1 = 0.5$ for full fine-tuning.

\paragraph{Selected-$\beta$ and score trajectories.}
We additionally provide visualizations of the selected hyperparameters and selection ratios over training for SGD+Momentum and Adam for the experiments in Figures~\ref{fig:sgd_momentum_test_acc_errbar} and~\ref{fig:adam_test_acc_errbar}.
Our adaptive $K$-Switch framework rapidly switches between the optimizer hyperparameters, and the candidate preferences drift along the learning process.
Figure~\ref{fig:beta-score-2x2} shows this drift of the cumulative selection ratios of each candidate over training.
We observe gradual changes in the optimizers' hyperparameter preferences towards equal selection ratios (0.5) as training progresses, while learning rate decay abruptly changes the preferences of each candidate.

\newpage
\section{Proofs omitted from the main text}
\label{sec:appx:proofs}
This section presents the proofs omitted from the main text.
The proofs are organized in the same order as the theorems appear in the main manuscript.

\paragraph{Roadmap.}
Proposition~\ref{prop:learning-power-as-inner-product} expands the causal convolution and uses Fubini--Tonelli to identify the filter--moment pairing.
Theorem~\ref{thm:dynamic-optimizer-under-convex-constraints} recognizes the objective as the support function of a compact convex family.
Corollary~\ref{cor:family-hull-reduction} is linearity over the closed convex hull.
Theorem~\ref{thm:greedy-optimal-sgdm} and Corollary~\ref{cor:greedy-optimal-adam} score the unit-energy Hilbert or weighted-diagonal section of the one-pole family.
Theorem~\ref{thm:finite-step-descent-bridge} is the descent lemma.
Corollary~\ref{cor:greedy-near-optimal} establishes local near-optimality of greedy alignment over norm-bounded candidate updates.
Proposition~\ref{prop:online-stability} transfers the Lipschitz bound to finite-sample scores.
The basic strategies are also summarized in Table~\ref{tab:proof-roadmap}.

\begin{table}[t]
\centering
\small
\caption{\small Summary of the theorems presented in the main text. The proof of each theorem is linked next to its title with button: [\texttt{proof}]. All the proofs are located in Appendix~\ref{sec:appx:proofs}.}
\label{tab:proof-roadmap}
\resizebox{\textwidth}{!}{%
\begin{tabular}{lll}
\toprule
\textbf{Result} & \textbf{What it says} & \textbf{In plain terms} \\
\midrule
Prop.~\ref{prop:learning-power-as-inner-product} & $P_t(Q)=\langle Q,R_t\rangle_{\gH}$ under A1--A3 & Alignment is a filter--statistics inner product. \\
Thm.~\ref{thm:dynamic-optimizer-under-convex-constraints} & $\max_{Q\in\gQ} P$ is the support function of $\gQ$ & A compact convex family has a greedy optimum, stable in $R$. \\
Cor.~\ref{cor:family-hull-reduction} & $\sup_{\mathrm{cl\,co}(\mathcal F\cup\{0\})} P = (\sup_{\mathcal F} P)_+$ & Scoring in the hyperparameterized set $\gF$ is exact. \\
Thm.~\ref{thm:greedy-optimal-sgdm} & $\beta^\star=\arg\max_\beta \sqrt{1-\beta^2}\,\mathbb E[\vg^\top \vm_\beta]$ & Hilbert-energy-normalized SGDM selection. \\
Cor.~\ref{cor:greedy-optimal-adam} & state-conditioned closed form for $(\beta_1,\beta_2)$ & Hilbert-energy-normalized Adam/AdamW selection. \\
Thm.~\ref{thm:finite-step-descent-bridge} & $|D_\eta-\eta A|\le L_s\eta^2\|\vu\|^2/2$ & Alignment is the first-order loss drop; $\eta$ must stay local. \\
Cor.~\ref{cor:greedy-near-optimal} & $D_\eta$ gap $L_s\eta^2 \rho^2$ on $\|\vu\|\le \rho$ & Greedy alignment is locally near-optimal for bounded updates. \\
Prop.~\ref{prop:online-stability} & gap $>2\varepsilon$ $\Rightarrow$ correct $\arg\max$ & Instantaneous scores work if the filter difference > noise. \\
\bottomrule
\end{tabular}%
}
\end{table}

\subsection{Proof of Proposition~\ref{prop:learning-power-as-inner-product}: Learning power is inner product}
\label{sec:appx:proof-of-proposition-learning-power-as-inner-product}

\begin{proof}[Proof of Proposition~\ref{prop:learning-power-as-inner-product}]
Here we provide a detailed derivation of \eqref{eq:sec_2:instantaneous_power} that was abbreviated in the main text.
We assume (A1) the gradient stream $\vg_t$ has finite and uniformly bounded second moments: $M := \sup_{t\in \mathbb Z} \E\|\vg_t\|^2 < \infty$, (A2) the optimizer filter $Q_t$ is absolutely summable in operator norm: $\sum_{k=0}^{\infty} \|Q_{t,k}\|_{\text{op}} < \infty$, and (A3) the lag-$k$ moment sequence $R_t = \{R_{t,k}\}_{k \geq 0}$ belongs to $\gH$, so that the final pairing is a Hilbert inner product.
The lag-$k$ moment is defined as $R_{t,k}=\E[\vg_t\,\vg_{t-k}^\top]$.
We start from the convolution definition of the dynamic optimizer:
\begin{equation}
\vu_t = \sum_{k=0}^\infty Q_{t,k}\,\vg_{t-k}.
\end{equation}
Then we can calculate the instantaneous power as follows:
\begin{align}
P_t(Q)
&= \E\big[\vg_t^\top \vu_t\big] \\
&= \E\Big[\vg_t^\top \sum_{k=0}^{\infty} Q_{t,k}\, \vg_{t-k}\Big] \\
&= \sum_{k=0}^{\infty} \E\big[\vg_t^\top Q_{t,k}\, \vg_{t-k}\big] \quad \text{(linearity of $\E$)} \\
&= \sum_{k=0}^{\infty} \mathrm{Tr}\!\big(Q_{t,k}\; \E[\vg_{t-k}\, \vg_t^\top]\big) \\
&= \sum_{k=0}^{\infty} \mathrm{Tr}\!\big(Q_{t,k}^\top\; \E[\vg_t\, \vg_{t-k}^\top]\big) \quad \text{(trace transpose)} \\
&= \sum_{k=0}^{\infty} \mathrm{Tr}\!\big(Q_{t,k}^\top\, R_{t,k}\big) \\
&= \langle Q_t, R_t \rangle_{\gH}.
\end{align}
The interchange of summation and expectation is justified by Fubini--Tonelli as follows. By Cauchy--Schwarz,
\begin{equation}
\E\big|\vg_t^\top Q_{t,k}\, \vg_{t-k}\big| \leq \|Q_{t,k}\|_{\text{op}} \sqrt{\E\|\vg_t\|^2} \sqrt{\E\|\vg_{t-k}\|^2} \leq M \|Q_{t,k}\|_{\text{op}}.
\end{equation}
Therefore,
\begin{equation}
\sum_{k=0}^{\infty} \E\big|\vg_t^\top Q_{t,k}\, \vg_{t-k}\big| \leq M \sum_{k=0}^{\infty} \|Q_{t,k}\|_{\text{op}} < \infty,
\end{equation}
which establishes absolute convergence.
The scalar-trace identity $\va^\top B \vc = \mathrm{Tr}(B \vc \va^\top)$ is used in the fourth line.
This completes the derivation of~\eqref{eq:sec_2:instantaneous_power}.
\end{proof}

\subsection{Proof of Theorem~\ref{thm:dynamic-optimizer-under-convex-constraints}: Greedy optimal first-order optimizers}
\label{sec:appx:proof-of-theorem-dynamic-optimizer-under-convex-constraints}

\begin{proof}[Proof of Theorem~\ref{thm:dynamic-optimizer-under-convex-constraints}]
We establish each claim in turn. Throughout, we assume $0 \in \gQ$ as stated in the theorem.

\emph{(i) Existence:}
Since $\gQ$ is compact and $Q \mapsto \langle Q, R \rangle_{\gH}$ is continuous, the maximum is attained by the Weierstrass extreme value theorem.
The optimal power $P^\star(R) = \sup_{Q \in \gQ} \langle Q, R \rangle_{\gH}$ is a supremum of linear functionals in $R$, hence sublinear (convex and positively homogeneous).
Finiteness follows from compactness of $\gQ$.

\emph{(ii) Construction:}
We establish the following conjugacy identities in the dynamic setting.

\begin{equation}
    \label{eq:sec_appx:6:conjugacy-identities}
    P^\star \;=\; \sigma_{\gQ} \;=\; \delta_{\gQ}^{*} \;=\; \gamma_{\gQ^\circ}, \qquad (\gamma_{\gQ})^* \;=\; \delta_{\gQ^\circ}.
\end{equation}

Here $\sigma_{\gQ}(R) = \sup_{Q \in \gQ} \langle Q, R \rangle_{\gH}$ denotes the support function of $\gQ$.

\textit{(1) Optimal power = support function = conjugate of indicator.}
By the definition of convex conjugate in $\gH$,
\begin{equation}
(\delta_{\gQ})^*(R) = \sup_{Q \in \gH} \{\langle Q, R \rangle_{\gH} - \delta_{\gQ}(Q)\} = \sup_{Q \in \gQ} \langle Q, R \rangle_{\gH} = \sigma_{\gQ}(R) = P^\star(R).
\end{equation}
Thus $P^\star = \sigma_{\gQ} = (\delta_{\gQ})^*$.

\textit{(2) Optimal power = gauge of polar.}
By the definition of polar in $\gH$, $R \in \gQ^\circ$ if and only if $\sup_{Q \in \gQ} \langle Q, R \rangle_{\gH} \leq 1$, i.e., $P^\star(R) \leq 1$.
Therefore
\begin{equation}
\gamma_{\gQ^\circ}(R) = \inf\{\lambda > 0 : R \in \lambda \gQ^\circ\} = \inf\{\lambda > 0 : P^\star(R) \leq \lambda\} = P^\star(R).
\end{equation}
Thus $P^\star = \gamma_{\gQ^\circ}$.

\textit{(3) Conjugate of gauge = indicator of polar.} We establish $(\gamma_{\gQ})^* = \delta_{\gQ^\circ}$. Consider two cases:
\begin{itemize}[leftmargin=1.5em]
\item If $R \in \gQ^\circ$, then for all $Q$,
\begin{equation}
\langle Q, R \rangle_{\gH} \leq \gamma_{\gQ}(Q) \cdot \sup_{Q' \in \gQ} \langle Q', R \rangle_{\gH} \leq \gamma_{\gQ}(Q),
\end{equation}
since $\sup_{Q' \in \gQ} \langle Q', R \rangle_{\gH} \leq 1$ by definition of polar.
Hence $\langle Q, R \rangle_{\gH} - \gamma_{\gQ}(Q) \leq 0$ for all $Q$, with equality at $Q = 0$ (using $0 \in \gQ$). Taking the supremum gives $(\gamma_{\gQ})^*(R) = 0 = \delta_{\gQ^\circ}(R)$.

\item If $R \notin \gQ^\circ$, there exists $Q_0 \in \gQ$ with $\langle Q_0, R \rangle_{\gH} > 1$. Since $Q_0 \in \gQ$, we have $\gamma_{\gQ}(Q_0) \leq 1$. For any $\alpha > 0$, we have $\gamma_{\gQ}(\alpha Q_0) = \alpha \gamma_{\gQ}(Q_0)$, and thus
\begin{align*}
\langle \alpha Q_0, R \rangle_{\gH} - \gamma_{\gQ}(\alpha Q_0) &= \alpha \langle Q_0, R \rangle_{\gH} - \alpha \gamma_{\gQ}(Q_0) \\
&= \alpha(\langle Q_0, R \rangle_{\gH} - \gamma_{\gQ}(Q_0)) \\
&\geq \alpha(\langle Q_0, R \rangle_{\gH} - 1) \to +\infty
\end{align*}
as $\alpha \to \infty$, since $\langle Q_0, R \rangle_{\gH} > 1$.
Hence $(\gamma_{\gQ})^*(R) = +\infty = \delta_{\gQ^\circ}(R)$.
\end{itemize}
Thus $(\gamma_{\gQ})^* = \delta_{\gQ^\circ}$.

This completes the proof of the conjugacy identities. We now prove the construction of the optimal optimizer using these identities.

Let $Q^\star \in \arg\max_{Q \in \gQ} \langle Q, R \rangle_{\gH}$. For any $M \in \gH$,
\begin{equation}
P^\star(M) = \max_{Q \in \gQ} \langle Q, M \rangle_{\gH} \geq \langle Q^\star, M \rangle_{\gH} = \langle Q^\star, R \rangle_{\gH} + \langle Q^\star, M - R \rangle_{\gH} = P^\star(R) + \langle Q^\star, M - R \rangle_{\gH},
\end{equation}
which is the defining inequality for $Q^\star \in \partial_R P^\star(R)$.
If the maximizer is unique, then $\partial_R P^\star(R) = \{Q^\star\}$. In finite-dimensional settings, or under standard support-function regularity conditions, this implies Gâteaux differentiability with $Q^\star = \nabla_{\!R}\, P^\star(R) = \nabla_{\!R}\, \gamma_{\gQ^\circ}(R)$.

\emph{(iii) Lipschitz continuity in symmetrized polar gauge:}
We establish the one-sided bounds first. Since $P^\star = \gamma_{\gQ^\circ}$ by the conjugacy identities, we have:
\begin{align}
P^\star(R) - P^\star(\hat{R}) &= \max_{Q \in \gQ} \langle Q, R \rangle_{\gH} - \max_{Q \in \gQ} \langle Q, \hat{R} \rangle_{\gH} \\
&\leq \max_{Q \in \gQ} \langle Q, R - \hat{R} \rangle_{\gH} \\
&= \gamma_{\gQ^\circ}(R - \hat{R}).
\end{align}
Similarly, $P^\star(\hat{R}) - P^\star(R) \leq \gamma_{\gQ^\circ}(\hat{R} - R)$. Therefore,
\begin{equation}
|P^\star(R) - P^\star(\hat{R})| \leq \max\{\gamma_{\gQ^\circ}(R - \hat{R}), \gamma_{\gQ^\circ}(\hat{R} - R)\} = \|R - \hat{R}\|_{\gQ^\circ}^{\mathrm{sym}}.
\end{equation}

Therefore, by using an estimated moment $\hat{R}$ instead of the true moment $R$, the error in optimal power can be bounded by $|P^\star(R) - P^\star(\hat{R})| \leq \|R - \hat{R}\|_{\gQ^\circ}^{\mathrm{sym}}$.
\end{proof}

\subsection{Proof of Corollary~\ref{cor:family-hull-reduction}: Family-hull reduction}
\label{sec:appx:proof-of-corollary-family-hull-reduction}

\begin{proof}[Proof of Corollary~\ref{cor:family-hull-reduction}]
Write any $Q \in \operatorname{co}(\mathcal{F} \cup \{0\})$ as $Q = \pi_0 \cdot 0 + \sum_{j=1}^m \pi_j Q_{\phi_j}$ with $\pi_0, \pi_j \ge 0$ and $\pi_0 + \sum_j \pi_j = 1$.
Let $M \coloneqq \sup_\phi \langle Q_\phi, R \rangle_{\gH}$.
By linearity, $\langle Q, R \rangle_{\gH} = \sum_j \pi_j \langle Q_{\phi_j}, R \rangle_{\gH}$.

\emph{Case $M \ge 0$:}
Each $\langle Q_{\phi_j}, R \rangle_{\gH} \le M$ and $\sum_j \pi_j \le 1$, so $\langle Q, R \rangle_{\gH} \le M$.
If the supremum is attained by some $Q_{\phi^\star}$ with $\langle Q_{\phi^\star}, R \rangle_{\gH} = M$, then this element attains the bound; otherwise the equality $\sup = M$ is understood at the level of suprema.

\emph{Case $M < 0$:}
Every $\langle Q_{\phi_j}, R \rangle_{\gH} < 0$, so $\langle Q, R \rangle_{\gH} \le 0$ with equality at $Q = 0$.

In both cases, $\sup_{Q \in \operatorname{co}(\mathcal{F} \cup \{0\})} \langle Q, R \rangle_{\gH} = M_+$.
Passing to the closure preserves the supremum by continuity of $Q \mapsto \langle Q, R \rangle_{\gH}$.
Proposition~\ref{prop:learning-power-as-inner-product} gives $\langle Q_\phi, R \rangle_{\gH} = \E[\vg_t^\top (Q_\phi \vg)_t]$.
\end{proof}

\subsection{Proof of Theorem~\ref{thm:greedy-optimal-sgdm}: Greedy optimal SGD+Momentum}
\label{sec:appx:proof-of-theorem-greedy-optimal-sgdm}

\begin{proof}[Proof of Theorem~\ref{thm:greedy-optimal-sgdm}]
We work in the impulse-space as defined in Section~\ref{sec:3_main}, i.e., a Hilbert space $(\gH,\langle\cdot,\cdot\rangle_{\gH})$ of causal LTI filters with matrix impulse response $\{q_k\}_{k\ge0}$ with a Hilbert norm $\|Q\|_{\gH}^2 = \sum_{k=0}^{\infty}\operatorname{Tr}(q_k^\top q_k)$, where $k$ is the lag index.

By the main-text definitions, the SGD+Momentum family is $\gQ_{\eta,\beta} = \gQ_\textnormal{1p} \cap \gQ_\textnormal{H}(B)$, where $\gQ_\textnormal{1p} = \{Q_{\beta,k} = \eta\beta^k I : \eta \ge 0,\, \beta \in \gB \subset [0,1)\}$ is the 1-pole family and $\gQ_\textnormal{H}(B) = \{Q : \|Q\|_{\gH} \le \sqrt{B}\}$ is the Hilbert trust region.
We solve $\max_{Q \in \gQ_{\eta,\beta}} \langle Q, R\rangle_{\gH}$ by parameterizing $Q = Q_{\text{SGD+M}}$ via the 1-pole impulse response and saturating the trust region.

\emph{Step 1 — Norm of the 1-pole optimizer.}
The impulse response is $Q_{\beta,k} = \eta\beta^k I$ in the family $\gQ_\textnormal{1p}$. By definition,
\begin{align}
\|Q_{\text{SGD+M}}\|_{\gH}^2 &= \sum_{k\ge0}\operatorname{Tr}(Q_{\beta,k}^\top Q_{\beta,k}) = \eta^2\sum_{k\ge0}\beta^{2k}\operatorname{Tr}(I) = \frac{\eta^2 d}{1-\beta^2},
\end{align}
where $d$ is the parameter dimension.
The trust region constraint $\|Q_{\text{SGD+M}}\|_{\gH} \le \sqrt{B}$ imposes
\begin{equation}
\eta \le \sqrt{B(1-\beta^2)/d}.
\end{equation}

\emph{Step 2 — Alignment with the moment operator.}
The inner product with $R$ is
\begin{align}
\langle Q_{\text{SGD+M}},R\rangle_{\gH} &= \sum_{k\ge0}\operatorname{Tr}(Q_{\beta,k}^\top R_{t,k}) = \eta\sum_{k\ge0}\beta^k\,\operatorname{Tr}(R_{t,k}) = \eta\sum_{k\ge0}\beta^k T_{t,k},
\end{align}
where $T_{t,k} \coloneqq \operatorname{Tr}(R_{t,k})$.

\emph{Step 3 — Reduce to 1-D search; saturate trust region under positive score.}
Define the score function $S_t(\beta) \coloneqq \sum_{k\ge0} \beta^k T_{t,k}$.
For fixed $\beta \in \gB$, if $S_t(\beta) > 0$, the objective is linear increasing in $\eta$ and the maximizer saturates the trust region; if $S_t(\beta) \le 0$, the best gain is $\eta = 0$.
Since $\beta = 0$ gives $S_t(0) = T_{t,0} = \E\|\vg_t\|^2 > 0$ whenever $0 \in \gB$ and the gradient is nonzero, the global nontrivial maximizer has positive score.

The trust region-normalized gain is
\begin{equation}
A(\beta;t) \coloneqq \frac{\langle Q_{\text{SGD+M}},R\rangle_{\gH}}{\|Q_{\text{SGD+M}}\|_{\gH}} = \frac{\sqrt{1-\beta^2}}{\sqrt{d}} \sum_{k\ge0}\beta^k T_{t,k} = \frac{\sqrt{1-\beta^2}}{\sqrt{d}} S_t(\beta).
\end{equation}
Hence the optimal momentum and learning rate are
\begin{equation}
\beta^\star_t = \arg\max_{\beta \in \gB}\, \sqrt{1-\beta^2}\,S_t(\beta), \quad
\eta^\star_t = \sqrt{B(1-(\beta^\star_t)^2)/d}.
\end{equation}

\emph{Step 4 — Solving $A$ for streaming gradients.}
Let $\vm_{\beta,t} = \vg_t + \beta\vg_{t-1} + \beta^2 \vg_{t-2} + \cdots$ be the unnormalized momentum at time $t$.
This can be obtained from a sequential filtering process:
\begin{equation}
\vm_{\beta,t} = \vg_t + \beta\, \vm_{\beta,t-1}, \quad \vm_{\beta,0} = \mathbf{0},
\end{equation}
which is exactly the same as how typical autograd frameworks implement momentum.
From $R_{t,k} = \mathbb{E}[\vg_t\, \vg_{t-k}^\top]$ and $T_{t,k} = \operatorname{Tr}(R_{t,k})$, we have
\begin{equation}
S_t(\beta) = \sum_{k=0}^{\infty} T_{t,k}\, \beta^k
= \sum_{k=0}^{\infty} \operatorname{Tr}(\mathbb{E}[\vg_t\, \vg_{t-k}^\top])\, \beta^k
= \operatorname{Tr}\!\left(\mathbb{E}\!\left[\vg_t \sum_{k=0}^{\infty} \beta^k \vg_{t-k}^\top\right]\right)
= \mathbb{E}[\vg_t^\top\, \vm_{\beta,t}],
\end{equation}
where the last equality uses $\operatorname{Tr}(\vg_t\, \vm_{\beta,t}^\top) = \vg_t^\top\, \vm_{\beta,t}$.
Substituting into $A(\beta;t) = \frac{\sqrt{1-\beta^2}}{\sqrt{d}} S_t(\beta)$, we can rewrite the optimal momentum as
\begin{equation}
\beta^\star_t = \arg\max_{\beta \in \gB} A(\beta; t) = \arg\max_{\beta \in \gB} \sqrt{1 - \beta^2}\, \mathbb{E}[\vg_t^\top\, \vm_{\beta,t}],
\end{equation}
where $\vm_{\beta,t} = \sum_{k=0}^{\infty} \beta^k \vg_{t-k}$ is the unnormalized momentum at time $t$ with momentum parameter $\beta$.
This completes the proof.
Note that, in theory, the expectation is taken over the entire possible gradient sequence $\vg_t$, which should be approximated in the real-world application.
\end{proof}

\subsection{Proof of Corollary~\ref{cor:ar1-momentum}: One-pole gradients recover one-pole momentum}
\label{sec:appx:proof-of-corollary-ar1-momentum}

\begin{proof}[Proof of Corollary~\ref{cor:ar1-momentum}]
By definition of $\vm_{\beta,t}$,
\begin{equation}
    A = \mathbb E[\vg_t^\top\vm_{\beta,t}]
    = \sum_{k\ge0}\beta^k \mathbb E[\vg_t^\top\vg_{t-k}]
    = r_0\sum_{k\ge0}(\beta\rho)^k
    = \frac{r_0}{1-\beta\rho}.
\end{equation}
The series converges since $|\beta\rho|<1$.
Hence
\begin{equation}
    \sqrt{1-\beta^2}\, \mathbb E[\vg_t^\top\vm_{\beta,t}] = r_0\frac{\sqrt{1-\beta^2}}{1-\beta\rho}.
\end{equation}
Since $r_0>0$, it suffices to maximize $\log A$:
\begin{equation}
    \frac{d}{d\beta}\log A(\beta)
    = -\frac{\beta}{1-\beta^2} + \frac{\rho}{1-\beta\rho}
    = \frac{\rho-\beta}{(1-\beta^2)(1-\beta\rho)}.
\end{equation}
The denominator is positive for $\beta,\rho\in(-1,1)$, so $A$ increases for $\beta<\rho$ and decreases for $\beta>\rho$.
Therefore the unique maximizer over $(-1,1)$ is $\beta^\star=\rho$.
Restricting to $\beta\in[0,\bar\beta]$ with $0<\bar\beta<1$ gives the projection $\max \{0, \min \{\bar\beta, \rho\}\}$.
\end{proof}

\subsection{Proof of Corollary~\ref{cor:greedy-optimal-adam}: Greedy optimal Adam/AdamW}
\label{sec:appx:proof-of-corollary-greedy-optimal-adam}

\begin{proof}[Proof of Corollary~\ref{cor:greedy-optimal-adam}]
We work in the impulse-space as defined in Section~\ref{sec:3_main}.
Throughout this proof, we condition on the realized second-moment state $\vc_{\beta_2,t}$ for each candidate $\beta_2$ and treat it as a fixed time-varying diagonal preconditioner.
All expectations below may therefore be read as conditional expectations given this state.

For the Adam family we use the weighted diagonal trust-region norm
\begin{equation}
\|Q\|_{\gD,\vc}^2 = \sum_{j=1}^d c_j \sum_{k\ge 0} |q_{j,k}|^2,
\end{equation}
where $k$ is the lag index.

By the main-text definitions, the Adam family is $\gQ_{\eta,\beta_1,\beta_2} = \gQ_\textnormal{D}(B,\vc) \cap \gQ_\textnormal{1p}(\vc)$, where $\gQ_\textnormal{D}(B,\vc) = \{\operatorname{diag}(q_{j}) : \sum_j c_j \sum_{k\ge 0} |q_{j,k}|^2 \le B\}$ is the elementwise weighted norm-bounded family and $\gQ_\textnormal{1p}(\vc) = \{q_{j,k} = \eta(1-\beta_1)\beta_1^k c_j^{-1} : \eta \ge 0,\, 0 < \beta_1 < 1\}$ is the Adam 1-pole family.
We solve $\max_{Q \in \gQ_{\eta,\beta_1,\beta_2}} \langle Q, R\rangle$ by parameterizing $Q = Q_{\text{Adam}}$ via the diagonal 1-pole impulse response and saturating the trust region.

\emph{Step 1 — Norm of the diagonal optimizer.}
For each parameter coordinate $j$, we have the running second moment and the coordinate-wise cost:
\begin{equation}
v_{\beta_2, t, j} = (1-\beta_2) \sum_{k=0}^{\infty} \beta_2^k g_{t-k,j}^2, \quad c_{\beta_2, t, j} = (v_{\beta_2, t, j} + \epsilon)^{1/2}.
\end{equation}
From the definition of $\gQ_\textnormal{1p}(\vc)$, the per-coordinate impulse response at time $t$ is:
\begin{equation}
q_{t,j,k} = \eta\, (1-\beta_1)\, \beta_1^k / c_{\beta_2, t, j}.
\end{equation}
Therefore, the weighted norm of $\gQ_\textnormal{D}(B,\vc)$ evaluates to:
\begin{equation}\label{eq:proof:norm-of-adam}
\|Q_{\text{Adam}}\|_{\gD,\vc}^2 = \sum_{j=1}^d c_{\beta_2,t,j} \sum_{k\ge 0}|q_{t,j,k}|^2
= \eta^2\, \frac{(1-\beta_1)^2}{1-\beta_1^2} \sum_{j=1}^d \frac{1}{c_{\beta_2,t,j}}
= \eta^2\, \frac{1-\beta_1}{1+\beta_1}\, W_{\beta_2, t},
\end{equation}
where $W_{\beta_2, t} \coloneqq \textstyle\sum_j 1/c_{\beta_2,t,j}$ is the normalization factor.

\emph{Step 2 — Alignment with the moment operator.}
Since $Q_{\text{Adam}}$ is diagonal, only the diagonal entries of $R_{t,k}$ enter the inner product; let $r_{t,k,j} \coloneqq R_{t,k,jj} = \mathbb{E}[g_{t,j}\, g_{t-k,j} \mid \vc_{\beta_2,t}]$. Then
\begin{align}
\langle Q_{\text{Adam}},R\rangle &= \sum_{k\ge0}\operatorname{Tr}(Q_k^\top R_{t,k})
= \sum_{k\ge0}\sum_{j=1}^d q_{t,j,k}\, r_{t,k,j} \\
&= \eta(1-\beta_1)\sum_{k\ge 0}\beta_1^k \sum_{j=1}^d \frac{r_{t,k,j}}{c_{\beta_2, t, j}} \\
&= \eta(1-\beta_1)\sum_{k\ge 0}\beta_1^k\, T_{t,k}(\beta_2),
\end{align}
where $T_{t,k}(\beta_2) \coloneqq \textstyle\sum_j r_{t,k,j}/c_{\beta_2,t,j}$ is the weighted trace of the moment operator.

\emph{Step 3 — Reduce to 1-D search; saturate trust region.}
For fixed $(\beta_1,\beta_2)$, define the score
\begin{equation}
S_t(\beta_1,\beta_2) \coloneqq \sum_{k\ge 0} \beta_1^k\, T_{t,k}(\beta_2).
\end{equation}
The inner product is linear in $\eta$ while the constraint is quadratic.
If $S_t(\beta_1,\beta_2) > 0$, the maximizer saturates the trust region; if $S_t(\beta_1,\beta_2) \le 0$, the best gain is $\eta = 0$.
In the closure of the admissible range, $\beta_1 = 0$ gives $S_t(0,\beta_2) = T_{t,0}(\beta_2) = \sum_j \mathbb{E}[g_{t,j}^2 \mid \vc_{\beta_2,t}]/c_{\beta_2,t,j} > 0$ whenever the gradient is nonzero, so the global nontrivial maximizer has positive score.

The trust-region-normalized gain is
\begin{equation}
A(\beta_1,\beta_2;t) \coloneqq \frac{\langle Q_{\text{Adam}},R\rangle}{\|Q_{\text{Adam}}\|_{\gD,\vc}} = \frac{\sqrt{1-\beta_1^2}}{\sqrt{W_{\beta_2,t}}}\, S_t(\beta_1,\beta_2).
\end{equation}
Therefore, we have the optimal Adam hyperparameters as
\begin{equation}
(\beta_1^\star,\beta_2^\star)_t = \arg\max_{0<(\beta_1,\beta_2)<1} A(\beta_1,\beta_2;t),
\end{equation}
and the corresponding trust-region-saturating learning rate is
\begin{equation}
\eta^\star_t = \sqrt{B}\, / \, \|Q_{\text{Adam}}\|_{\gD,\vc} = \sqrt{B}\, \sqrt{\frac{1+\beta_1^\star}{(1-\beta_1^\star)\, W_{\beta_2^\star,t}}}.
\end{equation}
This follows from equation~\ref{eq:proof:norm-of-adam} above and the diagonal trust region definition:
\begin{equation}
\gQ_\textnormal{D}(B,\vc) \coloneqq \{\operatorname{diag}(q_{j}) : \textstyle\sum_j c_j \textstyle\sum_{k\ge 0} |q_{j,k}|^2\le B\}.
\end{equation}

\emph{Step 4 — Solving $A$ for streaming gradients.}
Recall $r_{t,k,j} = R_{t,k,jj} = \mathbb{E}[g_{t,j}\, g_{t-k,j} \mid \vc_{\beta_2,t}]$ for the diagonal entries at time $t$ and lag $k$. Then
\begin{equation}
T_{t,k}(\beta_2) = \sum_{j=1}^d \frac{r_{t,k,j}}{c_{\beta_2,t,j}} = \sum_{j=1}^d \mathbb{E}\!\left[\frac{g_{t,j}\, g_{t-k,j}}{c_{\beta_2,t,j}} \,\Big|\, \vc_{\beta_2,t}\right].
\end{equation}
\begin{align}
S_t(\beta_1,\beta_2) &= \sum_{k=0}^{\infty} \beta_1^k\, T_{t,k}(\beta_2) \\
&= \sum_{k=0}^{\infty} \beta_1^k \sum_{j=1}^d \mathbb{E}\!\left[\frac{g_{t,j}\, g_{t-k,j}}{c_{\beta_2,t,j}} \,\Big|\, \vc_{\beta_2,t}\right] \\
&= \sum_{j=1}^d \mathbb{E}\!\left[\frac{g_{t,j}}{c_{\beta_2,t,j}} \sum_{k=0}^{\infty} \beta_1^k g_{t-k,j} \,\Big|\, \vc_{\beta_2,t}\right] \\
&= \frac{1}{1 - \beta_1} \sum_{j=1}^d \mathbb{E}\!\left[g_{t,j}\, \frac{\mu_{\beta_1,t,j}}{c_{\beta_2,t,j}} \,\Big|\, \vc_{\beta_2,t}\right] \\
&= \frac{1}{1 - \beta_1} \mathbb{E}\!\left[\sum_{j=1}^d g_{t,j}\, \frac{\mu_{\beta_1,t,j}}{(v_{\beta_2,t,j} + \epsilon)^{1/2}} \,\Big|\, \vc_{\beta_2,t}\right] \\
&= \frac{1}{1 - \beta_1} \mathbb{E}[\vg_t^\top\, \vu_{\beta_1,\beta_2,t} \mid \vc_{\beta_2,t}],
\end{align}
where
\begin{align}
\mu_{\beta_1,t,j} &= (1-\beta_1) \sum_{k=0}^{\infty} \beta_1^k g_{t-k,j} = \beta_1\, \mu_{\beta_1,t-1,j} + (1-\beta_1)\, g_{t,j}, \\
v_{\beta_2,t,j} &= (1-\beta_2) \sum_{k=0}^{\infty} \beta_2^k g_{t-k,j}^2 = \beta_2\, v_{\beta_2,t-1,j} + (1-\beta_2)\, g_{t,j}^2,
\end{align}
are the normalized first and second moments for coordinate $j$ at time $t$, and
\begin{equation}
\vu_{\beta_1,\beta_2,t} = \vmu_{\beta_1,t}\odot\vc_{\beta_2,t}^{-1}, \qquad u_{\beta_1,\beta_2,t,j} = \frac{\mu_{\beta_1,t,j}}{c_{\beta_2,t,j}} = \frac{\mu_{\beta_1,t,j}}{(v_{\beta_2,t,j} + \epsilon)^{1/2}},
\end{equation}
is the Adam update direction at time $t$, matching the main text.
Substituting into the expression for $A$:
\begin{align}
A(\beta_1,\beta_2; t) &= \frac{\sqrt{1-\beta_1^2}}{\sqrt{W_{\beta_2,t}}}\, S_t(\beta_1,\beta_2) \\
&= \frac{\sqrt{1-\beta_1^2}}{(1-\beta_1)\sqrt{W_{\beta_2,t}}}\, \mathbb{E}[\vg_t^\top\, \vu_{\beta_1,\beta_2,t} \mid \vc_{\beta_2,t}] \\
&= \sqrt{\frac{1+\beta_1}{1-\beta_1}}\, \frac{\mathbb{E}[\vg_t^\top\, \vu_{\beta_1,\beta_2,t} \mid \vc_{\beta_2,t}]}{\sqrt{W_{\beta_2,t}}}.
\end{align}
Therefore, we can rewrite the optimal Adam hyperparameters as
\begin{equation}
(\beta_1^\star,\beta_2^\star)_t = \arg\max_{0<\beta_1<1,\, 0<\beta_2<1} \sqrt{\frac{1+\beta_1}{1-\beta_1}}\, \frac{\mathbb{E}[\vg_t^\top\, \vu_{\beta_1,\beta_2,t} \mid \vc_{\beta_2,t}]}{\sqrt{W_{\beta_2,t}}}.
\end{equation}
This completes the proof.
Note that, in theory, the expectation is taken over the entire possible gradient sequence $\vg_t$ conditioned on the second-moment state, which should be approximated in the real-world application.
\end{proof}

\subsection{Proof of Theorem~\ref{thm:finite-step-descent-bridge}: Finite-step descent}
\label{sec:appx:proof-of-theorem-finite-step-descent-bridge}

\begin{proof}[Proof of Theorem~\ref{thm:finite-step-descent-bridge}]
Since $\gL$ has $L_s$-Lipschitz gradient, for any $\vx, \vh \in \R^d$,
\begin{equation}
\gL(\vx + \vh)
= \gL(\vx) + \nabla_\vx \gL(\vx)^\top \vh
+ \int_0^1 \bigl(\nabla_\vx \gL(\vx + s\vh) - \nabla_\vx \gL(\vx)\bigr)^\top \vh \, ds.
\end{equation}
Apply this with $\vx = \vtheta$ and $\vh = -\eta \vu$:
\begin{equation}
\gL(\vtheta - \eta \vu)
= \gL(\vtheta) - \eta \nabla_\vtheta \gL(\vtheta)^\top \vu
- \int_0^1 \bigl(\nabla_\vtheta \gL(\vtheta - s \eta \vu) - \nabla_\vtheta \gL(\vtheta)\bigr)^\top \eta \vu \, ds.
\end{equation}
Since $\vg = \nabla_\vtheta \gL(\vtheta)$,
\begin{equation}
D_\eta(\vu) - \eta A(\vu)
= \int_0^1 \bigl(\nabla_\vtheta \gL(\vtheta - s \eta \vu) - \nabla_\vtheta \gL(\vtheta)\bigr)^\top \eta \vu \, ds.
\end{equation}
Taking absolute values and using Lipschitz continuity of $\nabla_\vtheta \gL$:
\begin{align*}
\bigl| D_\eta(\vu) - \eta A(\vu) \bigr|
&\le \int_0^1 \|\nabla_\vtheta \gL(\vtheta - s \eta \vu) - \nabla_\vtheta \gL(\vtheta)\| \cdot \eta \|\vu\| \, ds \\
&\le \int_0^1 L_s s \eta \|\vu\| \cdot \eta \|\vu\| \, ds
= L_s \eta^2 \|\vu\|^2 \int_0^1 s \, ds
= \frac{L_s \eta^2}{2} \|\vu\|^2. \qedhere
\end{align*}
\end{proof}

\subsection{Proof of Corollary~\ref{cor:greedy-near-optimal}: Near-optimality of greedy alignment}
\label{sec:appx:proof-of-corollary-greedy-near-optimal}

\begin{proof}[Proof of Corollary~\ref{cor:greedy-near-optimal}]
For any $\vv \in \mathcal{U}$, by Theorem~\ref{thm:finite-step-descent-bridge},
$D_\eta(\vu^\star) \ge \eta A(\vu^\star) - \frac{L_s \eta^2}{2} \rho^2$.
Since $\vu^\star$ maximizes $A$, we have $A(\vu^\star) \ge A(\vv)$, so
$D_\eta(\vu^\star) \ge \eta A(\vv) - \frac{L_s \eta^2}{2} \rho^2$.
Again by Theorem~\ref{thm:finite-step-descent-bridge},
$D_\eta(\vv) \le \eta A(\vv) + \frac{L_s \eta^2}{2} \rho^2$.
Combining yields
$D_\eta(\vu^\star) \ge D_\eta(\vv) - L_s \eta^2 \rho^2$.
Since $\vv$ was arbitrary, the result follows.
\end{proof}

\subsection{Proof of Corollary~\ref{cor:kswitch-exact}: Exact finite-candidate $K$-Switch}
\label{sec:appx:proof-of-corollary-kswitch-exact}

\begin{proof}[Proof of Corollary~\ref{cor:kswitch-exact}]
The normalized maps $\tilde Q_{k,t}$ are causal because each $Q_{k,t}$ is causal and $a_{k,t}$ is a causal scalar normalization.
Consider any element $Q\in\operatorname{co}(\tilde{\mathcal F}_t)$.
By definition of the finite convex hull, there exist coefficients $\pi_k\ge0$, $\sum_{k=1}^K\pi_k=1$, such that
\begin{equation}
\label{eq:proof:convex-combination}
Q \;=\; \sum_{k=1}^K \pi_k\tilde Q_{k,t}.
\end{equation}
The action of this convex combination on the gradient stream is the pointwise convex combination of the corresponding update streams:
\begin{equation}
\label{eq:proof:pointwise-convex}
(Q \vg)_t \;=\; \sum_{k=1}^K \pi_k(\tilde Q_{k,t}\vg)_t.
\end{equation}
Therefore, by linearity of the inner product in the update and linearity of expectation,
\begin{equation}
\label{eq:proof:linearity-chain}
\begin{aligned}
P_t(Q)
&\;=\; \mathbb E\!\left[ \vg_t^\top (Q \vg)_t \right]  \\
&\;=\; \mathbb E\!\left[ \vg_t^\top \sum_{k=1}^K \pi_k(\tilde Q_{k,t}\vg)_t \right]  \\
&\;=\; \sum_{k=1}^K \pi_k \mathbb E\!\left[ \vg_t^\top(\tilde Q_{k,t}\vg)_t \right]  \\
&\;=\; \sum_{k=1}^K \pi_k P_t(\tilde Q_{k,t})  \\
&\;\le\; \max_{1\le k\le K}P_t(\tilde Q_{k,t}).
\end{aligned}
\end{equation}
Taking the supremum over all $Q\in\operatorname{co}(\tilde{\mathcal F}_t)$ gives
\begin{equation}
\label{eq:proof:sup-le-max}
\sup_{Q\in\operatorname{co}(\tilde{\mathcal F}_t)}P_t(Q)
\;\le\; \max_{1\le k\le K}P_t(\tilde Q_{k,t}).
\end{equation}
The reverse inequality holds because every $\tilde Q_{k,t}$ belongs to $\operatorname{co}(\tilde{\mathcal F}_t)$. Hence
\begin{equation}
\label{eq:proof:sup-eq-max}
\sup_{Q\in\operatorname{co}(\tilde{\mathcal F}_t)}P_t(Q)
\;=\; \max_{1\le k\le K}P_t(\tilde Q_{k,t}).
\end{equation}
Finally, by definition of $\tilde Q_{k,t}$,
\begin{equation}
\label{eq:proof:score-definition}
P_t(\tilde Q_{k,t})
\;=\; \mathbb E\!\left[ \vg_t^\top(\tilde Q_{k,t}\vg)_t \right]
\;=\; \mathbb E\!\left[ a_{k,t}\vg_t^\top \vu_{k,t} \right].
\end{equation}
This proves the displayed equality.

It remains to prove the active-support claim. Let
\begin{equation}
\label{eq:proof:M-definition}
P^\star \;\coloneqq\; \max_{1\le k\le K}P_t(\tilde Q_{k,t}).
\end{equation}
Suppose
\begin{equation}
\label{eq:proof:Qstar-definition}
Q^\star \;=\; \sum_{k=1}^K\pi_k\tilde Q_{k,t}, \qquad
\pi_k\ge0,\quad \sum_{k=1}^K\pi_k=1,
\end{equation}
attains the supremum over the convex hull.
Then
\begin{equation}\label{eq:proof:weighted-average}
P^\star
\;=\; P_t(Q^\star)
\;=\; \sum_{k=1}^K\pi_kP_t(\tilde Q_{k,t})
\;\le\; \sum_{k=1}^K\pi_k P^\star
\;=\; P^\star.
\end{equation}
Thus the weighted average equals its upper bound.
If there existed an active index $i$ with $\pi_i>0$ and $P_t(\tilde Q_{i,t}) < P^\star$, then the weighted average would be strictly smaller than $P^\star$, a contradiction.
Therefore every active component satisfies $P_t(\tilde Q_{k,t})= P^\star$.
The stated $K$-Switch selector is exactly the argmax over these finite candidate scores.
This completes the proof.
\end{proof}

\subsection{Proof of Proposition~\ref{prop:online-stability}: Online stability}
\label{sec:appx:proof-of-proposition-online-stability}

\begin{proof}
By the uniform error bound and optimality of $\hat k$ for the estimated scores,
\begin{equation}\label{eq:sec_4:online_stability}
A_{k^\star}
\;\le\; \hat A_{k^\star}+\varepsilon
\;\le\; \hat A_{\hat k}+\varepsilon
\;\le\; A_{\hat k}+2\varepsilon.
\end{equation}
Therefore, $A_{k^\star}-A_{\hat k}\le2\varepsilon$.
If the maximizer is unique with a gap $\Delta \coloneqq A_{k^\star}-\max_{j\ne k^\star}A_j > 2\varepsilon$ and $\hat k\ne k^\star$, then $A_{k^\star}-A_{\hat k}\ge\Delta>2\varepsilon$, contradicting the first claim.
Hence $\hat k=k^\star$.
\end{proof}

%%%%%%%%%%%%%%%%%%%%%%%%%%%%%%%%%%%%%%%%%%%%%%%%%%%%%%%%%%%%

\newpage
\section*{NeurIPS Paper Checklist}

\begin{enumerate}

\item {\bf Claims}
    \item[] Question: Do the main claims made in the abstract and introduction accurately reflect the paper's contributions and scope?
    \item[] Answer: \answerYes{}
    \item[] Justification:
    The abstract and introduction (Section~\ref{sec:1_intro}) state four contributions---(i) lifting greedy alignment to an operator-level principle, (ii) well-posedness and stability for compact families, (iii) reduction to hyperparameter scoring, and (iv) dynamic momentum rules---and each is supported by Sections~\ref{sec:2_prelim}, \ref{sec:3_main}, \ref{sec:4_impl}, and \ref{sec:5_exp}, respectively, with proofs in Appendix~\ref{sec:appx:proofs}.
    \item[] Guidelines:
    \begin{itemize}
        \item The answer \answerNA{} means that the abstract and introduction do not include the claims made in the paper.
        \item The abstract and/or introduction should clearly state the claims made, including the contributions made in the paper and important assumptions and limitations. A \answerNo{} or \answerNA{} answer to this question will not be perceived well by the reviewers.
        \item The claims made should match theoretical and experimental results, and reflect how much the results can be expected to generalize to other settings.
        \item It is fine to include aspirational goals as motivation as long as it is clear that these goals are not attained by the paper.
    \end{itemize}

\item {\bf Limitations}
    \item[] Question: Does the paper discuss the limitations of the work performed by the authors?
    \item[] Answer: \answerYes{}
    \item[] Justification:
    See Section~\ref{sec:7_limit} (``Limitations and scope'').
    \item[] Guidelines:
    \begin{itemize}
        \item The answer \answerNA{} means that the paper has no limitation while the answer \answerNo{} means that the paper has limitations, but those are not discussed in the paper.
        \item The authors are encouraged to create a separate ``Limitations'' section in their paper.
        \item The paper should point out any strong assumptions and how robust the results are to violations of these assumptions (e.g., independence assumptions, noiseless settings, model well-specification, asymptotic approximations only holding locally). The authors should reflect on how these assumptions might be violated in practice and what the implications would be.
        \item The authors should reflect on the scope of the claims made, e.g., if the approach was only tested on a few datasets or with a few runs. In general, empirical results often depend on implicit assumptions, which should be articulated.
        \item The authors should reflect on the factors that influence the performance of the approach. For example, a facial recognition algorithm may perform poorly when image resolution is low or images are taken in low lighting. Or a speech-to-text system might not be used reliably to provide closed captions for online lectures because it fails to handle technical jargon.
        \item The authors should discuss the computational efficiency of the proposed algorithms and how they scale with dataset size.
        \item If applicable, the authors should discuss possible limitations of their approach to address problems of privacy and fairness.
        \item While the authors might fear that complete honesty about limitations might be used by reviewers as grounds for rejection, a worse outcome might be that reviewers discover limitations that aren't acknowledged in the paper. The authors should use their best judgment and recognize that individual actions in favor of transparency play an important role in developing norms that preserve the integrity of the community. Reviewers will be specifically instructed to not penalize honesty concerning limitations.
    \end{itemize}

\item {\bf Theory assumptions and proofs}
    \item[] Question: For each theoretical result, does the paper provide the full set of assumptions and a complete (and correct) proof?
    \item[] Answer: \answerYes{}
    \item[] Justification:
    Assumptions (compactness of $\gQ$, stability/causality of linear filters, $\ell^1$-summability of $R$, etc.) are stated alongside each result in Sections~\ref{sec:2_prelim}--\ref{sec:3_main}.
    Complete proofs are provided in Appendix~\ref{sec:appx:proofs}.
    \item[] Guidelines:
    \begin{itemize}
        \item The answer \answerNA{} means that the paper does not include theoretical results.
        \item All the theorems, formulas, and proofs in the paper should be numbered and cross-referenced.
        \item All assumptions should be clearly stated or referenced in the statement of any theorems.
        \item The proofs can either appear in the main paper or the supplemental material, but if they appear in the supplemental material, the authors are encouraged to provide a short proof sketch to provide intuition.
        \item Inversely, any informal proof provided in the core of the paper should be complemented by formal proofs provided in appendix or supplemental material.
        \item Theorems and Lemmas that the proof relies upon should be properly referenced.
    \end{itemize}

    \item {\bf Experimental result reproducibility}
    \item[] Question: Does the paper fully disclose all the information needed to reproduce the main experimental results of the paper to the extent that it affects the main claims and/or conclusions of the paper (regardless of whether the code and data are provided or not)?
    \item[] Answer: \answerYes{}
    \item[] Justification:
    Algorithms are in Section~\ref{sec:4_impl}, implementation details are in Section~\ref{sec:5_exp} and in Appendix~\ref{sec:appx:exp:implementation} All datasets used (CIFAR-100, MetaMathQA-395K, GSM8K, Commonsense-170K, Cars, CUB-200, DTD, Food-101, RESISC45, SUN397) are publicly available with citations.
    \item[] Guidelines:
    \begin{itemize}
        \item The answer \answerNA{} means that the paper does not include experiments.
        \item If the paper includes experiments, a \answerNo{} answer to this question will not be perceived well by the reviewers: Making the paper reproducible is important, regardless of whether the code and data are provided or not.
        \item If the contribution is a dataset and\slash or model, the authors should describe the steps taken to make their results reproducible or verifiable.
        \item Depending on the contribution, reproducibility can be accomplished in various ways. For example, if the contribution is a novel architecture, describing the architecture fully might suffice, or if the contribution is a specific model and empirical evaluation, it may be necessary to either make it possible for others to replicate the model with the same dataset, or provide access to the model. In general. releasing code and data is often one good way to accomplish this, but reproducibility can also be provided via detailed instructions for how to replicate the results, access to a hosted model (e.g., in the case of a large language model), releasing of a model checkpoint, or other means that are appropriate to the research performed.
        \item While NeurIPS does not require releasing code, the conference does require all submissions to provide some reasonable avenue for reproducibility, which may depend on the nature of the contribution. For example
        \begin{enumerate}
            \item If the contribution is primarily a new algorithm, the paper should make it clear how to reproduce that algorithm.
            \item If the contribution is primarily a new model architecture, the paper should describe the architecture clearly and fully.
            \item If the contribution is a new model (e.g., a large language model), then there should either be a way to access this model for reproducing the results or a way to reproduce the model (e.g., with an open-source dataset or instructions for how to construct the dataset).
            \item We recognize that reproducibility may be tricky in some cases, in which case authors are welcome to describe the particular way they provide for reproducibility. In the case of closed-source models, it may be that access to the model is limited in some way (e.g., to registered users), but it should be possible for other researchers to have some path to reproducing or verifying the results.
        \end{enumerate}
    \end{itemize}

\item {\bf Open access to data and code}
    \item[] Question: Does the paper provide open access to the data and code, with sufficient instructions to faithfully reproduce the main experimental results, as described in supplemental material?
    \item[] Answer: \answerYes{}
    \item[] Justification:
    All datasets we used are public and cited in Section~\ref{sec:5_exp} and Appendix~\ref{sec:appx:exp}, and the algorithm and all hyperparameters are described in sufficient detail to reproduce results from scratch (Section~\ref{sec:4_impl} and Appendix~\ref{sec:appx:exp:implementation}).
    The code is in the supplementary material.
    \item[] Guidelines:
    \begin{itemize}
        \item The answer \answerNA{} means that paper does not include experiments requiring code.
        \item Please see the NeurIPS code and data submission guidelines (\url{https://neurips.cc/public/guides/CodeSubmissionPolicy}) for more details.
        \item While we encourage the release of code and data, we understand that this might not be possible, so \answerNo{} is an acceptable answer. Papers cannot be rejected simply for not including code, unless this is central to the contribution (e.g., for a new open-source benchmark).
        \item The instructions should contain the exact command and environment needed to run to reproduce the results. See the NeurIPS code and data submission guidelines (\url{https://neurips.cc/public/guides/CodeSubmissionPolicy}) for more details.
        \item The authors should provide instructions on data access and preparation, including how to access the raw data, preprocessed data, intermediate data, and generated data, etc.
        \item The authors should provide scripts to reproduce all experimental results for the new proposed method and baselines. If only a subset of experiments are reproducible, they should state which ones are omitted from the script and why.
        \item At submission time, to preserve anonymity, the authors should release anonymized versions (if applicable).
        \item Providing as much information as possible in supplemental material (appended to the paper) is recommended, but including URLs to data and code is permitted.
    \end{itemize}

\item {\bf Experimental setting/details}
    \item[] Question: Does the paper specify all the training and test details (e.g., data splits, hyperparameters, how they were chosen, type of optimizer) necessary to understand the results?
    \item[] Answer: \answerYes{}
    \item[] Justification:
    See Section~\ref{sec:5_exp} and Appendix~\ref{sec:appx:exp:implementation}.
    \item[] Guidelines:
    \begin{itemize}
        \item The answer \answerNA{} means that the paper does not include experiments.
        \item The experimental setting should be presented in the core of the paper to a level of detail that is necessary to appreciate the results and make sense of them.
        \item The full details can be provided either with the code, in appendix, or as supplemental material.
    \end{itemize}

\item {\bf Experiment statistical significance}
    \item[] Question: Does the paper report error bars suitably and correctly defined or other appropriate information about the statistical significance of the experiments?
    \item[] Answer: \answerYes{}
    \item[] Justification:
    All result tables in Section~\ref{sec:5_exp} (Tables~\ref{tab:gemma_mathqa}, \ref{tab:llama_mathqa}, \ref{tab:llm_commonsense}, \ref{tab:vit_tasks}) report mean $\pm$ standard deviation over multiple independent runs with different random seeds.
    Figures~\ref{fig:sgd_momentum_test_acc_errbar} and~\ref{fig:adam_test_acc_errbar} show error bars.
    \item[] Guidelines:
    \begin{itemize}
        \item The answer \answerNA{} means that the paper does not include experiments.
        \item The authors should answer \answerYes{} if the results are accompanied by error bars, confidence intervals, or statistical significance tests, at least for the experiments that support the main claims of the paper.
        \item The factors of variability that the error bars are capturing should be clearly stated (for example, train/test split, initialization, random drawing of some parameter, or overall run with given experimental conditions).
        \item The method for calculating the error bars should be explained (closed form formula, call to a library function, bootstrap, etc.)
        \item The assumptions made should be given (e.g., Normally distributed errors).
        \item It should be clear whether the error bar is the standard deviation or the standard error of the mean.
        \item It is OK to report 1-sigma error bars, but one should state it. The authors should preferably report a 2-sigma error bar than state that they have a 96\% CI, if the hypothesis of Normality of errors is not verified.
        \item For asymmetric distributions, the authors should be careful not to show in tables or figures symmetric error bars that would yield results that are out of range (e.g., negative error rates).
        \item If error bars are reported in tables or plots, the authors should explain in the text how they were calculated and reference the corresponding figures or tables in the text.
    \end{itemize}

\item {\bf Experiments compute resources}
    \item[] Question: For each experiment, does the paper provide sufficient information on the computer resources (type of compute workers, memory, time of execution) needed to reproduce the experiments?
    \item[] Answer: \answerYes{}
    \item[] Justification:
    Per-experiment GPU type and count are listed in the hyperparameter tables of Appendix~\ref{sec:appx:exp:implementation}, and compute time is reported in Section~\ref{sec:5_exp} and Appendix~\ref{sec:appx:exp:implementation}.
    \item[] Guidelines:
    \begin{itemize}
        \item The answer \answerNA{} means that the paper does not include experiments.
        \item The paper should indicate the type of compute workers CPU or GPU, internal cluster, or cloud provider, including relevant memory and storage.
        \item The paper should provide the amount of compute required for each of the individual experimental runs as well as estimate the total compute.
        \item The paper should disclose whether the full research project required more compute than the experiments reported in the paper (e.g., preliminary or failed experiments that didn't make it into the paper).
    \end{itemize}

\item {\bf Code of ethics}
    \item[] Question: Does the research conducted in the paper conform, in every respect, with the NeurIPS Code of Ethics \url{https://neurips.cc/public/EthicsGuidelines}?
    \item[] Answer: \answerYes{}
    \item[] Justification:
    Yes.
    \item[] Guidelines:
    \begin{itemize}
        \item The answer \answerNA{} means that the authors have not reviewed the NeurIPS Code of Ethics.
        \item If the authors answer \answerNo, they should explain the special circumstances that require a deviation from the Code of Ethics.
        \item The authors should make sure to preserve anonymity (e.g., if there is a special consideration due to laws or regulations in their jurisdiction).
    \end{itemize}

\item {\bf Broader impacts}
    \item[] Question: Does the paper discuss both potential positive societal impacts and negative societal impacts of the work performed?
    \item[] Answer: \answerYes{}
    \item[] Justification:
    We have a ``Broader impact'' section in the main paper.
    \item[] Guidelines:
    \begin{itemize}
        \item The answer \answerNA{} means that there is no societal impact of the work performed.
        \item If the authors answer \answerNA{} or \answerNo, they should explain why their work has no societal impact or why the paper does not address societal impact.
        \item Examples of negative societal impacts include potential malicious or unintended uses (e.g., disinformation, generating fake profiles, surveillance), fairness considerations (e.g., deployment of technologies that could make decisions that unfairly impact specific groups), privacy considerations, and security considerations.
        \item The conference expects that many papers will be foundational research and not tied to particular applications, let alone deployments. However, if there is a direct path to any negative applications, the authors should point it out. For example, it is legitimate to point out that an improvement in the quality of generative models could be used to generate Deepfakes for disinformation. On the other hand, it is not needed to point out that a generic algorithm for optimizing neural networks could enable people to train models that generate Deepfakes faster.
        \item The authors should consider possible harms that could arise when the technology is being used as intended and functioning correctly, harms that could arise when the technology is being used as intended but gives incorrect results, and harms following from (intentional or unintentional) misuse of the technology.
        \item If there are negative societal impacts, the authors could also discuss possible mitigation strategies (e.g., gated release of models, providing defenses in addition to attacks, mechanisms for monitoring misuse, mechanisms to monitor how a system learns from feedback over time, improving the efficiency and accessibility of ML).
    \end{itemize}

\item {\bf Safeguards}
    \item[] Question: Does the paper describe safeguards that have been put in place for responsible release of data or models that have a high risk for misuse (e.g., pre-trained language models, image generators, or scraped datasets)?
    \item[] Answer: \answerNA{}
    \item[] Justification:
    The paper releases no new pre-trained models, image generators, or scraped datasets. Our contribution is an optimizer-selection algorithm.
    \item[] Guidelines:
    \begin{itemize}
        \item The answer \answerNA{} means that the paper poses no such risks.
        \item Released models that have a high risk for misuse or dual-use should be released with necessary safeguards to allow for controlled use of the model, for example by requiring that users adhere to usage guidelines or restrictions to access the model or implementing safety filters.
        \item Datasets that have been scraped from the Internet could pose safety risks. The authors should describe how they avoided releasing unsafe images.
        \item We recognize that providing effective safeguards is challenging, and many papers do not require this, but we encourage authors to take this into account and make a best faith effort.
    \end{itemize}

\item {\bf Licenses for existing assets}
    \item[] Question: Are the creators or original owners of assets (e.g., code, data, models), used in the paper, properly credited and are the license and terms of use explicitly mentioned and properly respected?
    \item[] Answer: \answerYes{}
    \item[] Justification:
    Every dataset and every pre-trained model is cited in Section~\ref{sec:5_exp} and Appendix~\ref{sec:appx:exp}, and all are used under their respective public licenses.
    \item[] Guidelines:
    \begin{itemize}
        \item The answer \answerNA{} means that the paper does not use existing assets.
        \item The authors should cite the original paper that produced the code package or dataset.
        \item The authors should state which version of the asset is used and, if possible, include a URL.
        \item The name of the license (e.g., CC-BY 4.0) should be included for each asset.
        \item For scraped data from a particular source (e.g., website), the copyright and terms of service of that source should be provided.
        \item If assets are released, the license, copyright information, and terms of use in the package should be provided. For popular datasets, \url{paperswithcode.com/datasets} has curated licenses for some datasets. Their licensing guide can help determine the license of a dataset.
        \item For existing datasets that are re-packaged, both the original license and the license of the derived asset (if it has changed) should be provided.
        \item If this information is not available online, the authors are encouraged to reach out to the asset's creators.
    \end{itemize}

\item {\bf New assets}
    \item[] Question: Are new assets introduced in the paper well documented and is the documentation provided alongside the assets?
    \item[] Answer: \answerNA{}
    \item[] Justification:
    The paper does not release new datasets, model checkpoints, or other assets. This is a theoretical contribution.
    \item[] Guidelines:
    \begin{itemize}
        \item The answer \answerNA{} means that the paper does not release new assets.
        \item Researchers should communicate the details of the dataset\slash code\slash model as part of their submissions via structured templates. This includes details about training, license, limitations, etc.
        \item The paper should discuss whether and how consent was obtained from people whose asset is used.
        \item At submission time, remember to anonymize your assets (if applicable). You can either create an anonymized URL or include an anonymized zip file.
    \end{itemize}

\item {\bf Crowdsourcing and research with human subjects}
    \item[] Question: For crowdsourcing experiments and research with human subjects, does the paper include the full text of instructions given to participants and screenshots, if applicable, as well as details about compensation (if any)?
    \item[] Answer: \answerNA{}
    \item[] Justification:
    The paper involves no crowdsourcing experiments or research with human subjects.
    \item[] Guidelines:
    \begin{itemize}
        \item The answer \answerNA{} means that the paper does not involve crowdsourcing nor research with human subjects.
        \item Including this information in the supplemental material is fine, but if the main contribution of the paper involves human subjects, then as much detail as possible should be included in the main paper.
        \item According to the NeurIPS Code of Ethics, workers involved in data collection, curation, or other labor should be paid at least the minimum wage in the country of the data collector.
    \end{itemize}

\item {\bf Institutional review board (IRB) approvals or equivalent for research with human subjects}
    \item[] Question: Does the paper describe potential risks incurred by study participants, whether such risks were disclosed to the subjects, and whether Institutional Review Board (IRB) approvals (or an equivalent approval/review based on the requirements of your country or institution) were obtained?
    \item[] Answer: \answerNA{}
    \item[] Justification:
    The paper does not involve human subjects or crowdsourcing, so IRB (or equivalent) approval is not applicable.
    \item[] Guidelines:
    \begin{itemize}
        \item The answer \answerNA{} means that the paper does not involve crowdsourcing nor research with human subjects.
        \item Depending on the country in which research is conducted, IRB approval (or equivalent) may be required for any human subjects research. If you obtained IRB approval, you should clearly state this in the paper.
        \item We recognize that the procedures for this may vary significantly between institutions and locations, and we expect authors to adhere to the NeurIPS Code of Ethics and the guidelines for their institution.
        \item For initial submissions, do not include any information that would break anonymity (if applicable), such as the institution conducting the review.
    \end{itemize}

\item {\bf Declaration of LLM usage}
    \item[] Question: Does the paper describe the usage of LLMs if it is an important, original, or non-standard component of the core methods in this research? Note that if the LLM is used only for writing, editing, or formatting purposes and does \emph{not} impact the core methodology, scientific rigor, or originality of the research, declaration is not required.
    \item[] Answer: \answerNA{}
    \item[] Justification:
    LLMs are used only for writing, editing, or formatting purposes.
    \item[] Guidelines:
    \begin{itemize}
        \item The answer \answerNA{} means that the core method development in this research does not involve LLMs as any important, original, or non-standard components.
        \item Please refer to our LLM policy in the NeurIPS handbook for what should or should not be described.
    \end{itemize}

\end{enumerate}

\end{document}